\documentclass[a4paper,fleqn]{cas-dc}
\usepackage{graphicx}
\usepackage[flushleft]{threeparttable}
\usepackage{multirow}
\usepackage{float}
\usepackage{hyperref}
\usepackage[authoryear]{natbib}
\usepackage{mathtools, cuted}

\DeclareMathOperator*{\argmin}{arg\,min}

\def\tsc#1{\csdef{#1}{\textsc{\lowercase{#1}}\xspace}}
\tsc{WGM}
\tsc{QE}
\tsc{EP}
\tsc{PMS}
\tsc{BEC}
\tsc{DE}

\begin{document}
\let\WriteBookmarks\relax
\def\floatpagepagefraction{1}
\def\textpagefraction{.001}
\shorttitle{A Probabilistic Representation of Deep Learning for Improving The Information Theoretic Interpretability}
\shortauthors{Xinjie Lan et~al.}

\title [mode = title]{A Probabilistic Representation of Deep Learning \\ for Improving The Information Theoretic Interpretability}                      



\author{Xinjie Lan}[
                  orcid=0000-0001-7600-106
                  ]
\ead{lxjbit@udel.edu}






\author{Kenneth E. Barner}




\address{Department of Electrical and Computer Engineering, University of Delaware, Newark, DE, USA, 19711}

\begin{abstract}
In this paper, we propose a probabilistic representation of MultiLayer Perceptrons (MLPs) to improve the information theoretic interpretability.
Above all, we demonstrate that the activations being \textit{i.i.d.}\ is not valid for all the hidden layers of MLPs, thus the existing mutual information estimators based on non-parametric inference methods, e.g., empirical distributions and Kernel Density Estimate (KDE), are invalid for measuring the information flow in MLPs.  
Moreover, we introduce explicit probabilistic explanations for MLPs: (i) we define the probability space $(\Omega_F, \mathcal{T}, P_F)$ for a fully connected layer $\boldsymbol{f}$ and demonstrate the great effect of an activation function of $\boldsymbol{f}$ on the probability measure $P_F$; (ii) we prove the entire architecture of MLPs as a Gibbs distribution $P$; and (iii) the back-propagation aims to optimize the sample space $\Omega_F$ of all the fully connected layers of MLPs for learning an optimal Gibbs distribution $P^*$ to express the statistical connection between the input and the label. 
Based on the probabilistic explanations for MLPs, we improve the information theoretic interpretability of MLPs in three aspects: (i) the random variable of $\boldsymbol{f}$ is discrete and the corresponding entropy is finite; (ii) the information bottleneck theory cannot correctly explain the information flow in MLPs if we take into account the back-propagation; and (iii) we propose novel information theoretic explanations for the generalization of MLPs.
Finally, we demonstrate the proposed probabilistic representation and information theoretic explanations for MLPs in a synthetic dataset and benchmark datasets.
\end{abstract}




\maketitle

\section{Introduction}

Improving the interpretability of Deep Neural Networks (DNNs) is a fundamental issue of deep learning. 
Recently, numerous efforts have been devoted to explaining DNNs from the view point of information theory.
In the seminal work, \citet{DNN-information} initially use the Information Bottleneck (IB) theory to clarify the internal logic of DNNs.
Specifically, they claim that DNNs optimize an IB tradeoff between compression and prediction, and the generalization performance of DNNs is causally related to the compression.
However, the IB explanation causes serious controversies, especially \citet{IP-argue} question the validity of the IB explanations by some counter-examples, and \citet{goldfeld2019estimating} doubt the causality between the compression and the generalization performance of DNNs.

Basically, the above controversies stem from different probabilistic models for the hidden layer of DNNs.
Due to the complicated architecture of DNNs, it is extremely hard to establish an explicit probabilistic model for the hidden layer of DNNs.
As a result, all the previous works have to adopt non-parametric statistics to estimate the mutual information.
\citet{DNN-information} model the distribution of a hidden layer as the empirical distribution (a.k.a.\ the binning method) of the activations of the layer, whereas \citet{IP-argue} model the distribution as Kernel Density Estimation (KDE), and \citet{goldfeld2019estimating} model the distribution as the convolution between the empirical distribution and additive Gaussian noise.
Inevitably, different probabilistic models derive different information theoretic explanations for DNNs, thereby leading to controversies.

\pagebreak
Notably, the non-parametric statistical models lack solid theoretical basis in the context of DNNs.
As two classical non-parametric inference algorithms \citep{wasserman2006all}, the empirical distribution and KDE approach the true distribution only if the samples are independently and identically distributed (\textit{i.i.d.}). 
Specifically, the prerequisite of applying the non-parametric statistics in DNNs is that the activations of a hidden layer are \textit{i.i.d.}\ samples of the true distribution of the layer.
However, none of previous works explicitly demonstrates the prerequisite.

Moreover, the unclear definition for the random variable of a hidden layer results in an information theoretic issue \citep{chelombiev2018adaptive}.
Specifically, a random variable is a measurable function ${F}: \Omega \rightarrow E$ mapping the sample space $\Omega$ to the measurable space $E$.
All the previous works simply assume the activations of a hidden layer as $E$ but not specify $\Omega$, which indciates $F$ as a continuous random variable because the activations are continuous.
As a result, the conditional distribution $P(F|\boldsymbol{X})$ would be a delta function under the assumption that DNNs are deterministic models, thereby the mutual information $I(\boldsymbol{X}, F) = \infty$, where $\boldsymbol{X}$ is the random variable of the input. 
However, that contradicts experimental results $I(\boldsymbol{X}, F) < \infty$.

To resolve the above information theoretic controversies and further improve the interpretability for DNNs, this paper proposes a probabilistic representation for feedforward fully connected DNNs, i.e., the MultiLayer Perceptrons (MLPs), in three aspects: (i) we thoroughly study the \textit{i.i.d.}\ property of the activations of a fully connected layer, (ii) we define the probability space for a fully connected layer, and (iii) we explicitly propose probabilistic explanations for MLPs and the back-propagation training algorithm.

\pagebreak
First, we demonstrate that the correlation of activations with the same label becomes larger as the layer containing the activations is closer to the output.
Therefore, activations  being \textit{i.i.d.}\ is not valid for all the hidden layers of MLPs.
In other words, the existing mutual information estimators based on non-parametric statistics are not valid for all the hidden layers of MLPs as the activations of hidden layers cannot satisfy the prerequisite.

Second, we define the probability space $(\Omega_F, \mathcal{T}, P_F)$ for a fully connected layer $\boldsymbol{f}$ with $N$ neurons given the input $\boldsymbol{x}$.
Let the experiment be $\boldsymbol{f}$ extracting a single feature of $\boldsymbol{x}$, $(\Omega_F, \mathcal{T}, P_F)$ is defined as follows: 
the sample space $\Omega_F$ consists of $N$ possible outcomes (i.e., features), and each outcome is defined by the weights of each neuron; 
the event space $\mathcal{T}$ is the $\sigma$-algebra; 
and the probability measure $P_F$ is a Gibbs measure for quantifying the probability of each outcome occurring the experiment. 
Notably, the activation function of $\boldsymbol{f}$ has a great effect on $P_F$, because an activation equals the negative energy function of $P_F$.

Third, we propose probabilistic explanations for MLPs and the back-propagation training: (i) we prove the entire architecture of MLPs as a Gibbs distribution based on the Gibbs distribution $P_F$ for each layer;
and (ii) we show that the back-propagation training aims to optimize the sample space of all the layers of MLPs for modeling the statistical connection between the input $\boldsymbol{x}$ and the label $\boldsymbol{y}$, because the weights of each layer define sample space.

In summary, the three probabilistic explanations for fully connected layers and MLPs establish a solid probabilistic foundation for explaining MLPs in an information theoretic way.
Based on the probabilistic foundation, we propose three novel information theoretic explanations for MLPs.

Above all, we demonstrate that the entropy of $F$ is finite, i.e., $H(F) < \infty$.
Based on $(\Omega_F, \mathcal{T}, P_F)$, we can explicitly define the random variable of $\boldsymbol{f}$ as ${F}: \Omega_F \rightarrow E_F$, where $E_F$ denotes discrete measurable space, thus $F$ is a discrete random variable and $H(F) < \infty$.
As a result, we resolve the controversy regarding $F$ being continuous.

Furthermore, we demonstrate that the information flow of $\boldsymbol{X}$ and $Y$ in MLPs cannot satisfy IB if taking into account the back-propagation training.
Specifically, the probabilistic explanation for the back-propagation training indicates that $\Omega_F$ depends on both $\boldsymbol{x}$ and $y$, thus $F$ depends on both $\boldsymbol{X}$ and $Y$, where $Y$ is the random variable of $y$. However, IB requires that $F$ is independent on $Y$ given $\boldsymbol{X}$, 

In addition, we further confirm none causal relationship between the compression and the generalization of MLPs.
Alternatively, we demonstrate that the performance of a MLP depends on the mutual information between the MLP and $\boldsymbol{X}$, i.e., $I(\boldsymbol{X}, \text{MLP})$.
More specifically, we demonstrate all the information of $Y$ coming from $\boldsymbol{X}$, i.e., $H(Y) = I(\boldsymbol{X}, Y)$ (the relation is visualized by the Venn diagram in Figure \ref{Img_mlp_information_venn}), thus $I(\boldsymbol{X}, \text{MLP})$ can be divided into two parts $I(Y, \text{MLP})$ and $I(\bar{X}, \text{MLP})$, where $\bar{X} = Y^c \cap \boldsymbol{X}$ denotes the relative complement $Y$ in $\boldsymbol{X}$.
We demonstrate that the performance of the MLP on training dataset depends on $I(Y, \text{MLP})$, and the generalization of the MLP depends on $I(\bar{X}, \text{MLP})$.

\pagebreak
Finally, we generate a synthetic dataset to demonstrate the theoretical explanations for MLPs.
Since the dataset only has four simple features, we can validate the probabilistic explanations for MLPs by visualizing the weights of MLPs.
In addition, the four features has equal probability, thus the dataset has fixed entropy.
As a result, we can demonstrate the information theoretic explanations for MLPs.

The rest of the paper is organized as follows.
Section \ref{related works} briefly discusses the related works.
Section \ref{prob_exp} and \ref{info_exp} propose the probabilistic and information theoretic explanations for MLPs, respectively.
Section \ref{mi_calculation} specifies the mutual information estimators based on $(\Omega_F, \mathcal{T}, P_F)$ for a fully connected layer.
Section \ref{experiments} validates the probabilistic and information theoretic explanations for MLPs on the synthetic dataset and benchmark dataset MNIST and Fashion-MNIST.
Section \ref{conclusions} concludes the paper and discusses future work.

\begin{figure}
\centering
\includegraphics[scale=0.71]{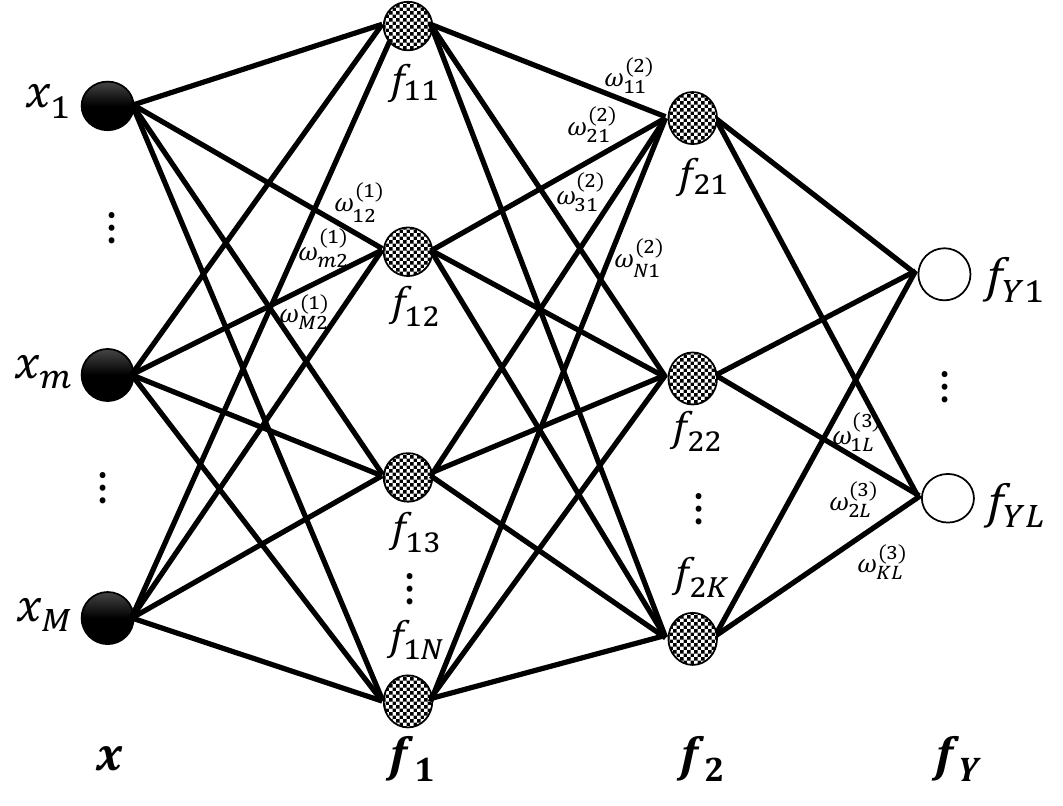}
\caption{
The input layer $\boldsymbol{x}$ has $M$ nodes, and $\boldsymbol{f_1}$ has $N$ neurons $\{f_{1n} = \sigma_1[g_{1n}(\boldsymbol{x})]\}_{n=1}^{N}$, where $g_{1n}(\boldsymbol{x}) = \sum_{m=1}^M\omega^{(1)}_{mn} \cdot x_m + b_{1n}$ is the $n$th linear function with $\omega^{(1)}_{mn}$ being the weight of the edge between $x_m$ and $f_{1n}$, and $b_{1n}$ being the bias.
$\sigma_1(\cdot)$ is a non-linear activation function, e.g., the ReLU function.
Similarly, $\boldsymbol{f_{2}} = \{f_{2k}=\sigma_2[g_{2k}(\boldsymbol{f_1})]\}_{k=1}^K$ has $K$ neurons, where $g_{2k}(\boldsymbol{f_1}) = \sum_{n=1}^N\omega^{(2)}_{nk} \cdot f_{1n} + b_{2k}$.
In addition, $\boldsymbol{f_Y}$ is the softmax, thus $f_{yl} = \frac{1}{\boldsymbol{Z_Y}}\text{exp}(g_{yl})$ where $g_{yl} = \sum_{k=1}^K\omega^{(3)}_{kl} \cdot f_{2k} + b_{yl}$ and $\boldsymbol{Z_Y} = \sum_{l=1}^L\text{exp}(g_{yl})$ is the partition function.
}
\label{fig_mlp_bhm}
\end{figure}

\textbf{Preliminaries.} $P(\boldsymbol{X}, {Y}) = P(Y|\boldsymbol{X})P(\boldsymbol{X})$ is an unknown joint distribution between two random variables $\boldsymbol{X}$ and ${Y}$. 
A dataset $\boldsymbol{\mathcal{D}} = \{(\boldsymbol{x}^j, {y}^j)| \boldsymbol{x}^j \in {\mathbb{R}}^{M}, {y}^j \in {\mathbb{R}}\}_{j=1}^{J}$ consists of $J$ \textit{i.i.d.}\ samples generated from $P(\boldsymbol{X}, {Y})$ with finite $L$ classes, i.e., $y^j \in \{1, \cdots, L\}$.
A neural network with $I$ hidden layers is denoted as $\text{DNN} = \{\boldsymbol{x; f_1; ...; f_I; f_Y}\}$ and trained by $\boldsymbol{\mathcal{D}}$, where $(\boldsymbol{x}^j,{y}^j) \in \boldsymbol{\mathcal{D}}$ are the input of the DNN and the label, respectively, thus $\boldsymbol{x} \sim P(\boldsymbol{X})$ and the DNN aims to learn $P(Y|\boldsymbol{X})$ with the one-hot format, i.e., if $l = y^j$, $P_{Y|\boldsymbol{X}}(l|\boldsymbol{x}^j) = 1$; otherwise $P_{Y|\boldsymbol{X}}(l|\boldsymbol{x}^j) = 0$.
We use the $\text{MLP} = \{\boldsymbol{x; f_1; f_2; f_Y}\}$ in Figure \ref{fig_mlp_bhm} for most theoretical derivations unless otherwise specified.
In addition, $H(\boldsymbol{X})$ is the entropy of $\boldsymbol{X}$, 
$I(\boldsymbol{X}, F_i)$ and $I(Y, F_i)$ are the mutual information between $\boldsymbol{X}$, $Y$ and ${F_i}$, where ${F_i}$ is the random variable of the hidden layer $\boldsymbol{f_i}$.

\section{Related work} 
\label{related works}

\subsection{Information theoretic explanations for DNNs}

IB aims to optimize a random variable ${F}$ as a compressed representation of $\boldsymbol{X}$ such that it can minimize the information of $\boldsymbol{X}$ while preserve the information of ${Y}$ \citep{slonim2002information}.
Since ${F}$ is a compressed representation of $\boldsymbol{X}$, it is entirely determined by $\boldsymbol{X}$, i.e, $P({F}|\boldsymbol{X}, {Y}) = P(\boldsymbol{F}| \boldsymbol{X})$, thus the joint distribution $P(\boldsymbol{X}, {Y}, {F})$ can be formulated as
\begin{equation}
\begin{split}
P(\boldsymbol{X}, {Y}, {F}) &= P(\boldsymbol{X}, {Y})P({F}| \boldsymbol{X}) \\ 
&= P({Y})P(\boldsymbol{X}|{Y})P({F}| \boldsymbol{X}).
\end{split}
\end{equation}
As a result, the corresponding Markov chain can be described as ${Y} \boldsymbol{\leftrightarrow} \boldsymbol{X} \boldsymbol{\leftrightarrow} {F}$ and IB can be formulated as 
\begin{equation}
P^{*}({F}|\boldsymbol{X}) = \argmin_{P({F}|\boldsymbol{X})} I(\boldsymbol{X}, F) - \beta I(Y, F),
\end{equation}
where $\beta$ is the Lagrange multiplier controlling the tradeoff between $I(\boldsymbol{X}, F)$ and $I(Y, F)$.

The key to validating the IB explanation for MLPs is to precisely measure $I(\boldsymbol{X}, F_i)$ and $I(Y, F_i)$.
Ideally, we should specify $F_i: \Omega_{F_i} \rightarrow E_{F_i}$ before deriving $I({F_i}, \boldsymbol{X})$ and $I({F_i}, {Y})$. 
However, the complicated architecture of MLPs makes it hard to specify $F_i$.
Alternatively, most previous works use non-parametric inference to estimate $I(\boldsymbol{X}, F_i)$ and $I(Y, F_i)$.

Based on a classical non-parametric inference method, namely the empirical distribution of the activations of $\boldsymbol{f}_i$, \citet{DNN-information} experimentally show that the $\text{MLP} = \{\boldsymbol{x; f_1; f_2; f_Y}\}$ shown in Figure \ref{fig_mlp_bhm} satisfies IB and corresponds to a Markov chain 
\begin{equation}
\label{markov_chain_dnn}
{Y} {\leftrightarrow} \boldsymbol{X} {\leftrightarrow} {F_1} {\leftrightarrow} {F_2} \boldsymbol{\leftrightarrow} {F_Y}.
\end{equation}
As a result, the information flow in the MLP should satisfies the two Data Processing Inequalities (DPIs)
\begin{equation}
\label{dpi_dnn}
\begin{split}
H(\boldsymbol{X}) \geq I(\boldsymbol{X}, {F_1}) &\geq I(\boldsymbol{X}, {F_2}) \geq I(\boldsymbol{X}, {F_Y}), \\
I({Y}, \boldsymbol{X}) \geq I({Y}, {F_1}) &\geq I({Y}, {F_2}) \geq I({Y}, {F_Y}).
\end{split}
\end{equation}
Furthermore, they claim that most of training epochs focus on learning a compressed representation of input for fitting the labels, and the generalization performance of DNNs is causally related to the compression phase.

Meanwhile, other non-parametric inference methods are also used to estimate the mutual information. 
For instance, \citet{IP-argue} use Gaussian KDE to estimate $I(\boldsymbol{X}, F_i)$ and $I(Y, F_i)$, \citet{goldfeld2019estimating} choose the convolution of the empirical distribution and additive Gaussian noise to estimate $I(\boldsymbol{X}, F_i)$, and \citet{chelombiev2018adaptive} propose some adaptive techniques for optimizing the mutual information estimators based on empirical distributions and KDE.

However, the information theoretic explanations for DNNs based on non-parametric inference have several limitations.
First, it is invalid for non-saturating activation functions, e.g., the widely used ReLU. 
Second, the causal relation between generalization and compression cannot be validated by KDE and other recent works \citep{gabrie2018entropy}.

Except the classical non-parametric inference methods, recent works propose some new mutual information estimators.
For instance, \citet{yu2019multivariate} propose the matrix-based R\'{e}nyi $\alpha$-entropy to estimate $I(\boldsymbol{X}, F_i)$  without probabilistic modeling $\boldsymbol{f}_i$, in which Shannon entropy is a special case of R\'{e}nyi $\alpha$-entropy when $\alpha \rightarrow 1$ \citep{yu2019multivariate, yu2020understanding}.
\citet{gabrie2018entropy} propose the heuristic replica method to estimate $I(\boldsymbol{X}, F_i)$ in statistical feedforward
neural networks \citep{kabashima2008inference, manoel2017multi}.


\subsection{Probabilistic explanations for DNNs}

Probabilistic modeling the hidden layer of DNNs is a fundamental question of deep learning theory.
Numerous probabilistic models have been proposed to explain DNNs, e.g., Gaussian process \citep{DNN_GP, Mattews_GP, CNN_GP2}, mixture model \citep{DRMM, DMFA, DGMM}, and Gibbs distribution \citep{Boltzmann_machine, yaida2019non}.
 
As a fundamental probabilistic graphic model, the Gibbs distribution (a.k.a., Boltzmann distribution, the energy based model, or the renormalization group) formulates the dependence within $\boldsymbol{X}$ by associating an energy $E(\boldsymbol{x; \theta})$ to each dependence structure \citep{Geman}.
\begin{equation} 
\label{Gibbs} 
P(\boldsymbol{X}; \boldsymbol{\theta}) = \frac {1}{Z(\boldsymbol{\theta})}\text{exp}[-E(\boldsymbol{x; \theta})],
\end{equation}
where $E(\boldsymbol{x; \theta})$ is the energy function, $\boldsymbol{\theta}$ are the parameters, and the partition function is $Z(\boldsymbol{\theta}) = \sum_{\boldsymbol{x}}{\text{exp}[-E(\boldsymbol{x; \theta})]}$\footnote{We only consider the discrete case in  the paper.}. 

The Gibbs distribution has three appealing properties.
First, it can be easily reformulated as various probabilistic models by redefining $E(\boldsymbol{x; \theta})$, which allows us to clarify the complicated architecture of a hidden layer. 
For example, if the energy function is defined as  the summation of multiple functions, namely ${E(\boldsymbol{x; \theta}) = -\sum_{k}f_k(\boldsymbol{x}; \boldsymbol{\theta}_k)}$, the Gibbs distribution would be the Product of Experts (PoE) model, i.e., ${ P(\boldsymbol{x}; \boldsymbol{\theta}) =  \frac {1}{Z(\boldsymbol{\theta})}\prod_{k}{F}_k}$, where ${F}_k = \text{exp}[-f_k(\boldsymbol{x}; \boldsymbol{\theta}_k)]$ and ${ Z(\boldsymbol{\theta}) = \prod_k Z(\boldsymbol{\theta}_k)}$ \citep{CD}. 
Second, since $Z(\boldsymbol{\theta})$ only depends on $\boldsymbol{\theta}$, the deterministic function $E(\boldsymbol{x; \theta})$ is a sufficient statistics of $P(\boldsymbol{X}; \boldsymbol{\theta})$.
The property allows us to explain a deterministic hidden layer in a probabilistic way.
Third, the energy minimization is a typical optimization for $\boldsymbol{\theta}$, namely $\boldsymbol{\theta^*} = \argmin_{\boldsymbol{\theta}}E(\boldsymbol{x; \theta})$ \citep{energy_learning}, which allows us to explain the back-propagation training, because the energy minimization can be implemented by the gradient descent algorithm as long as $E(\boldsymbol{x}; \boldsymbol{\theta})$ is differentiable.

To the best of our knowledge, \cite{Boltzmann_machine} initially explain the distribution of hidden layers as a Gibbs distribution in the Restricted Boltzmann Machine (RBM).
\cite{lin2017does} clarify certain advantages of DNNs based on the Gibbs distribution. 
Notably, \cite{yaida2019non} indirectly demonstrates the distribution of a fully connected layer as a Gibbs distribution.
However, there is few work to extend the Gibbs explanation to complicated hidden layers, e.g., fully connected layers and convolutional layers. 

\begin{figure*}
\centering
\includegraphics[scale=0.55]{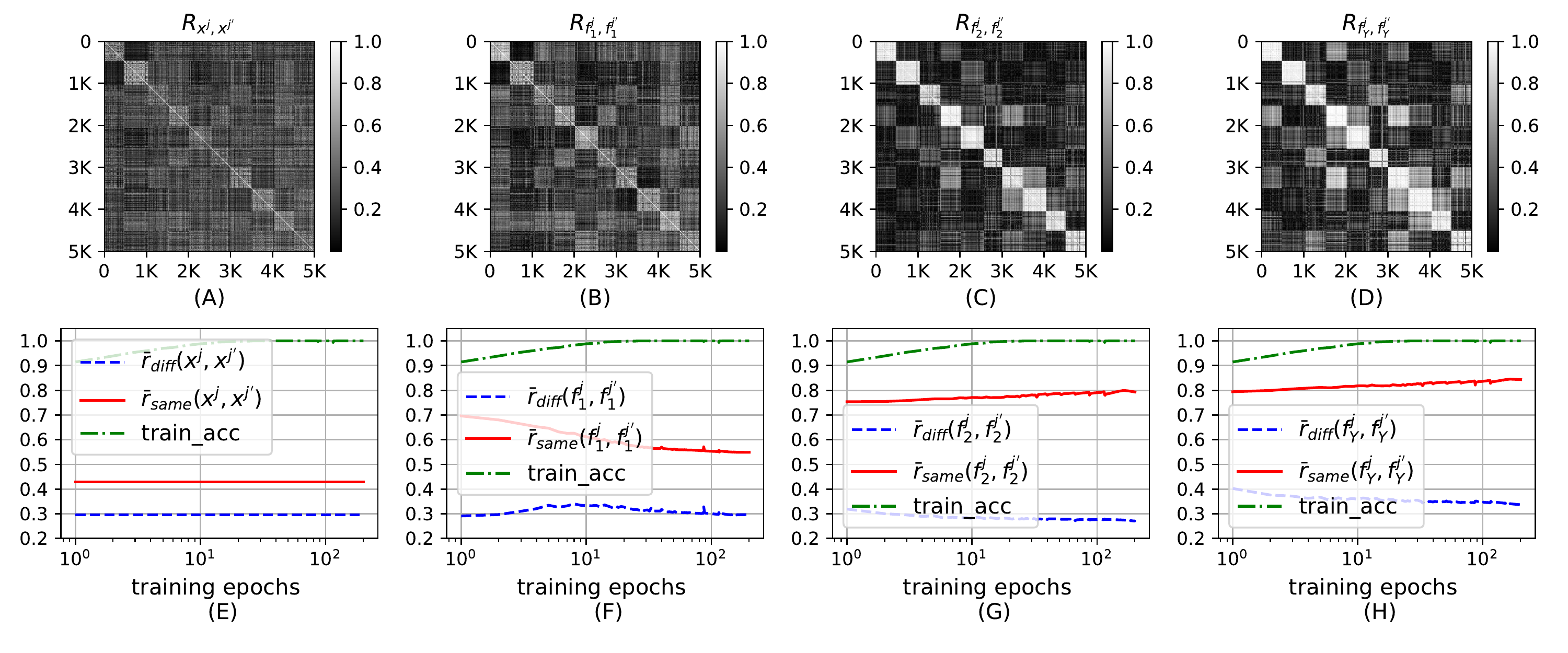}
\caption{\small{
(A) visualizes the sample correlation matrix $R_{\boldsymbol{x}^j, \boldsymbol{x}^{j'}}$ given the 5000 testing dataset $\{\boldsymbol{x}^j\}_{j=1}^{5000}$.
(B)-(D) visualize the three sample correlation matrix $R_{\boldsymbol{f}_i^j, \boldsymbol{f}_i^{j'}}$ for the three layers given $\{\boldsymbol{x}^j\}_{j=1}^{5000}$, respectively.
(E) visualizes the average sample correlation of $\{\boldsymbol{x}^j\}_{j=1}^{5000}$ with the same labels and with different labels.
(F)-(H) visualize the average sample correlation of $\{\boldsymbol{f}_i^j\}_{j=1}^{5000}$ for the three layers with the same labels and with different labels. 
}}
\label{fig_mlp_iid1}
\end{figure*}

\section{Novel probabilistic explanations} 
\label{prob_exp}

In this section, we present three theoretical results:  (i) we demonstrate that activations being \textit{i.i.d.}\ is not valid for all the layers of MLPs, thus non-parametric inference cannot model the distributions of all the fully connected layers of MLPs; (ii) we define the probability space $(\Omega_F, \mathcal{T}, P_F)$ for a fully connected layer, and propose a probabilistic explanation for the entire architecture of MLPs based on the Gibbs measure $P_F$; and (iii) we introduce a probabilistic explanation for the back-propagation training based on the sample space $\Omega_F$.

\subsection{Activations are not \textit{i.i.d.}}

Given an input $\boldsymbol{x}^j \in {\mathbb{R}}^{M}$, we define the corresponding multivariate random variable ${\textstyle \boldsymbol{X}^j = [X^j_1, \cdots, X^j_M]}$, where $X^j_m$ is the scalar-valued random variable of $x^j_m$.
In the context of frequentist probability, all the parameters of MLPs are viewed as constants, thus the random variable of $g_{1n}(\boldsymbol{x}^j) = \sum_{m=1}^M\omega^{(1)}_{mn} \cdot x_m^j + b_{1n}$ is defined as ${\textstyle G^j_{1n} = \sum_{m=1}^M \omega^{(1)}_{mn}X^j_{m} + b_{1n}}$ and the random variable of the activation $f^j_{1n} = \sigma_1(g^j_{1n})$ is defined as $F^j_{1n} = \sigma_1(G^j_{1n})$. 
Therefore, the multivariate random variable of $\boldsymbol{f}^j_1 = [f^j_{11}, \cdots, f^j_{1N}]$ can be defined as ${\textstyle \boldsymbol{F}^j_1 = [F^j_{11}, \cdots, F^j_{1N}]}$.
Similarly, we define the multivariate random variable of ${\textstyle \boldsymbol{f}^j_2}$ as ${\textstyle \boldsymbol{F}^j_2 = [F^j_{21}, \cdots, F^j_{2K}]}$ and the multivariate random variable of $\boldsymbol{f}^j_Y$ as ${\textstyle \boldsymbol{F}^j_Y = [F^j_{y1}, \cdots, F^j_{yL}]}$.

Samples being \textit{i.i.d.} is the prerequisite of non-parametric inference methods, e.g., the empirical distribution and KDE.
In the context of MLPs, all the previous works regard the activations of a layer as the samples of the random variable of the layer.
As a result, activations being \textit{i.i.d.} should be the prerequisite of applying non-parametric inference methods to estimate the true distribution of the layer.

Since the necessary condition for samples being \textit{i.i.d.}\ is uncorrelation, we can use the sample correlation to examine if activations being \textit{i.i.d.}.
More specifically, if $\boldsymbol{F}^j_i$ and $\boldsymbol{F}^{j'}_i$ are \textit{i.i.d.}, the sample correlation $r_{\boldsymbol{f}^j_i, \boldsymbol{f}^{j'}_i}$ must be zero, 
\begin{equation}
{
r_{\boldsymbol{f}^j_i, \boldsymbol{f}^{j'}_i} = \frac{\sum_{n=1}^N ({f}^j_{in} - \bar{\boldsymbol{f}^j_i})({f}^{j'}_{in} - \bar{\boldsymbol{f}^{j'}_i})}{\sqrt{{\sum_{n=1}^N ({f}^j_{in} - \bar{\boldsymbol{f}^j_i})}^2 \sum_{n=1}^N {({f}^{j'}_{in} - \bar{\boldsymbol{f}^{j'}_i})}^2}},
}
\end{equation}
where $\boldsymbol{f}^j_i$ and $\boldsymbol{f}^{j'}_i$ are two activation samples of $\boldsymbol{F}^j_i$ and $\boldsymbol{F}^{j'}_i$ given two sample inputs $\boldsymbol{x}^j$ and $\boldsymbol{x}^{j'}$, $\bar{\boldsymbol{f}^j_i} = \frac{1}{N}\sum_{n=1}^N {f}^j_{in}$, and $N$ is the number neurons of $\boldsymbol{f}_i$.

We specify the MLP to classify the benchmark MNIST dataset \citep{lecun_cnns}.
Since the dimension of each image is $28 \times 28$, the number of the input nodes is $M = 784$.
In addition, $\boldsymbol{f_1}$, $\boldsymbol{f_2}$, and $\boldsymbol{f_Y}$ have $N = 128$, $K= 64$, and $L = 10$ neurons/nodes, respectively.
All the activation functions are sigmoid.
After 200 training epochs, we derive ${\textstyle r_{\boldsymbol{f}^j_i, \boldsymbol{f}^{j'}_i}}$ on the 5000 testing images $\{\boldsymbol{x}^j\}_{j=1}^{5000}$ and define the matrix $R_{\boldsymbol{f}^j_i, \boldsymbol{f}^{j'}_i}$ to contain all the ${\textstyle |r_{\boldsymbol{f}^j_i, \boldsymbol{f}^{j'}_i}|}$.

As a result, we can examine if $\boldsymbol{F}^j_i$ and $\boldsymbol{F}^{j'}_i$ being \textit{i.i.d.}\ by checking if most elements in $R_{\boldsymbol{f}^j_i, \boldsymbol{f}^{j'}_i}$ are close to zero.
In addition, we rearrange the order of $\{\boldsymbol{x}^j\}_{j=1}^{5000}$ such that images with the same label have consecutive index, i.e., images with label $l$ has the index $[l\times 500, (l+1)\times500]$, thus we can easily check the correlation of activations with the same label.

We demonstrate that the correlation of activations with the same label becomes larger as the layer is closer to the output.
In other words, activations being \textit{i.i.d.}\ is not valid for all the layers of the MLP, thus non-parametric inference cannot correctly model the true distribution of all the layers.

More specifically, Figure \ref{fig_mlp_iid1}(A) shows that the correlation of each pair of testing images $\{\boldsymbol{x}^j\}_{j=1}^{5000}$, i.e., $r_{\boldsymbol{x}^j, \boldsymbol{x}^{j'}}$, is close to zero. 
Figure \ref{fig_mlp_iid1}(E) shows $\bar{r}_{\text{diff}}(\boldsymbol{x}^j, \boldsymbol{x}^{j'})$ and $\bar{r}_{\text{same}}(\boldsymbol{x}^j, \boldsymbol{x}^{j'})$, which denote the average sample correlation of $\{\boldsymbol{x}^j\}_{j=1}^{5000}$ with different labels and with the same label, respectively.
\begin{equation}
{
\bar{r}_{\text{diff}}(\boldsymbol{x}^j, \boldsymbol{x}^{j'}) = \frac{1}{N_{\text{diff}}} \sum_{l=0}^{L-1}\sum_{y^j \neq y^{j'}}r_{\boldsymbol{x}^j, \boldsymbol{x}^{j'}}
}
\end{equation}
\begin{equation}
{
\bar{r}_{\text{same}}(\boldsymbol{x}^j, \boldsymbol{x}^{j'}) = \frac{1}{N_{\text{same}}} \sum_{l=0}^{L-1}\sum_{y^j = y^{j'} = l}r_{\boldsymbol{x}^j, \boldsymbol{x}^{j'}}
}
\end{equation}
where $N_{\text{diff}}$ and $N_{\text{same}}$ are the total number of pairs $(\boldsymbol{x}^j, \boldsymbol{x}^{j'})$ with different labels and the same label, respectively.
We observe that $r_{\text{diff}}(\boldsymbol{x}^j, \boldsymbol{x}^{j'})$ is around 0.29 and $r_{\text{same}}(\boldsymbol{x}^j, \boldsymbol{x}^{j'})$ is around 0.43 in Figure \ref{fig_mlp_iid1}(E).
In summary, the correlation coefficients of $\{\boldsymbol{x}^j\}_{j=1}^{5000}$ are low, thus \textit{i.i.d.} can be viewed as a valid assumption for $\{\boldsymbol{x}^j\}_{j=1}^{5000}$.

In terms of the correlation of activations with the same label in different layers, Figure \ref{fig_mlp_iid1}(B)-(D) show an ascending trend as the layer is closer to the output.
For instance, the pixels at the top-left corner of $R_{\boldsymbol{f}^j_i, \boldsymbol{f}^{j'}_i}$ becomes lighter as the layer is closer to the output, i.e., the correlation of the activations with the label 0 becomes larger.
In addition, Figure \ref{fig_mlp_iid1}(F)-(J) also demonstrate the ascending trend, i.e., ${\textstyle \bar{r}_{\text{same}}(\boldsymbol{f}^j_1, \boldsymbol{f}^{j'}_1)}$ converges to 0.55, $\bar{r}_{\text{same}}(\boldsymbol{f}^j_2, \boldsymbol{f}^{j'}_2)$ converges to 0.79, and $\bar{r}_{\text{same}}(\boldsymbol{f}^j_Y, \boldsymbol{f}^{j'}_Y)$ converges to 0.84.

As a comparison, Figure \ref{fig_mlp_iid1}(B)-(D) show the correlation of activations with different labels being relatively stable in different layers, which is further validated by Figure \ref{fig_mlp_iid1}(F)-(J) showing that $\bar{r}_{\text{diff}}(\boldsymbol{f}^j_1, \boldsymbol{f}^{j'}_1)$, $\bar{r}_{\text{diff}}(\boldsymbol{f}^j_2, \boldsymbol{f}^{j'}_2)$, and $\bar{r}_{\text{diff}}(\boldsymbol{f}^j_Y, \boldsymbol{f}^{j'}_Y)$ converge to 0.29,  0.27, and 0.33, respectively.


In summary, the correlation of activations with the same label becomes larger as the layer is closer to the output, thus activations being \textit{i.i.d.}\ is not valid for all the layers of the MLP.
In addition, we derive the same result based on more complicated MLPs on the benchmark Fashion-MINST dataset in Appendix \ref{MLP_IID}.
As a result, non-parametric inference, e.g., the empirical distribution and KDE, cannot correctly model the true distribution of all the layers, thus they are invalid for estimating the mutual information between each layer and the input/labels.
Notably, this section further confirms the necessity for establishing a slid probabilistic foundation for deriving information theoretic explanations for DNNs.

\subsection{Probabilistic explanations for MLPs}

This section proposes three probabilistic explanations for MLPs: (i) we define the probability space $(\Omega_F, \mathcal{T}, P_F)$ for a fully connected layer, (ii) we prove the entire architecture of MLPs as a Gibbs distribution $P$, and (iii) we demonstrate that the back-propagation aims to optimize the sample space of each layer to learn an optimal Gibbs distribution $P^*$ for describing the statistical connection between $\boldsymbol{X}$ and $Y$.

\subsubsection{The probability space $(\Omega_F, \mathcal{T}, P_F)$ for a layer}
\label{prob_space}
In this section, we define the probability space $(\Omega_F, \mathcal{T}, P_F)$ for a fully connected layer, and prove the probability space being valid for all the fully connected layers of MLPs.

\textbf{Definition.} Given a fully connected layer $\boldsymbol{f}$ consisting of $N$ neurons $\{f_n = \sigma[g_{n}(\boldsymbol{x})]\}_{n=1}^N$, where $\sigma(\cdot)$ is an activation function, e.g, the sigmoid function, $\boldsymbol{x} = \{x_m\}_{m=1}^M \in \mathbb{R}^M$ is the input of $\boldsymbol{f}$, $g_{n}(\boldsymbol{x})=\sum_{m=1}^M\omega_{mn}\cdot x_m + b_n$ is the $n$th linear filter with $\omega_{mn}$ being the weight and $b_n$ being the bias, let $\boldsymbol{f}$ extracting a single feature of $\boldsymbol{x}$ be an experiment,
we define the probability space $(\Omega_F, \mathcal{T}, P_F)$ for $\boldsymbol{f}$ as follows.

First, the sample space $\Omega_F$ includes $N$ possible outcomes $\{\boldsymbol{\omega}_n\}_{n=1}^N = \{\{{\omega}_{mn}\}_{m=1}^{M}\}_{n=1}^N$ defined by the weights of the $N$ neurons. 
Since a scalar value cannot describe the feature of $\boldsymbol{x}$, we do not take into account $b_n$ for defining $\Omega_F$.
In terms of machine learning, $\boldsymbol{\omega}_n$ defines a possible feature of $\boldsymbol{x}$.
In particular, the definition of the experiment guarantees that the possible outcomes are mutually exclusive (i.e., only one outcome will occur on each trial of the experiment). 

Second, we define the event space $\mathcal{T}$ as the $\sigma$-algebra.
For example, if $\boldsymbol{f}$ has $N = 2$ neurons and $\Omega = \{\boldsymbol{\omega}_1, \boldsymbol{\omega}_2\}$, ${\textstyle \mathcal{T} = \{\emptyset, \{\boldsymbol{\omega}_1\}, \{\boldsymbol{\omega}_2\}, \{\{\boldsymbol{\omega}_1, \boldsymbol{\omega}_2\}\}}$ means that neither of the outcomes, one of the outcomes, or both of the outcomes could happen, respectively.

Third, the probability measure $P_F$ is the Gibbs measure to quantify the probability of $\{\boldsymbol{\omega}_n\}_{n=1}^N$ occurring in $\boldsymbol{x}$.
\begin{equation} 
\label{Gibbs_f}
\begin{split}
P_F(\boldsymbol{\omega}_n) &= \frac{1}{Z_{F}}\text{exp}(f_{n}) = \frac{1}{Z_{F}}\text{exp}[\sigma(g_{n}(\boldsymbol{x}))] \\
&= \frac{1}{Z_{F}}\text{exp}[\sigma(\langle \boldsymbol{\omega}_n, \boldsymbol{x} \rangle + b_n)]\\
\end{split}
\end{equation}
where $\langle \cdot, \cdot \rangle$ denotes the inner product and $Z_{F} = \sum_{n=1}^N\text{exp}(f_{n})$ is the partition function.

\textbf{Proof.} We use the mathematical induction to prove the probability space for all the fully connected layers of the MLP in the backward direction.
Given three probability space $(\Omega_{F_1}, \mathcal{T}, P_{F_1})$, $(\Omega_{F_2}, \mathcal{T}, P_{F_2})$, and $(\Omega_{F_Y}, \mathcal{T}, P_{F_Y})$ for the three layers $\boldsymbol{f_1}$, $\boldsymbol{f_2}$, and $\boldsymbol{f_Y}$, respectively, we first prove $P_{{F_Y}}$ as a Gibbs distribution, and then we prove $P_{{F_2}}$ and $P_{{F_1}}$ being Gibbs distributions based on $P_{{F_Y}}$ and $P_{{F_2}}$, respectively.

Since the output layer $\boldsymbol{f_Y}$ is the softmax, each output node $f_{yl}$ can be formulated as 
\begin{equation}
\label{Gibbs_fY}
\begin{split}
f_{yl} &= \frac{1}{Z_{F_Y}}\text{exp}(g_{yl})= \frac{1}{Z_{F_Y}}\text{exp}[\sum_{k=1}^K\omega^{(3)}_{kl} \cdot f_{2k} + b_{yl}]\\
&= \frac{1}{Z_{F_Y}}\text{exp}[\langle \boldsymbol{\omega}^{(3)}_l, \boldsymbol{f_2} \rangle + b_{yl}],
\end{split}
\end{equation}
where $Z_{F_Y} = \sum_{l=1}^L\text{exp}(g_{yl})$ is the partition function and $\boldsymbol{\omega}^{(3)}_{l} = \{\omega^{(3)}_{kl}\}_{k=1}^K$.
Comparing Equation \ref{Gibbs_f} and \ref{Gibbs_fY},  we can derive that $\boldsymbol{f_Y}$ forms a Gibbs distribution $P_{F_Y}(\boldsymbol{\omega}^{(3)}_l) = f_{yl}$ to measure the probability of $\boldsymbol{\omega}^{(3)}_l$ occurring in $\boldsymbol{f_2}$, which is consistent with the definition of $(\Omega_{F_Y}, \mathcal{T}, P_{F_Y})$.

\pagebreak
Based on the properties of exponential functions, i.e., ${\textstyle \text{exp}(a+b) = \text{exp}(a)\cdot \text{exp}(b)}$ and ${\textstyle \text{exp}(a\cdot b) = [\text{exp}(b)]^{a}}$, we can reformulate $P_{\boldsymbol{F_Y}}(\boldsymbol{\omega}^{(3)}_l)$ as 
\begin{equation} 
P_{{F_Y}}(\boldsymbol{\omega}^{(3)}_l) = \frac {1}{Z'_{F_Y}}\prod_{k=1}^K[\text{exp}(f_{2k})]^{\omega^{(3)}_{kl}},
\end{equation}
where $Z'_{F_Y} = Z_{F_Y}/\text{exp}(b_{yl})$. 
Since $\{\text{exp}(f_{2k})\}_{k=1}^K$ are scalar, we can introduce a new partition function ${\textstyle Z_{{F_2}} = \sum_{k=1}^K\text{exp}(f_{2k})}$ such that $\{\frac{1}{Z_{{F_2}}}\text{exp}(f_{2k})\}_{k=1}^K$ becomes a probability measure, thus we can reformulate $P_{{F_Y}}(\boldsymbol{\omega}^{(3)}_l) $ as a PoE model
\begin{equation} 
\label{FoE_fY}
P_{{F_Y}}(\boldsymbol{\omega}^{(3)}_l) =\frac {1}{Z''_{F_Y}}\prod_{k=1}^K[\frac{1}{Z_{{F_2}}}\text{exp}(f_{2k})]^{\omega^{(3)}_{kl}},
\end{equation}
where ${\textstyle Z''_{F_Y} = Z_{F_Y}/[\text{exp}(b_{yl}) \cdot \prod_{k=1}^K[Z_{{F_2}}]^{\omega^{(3)}_{kl}}}]$, especially each expert is defined as $\frac{1}{Z_{{F_2}}}\text{exp}(f_{2k})$.

It is noteworthy that all the experts $\{\frac{1}{Z_{{F_2}}}\text{exp}(f_{2k})\}_{k=1}^K$ form a probability measure and establish an exact one-to-one correspondence to all the neurons in $\boldsymbol{f_2}$, thus the distribution of $\boldsymbol{f_2}$ can be expressed as
\begin{equation}
\label{Gibbs_f2}
P_{{F_2}} = \{\frac {1}{Z_{{F_2}}}\text{exp}(f_{2k})\}_{k=1}^K.
\end{equation}

Since $\{f_{2k} = \sigma_2(\sum_{n=1}^N\omega^{(2)}_{nk} \cdot f_{1n} + b_{2k})\}_{k=1}^K$, $P_{{F_2}}$ can be extended as 
\begin{equation}
\begin{split}
P_{{F_2}}(\boldsymbol{\omega}^{(2)}_{k}) &= \frac {1}{Z_{{F_2}}}\text{exp}[\sigma_2(\sum_{n=1}^N\omega^{(2)}_{nk} \cdot f_{1n} + b_{2k})]\\
&= \frac {1}{Z_{{F_2}}}\text{exp}[\sigma_2(\langle \boldsymbol{\omega}^{(2)}_k, \boldsymbol{f_1} \rangle + b_{2k})].
\end{split}
\end{equation}
where $Z_{{F_2}} = \sum_{k=1}^K\text{exp}(f_{2k})$ is the partition function and $\boldsymbol{\omega}^{(2)}_{k} = \{\omega^{(2)}_{nk}\}_{n=1}^N$.
We can conclude that $\boldsymbol{f_2}$ corresponds to a Gibbs distribution $P_{{F_2}}(\boldsymbol{\omega}^{(2)}_k)$ to measure the probability of $\boldsymbol{\omega}^{(2)}_k$ occurring in $\boldsymbol{f_1}$, which is consistent with $(\Omega_{F_2}, \mathcal{T}, P_{\boldsymbol{F_2}})$.

Due to the non-linearity of the activation function $\sigma_2(\cdot)$, we cannot derive $P_{{F_2}}(\boldsymbol{\omega}^{(2)}_{k})$ being a PoE model only based on the properties of exponential functions.
Alternatively, the equivalence between the gradient descent algorithm and the first order approximation \citep{GD_Taylor} indicates that $\sigma_2(\sum_{n=1}^N\omega^{(2)}_{nk} \cdot f_{1n} + b_{2k})]$ can be approximated as 
\begin{equation} 
\sigma_2(\sum_{n=1}^N\omega^{(2)}_{nk} \cdot f_{1n} + b_{2k})] \approx C_{21} \cdot [\sum_{n=1}^N\omega^{(2)}_{nk} \cdot f_{1n} + b_{2k}]  + C_{22},
\end{equation}
where $C_{21}$ and $C_{22}$ only depend on the activations $\{f_{1n}\}_{n=1}^N$ in the previous training iteration, thus they can be regarded as constants and absorbed by $\omega^{(2)}_{nk}$ and $b_{2k}$.
The proof for the approximation is included in Appendix \ref{GD_Approx}.

Therefore, $P_{F_2}$ still can be modeled as a PoE model
\begin{equation} 
\label{FoE_f2}
P_{{F_2}} = \{\frac {1}{Z''_{\boldsymbol{F_2}}}\prod_{n=1}^N[\frac{1}{Z_{\boldsymbol{F_1}}}\text{exp}(f_{1n})]^{\omega^{(2)}_{nk}}\}_{k=1}^K,
\end{equation}
where ${Z''_{{F_2}} = Z_{{F_2}}/[\text{exp}(b_{2k})\cdot \prod_{n=1}^NZ_{{F_1}}^{\omega^{(2)}_{nk}}}]$ and the partition function is ${ Z_{{F_1}} = \sum_{n=1}^N\text{exp}(f_{1n})}$.
Similar to $P_{{F_2}}(\boldsymbol{\omega}^{(2)}_{k})$, we can derive the probability measure of $\boldsymbol{f_1}$ as
\begin{equation} 
\label{Gibbs_f1}
\begin{split}
P_{{F_1}}(\boldsymbol{\omega}^{(1)}_{n}) &= \frac{1}{Z_{{F_1}}}\text{exp}(f_{1n})\\
&= \frac{1}{Z_{{F_1}}}\text{exp}[\sigma_1(\sum_{m=1}^M \omega^{(1)}_{mn} \cdot x_m + b_{1n})]\\
&= \frac{1}{Z_{{F_1}}}\text{exp}[\sigma_1(\langle \boldsymbol{\omega}^{(1)}_{n}, \boldsymbol{x} \rangle + b_{1n})],\\
\end{split}
\end{equation}
where $Z_{F_1} = \sum_{n=1}^N\text{exp}(f_{1n})$ is the partition function and $\boldsymbol{\omega}^{(1)}_{n} = \{\omega^{(1)}_{mn}\}_{m=1}^M$.
We can conclude that $\boldsymbol{f_1}$ corresponds to a Gibbs distribution $P_{{F_1}}(\boldsymbol{\omega}^{(1)}_n)$ to measure the probability of $\boldsymbol{\omega}^{(1)}_n$ occurring in $\boldsymbol{x}$, which is consistent with $(\Omega_{F_1}, \mathcal{T}, P_{\boldsymbol{F_1}})$.
Overall, we prove the proposed probability space for all the fully connected layers in the MLP. 
Notably, we can easily extend the probability space to an arbitrary fully connected layer through properly changing the script.

Based on $(\Omega_F, \mathcal{T}, P_F)$ for a fully connected layer $\boldsymbol{f}$, we can specify the corresponding random variable $F: \Omega_F \rightarrow E_F$.
More specifically, since $\Omega_F = \{\boldsymbol{\omega}_n\}_{n=1}^N$ includes finite $N$ possible outcomes, $E_F$ is a discrete measurable space and $F$ is a discrete random variable, e.g., $P(F = n)$ denotes the probability of $\boldsymbol{\omega}_n$ occurring in the experiment.

\begin{figure*}
\centering
\includegraphics[scale=0.5]{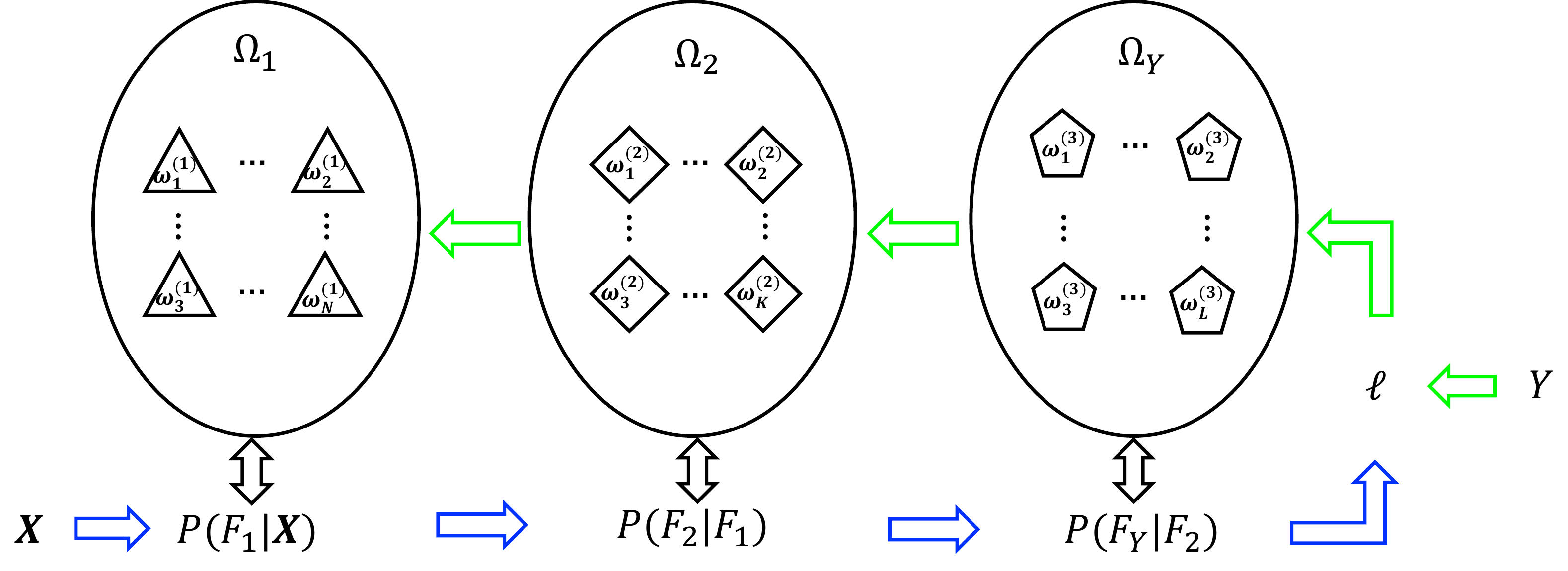}
\caption{\small{
The probabilistic explanation for the $\text{MLP} = \{\boldsymbol{x; f_1; f_2; f_Y}\}$ and the training algorithm. $\ell$ denotes the loss function.
$P({F_i|F_{i-1}})$ is the distribution of the layer $\boldsymbol{f_i}$ given its previous layer.
The oval above $P({F_i})$ represents the corresponding sample space ${\Omega_i}$, which consists of possible outcomes defined by the weights of neurons. 
For example, ${\Omega_1} = \{\boldsymbol{\omega}^{(1)}_{n}\}_{n=1}^N$ and $\boldsymbol{\omega}^{(1)}_{n} = \{\omega_{mn}\}_{m=1}^{M}$. 
}}
\label{Img_mlp_prob_space}
\end{figure*}

\subsubsection{The probabilistic explanation for the entire architecture of the MLP}
\label{probabilistic_mlp}

Since $\Omega_F$ is defined by $\omega_{mn}$, it is fixed if not considering parameters updating.
Therefore, ${F_{i+1}}$ is entirely determined by ${F_{i}}$ in the $\text{MLP} = \{\boldsymbol{x; f_1;f_2; f_Y}\}$ without considering the back-propagation training, and the MLP forms the Markov chain
$\boldsymbol{X} \boldsymbol{\leftrightarrow} {F_1} \boldsymbol{\leftrightarrow} {F_2} {\leftrightarrow} {F_Y}$, thus the entire architecture of the MLP corresponds to a joint distribution
\begin{equation}
P(F_Y, F_2, F_1|\boldsymbol{X}) = P({F_Y|F_2})P({F_2|F_1})P(F_1|\boldsymbol{X}). 
\end{equation}

Subsequently, we can derive the marginal distribution $P(F_Y|\boldsymbol{X})$ still being a Gibbs distribution
\begin{equation} 
\label{pdf_posterior1}
\begin{split}
P_{F_Y|\boldsymbol{X}}(l|\boldsymbol{x}) &= \sum_{k=1}^K\sum_{n=1}^N P(F_Y=l, F_2k, F_1=n|\boldsymbol{X}=\boldsymbol{x}) \\
&= \frac{1}{Z_{\text{MLP}}(\boldsymbol{x})}\text{exp}[f_{yl}(\boldsymbol{f_2}(\boldsymbol{f_1}(\boldsymbol{x})))],
\end{split}
\end{equation}
where $Z_{\text{MLP}}(\boldsymbol{x}) = \sum_{l=1}^L\sum_{k=1}^K\sum_{n=1}^N P_{F_Y, F_2, F_1|\boldsymbol{X}}(l, k, n|\boldsymbol{x})$ is the partition function. 
Notably, the energy function $E_{yl}(\boldsymbol{x}) = -f_{yl}(\boldsymbol{f_2}(\boldsymbol{f_1}(\boldsymbol{x})))$ indicates that $P(F_Y|\boldsymbol{X})$ is determined by the entire architecture of the MLP.
The detailed derivation of $P(F_Y|\boldsymbol{X})$ is presented in Appendix \ref{posterior_MLP}.

\subsubsection{Probabilistic explanations for training}
\label{prob_training}

The back-propagation training \citep{backpropagation} updates the parameters of a hidden layer in the back-forward direction.
In the $\text{MLP} = \{\boldsymbol{x; f_1; f_2; f_Y}\}$, the weights of each layer are updated as 
\begin{equation} 
\boldsymbol{\omega}(T+1) = \boldsymbol{\omega}(T) - \alpha \frac{\partial \ell}{\partial \boldsymbol{\omega}(T)}
\end{equation}
where $\boldsymbol{\omega}(T)$ denotes the learned weights in the $T$th training iteration, $\ell$ is the cross entropy loss function, and $\alpha$ is the learning rate.
Specifically, the gradient of $\ell$ with respect to the weight of each layer in the MLP are formulated as
\begin{equation}
\label{gradient_ell}
\begin{split}
\frac{\partial \ell}{\partial \omega^{(3)}_{kl}} & = [f_{yl} - P_{Y|\boldsymbol{X}}(l|\boldsymbol{x})]f_{2k},\\
\frac{\partial \ell}{\partial \omega^{(2)}_{nk}} &= \sum_{l=1}^L [f_{yl} - P_{Y|\boldsymbol{X}}(l|\boldsymbol{x})]\omega^{(3)}_{kl}\sigma'_2(g_{2k})f_{1n},\\
\frac{\partial \ell}{\partial \omega^{(1)}_{mn}} &= \sum_{k=1}^K \sum_{l=1}^L [f_{yl} - P_{Y|\boldsymbol{X}}(l|\boldsymbol{x})]\omega^{(3)}_{kl}\sigma'_2(g_{2k})\omega^{(2)}_{nk}\sigma'_1(g_{1n})x_m.\\
\end{split}
\end{equation}
where $P_{Y|\boldsymbol{X}}(l|\boldsymbol{x})$ is the conditional probability of the label $y$ given the input $\boldsymbol{x}$, i.e., $P_{Y|\boldsymbol{X}}(l|\boldsymbol{x}) = 1$ if $l = y$, otherwise $P_{Y|\boldsymbol{X}}(l|\boldsymbol{x}) = 0$.
The derivation is presented in Appendix \ref{bp}.

Since weights are randomly initialized before training, i.e., $\boldsymbol{\omega}(0)$ are random values, $\boldsymbol{\omega}(T+1)$ are entirely determined by all the gradients before $T+1$, i.e., $\{\frac{\partial \ell}{\partial \boldsymbol{\omega}(t)}\}_{t=1}^T$.
Therefore, we conclude that $\Omega_F$ is determined by $\{\frac{\partial \ell}{\partial \boldsymbol{\omega}(t)}\}_{t=1}^T$ because $\Omega_F$ is defined by $\boldsymbol{\omega}$.

As a result, we can derive that $\Omega_{F_Y}$ is a function $y$ and $\Omega_{F_i}$ is a function of $\Omega_{F_{i+1}}$. 
First, since $\frac{\partial \ell}{\partial \omega^{(3)}_{kl}}$ is a function of $P_{Y|\boldsymbol{X}}(l|\boldsymbol{x})$, $\Omega_{F_Y}$ can be viewed as a function of $y$ based on the definition of $P_{Y|\boldsymbol{X}}(l|\boldsymbol{x})$.
Second, based on Equation \ref{gradient_ell}, we can reformulate $\frac{\partial \ell}{\partial \omega^{(2)}_{nk}}$ as

\begin{equation}
\label{gradient_32}
\frac{\partial \ell}{\partial \omega^{(2)}_{nk}} = \sum_{l=1}^L \frac{\partial \ell}{\partial \omega^{(3)}_{kl}}\omega^{(3)}_{kl} \frac{\sigma'_2(g_{2k})}{f_{2k}}f_{1n}.
\end{equation}
Equation \ref{gradient_32} indicates $\frac{\partial \ell}{\partial \omega^{(2)}_{nk}}$ is a function of $\frac{\partial \ell}{\partial \omega^{(3)}_{kl}}$, thus $\Omega_{F_2}$ is a function of $\Omega_Y$.
Similarly, we can derive
\begin{equation}
\label{gradient_21}
\frac{\partial \ell}{\partial \omega^{(1)}_{mn}} = \sum_{k=1}^K \frac{\partial \ell}{\partial \omega^{(2)}_{nk}}\omega^{(2)}_{nk}\frac{\sigma'_1(g_{1n})}{f_{1n}}x_m,
\end{equation}
which indicates that $\Omega_{F_1}$ is a function of $\Omega_{F_2}$.

In addition, we can derive $\Omega_{F_1}$, $\Omega_{F_2}$, and $\Omega_{F_Y}$ depending on the input $\boldsymbol{x}$.
Equation (\ref{gradient_ell}) shows that the gradient of $\ell$ with respect to the weight of a layer is a function of the input of the layer, i.e., $\frac{\partial \ell}{\partial \omega^{(1)}_{mn}}$ is a function of $x_m$, $\frac{\partial \ell}{\partial \omega^{(2)}_{nk}}$ is a function of $f_{1n}$, and $\frac{\partial \ell}{\partial \omega^{(3)}_{kl}}$ is a function of $f_{2k}$.
$f_{1n} = \sigma_1[g_{1n}(\boldsymbol{x})]$ and $f_{2k} = \sigma_2[g_{2k}(\boldsymbol{f_1})]$ imply that $\frac{\partial \ell}{\partial \omega^{(1)}_{mn}}$, $\frac{\partial \ell}{\partial \omega^{(2)}_{nk}}$, and $\frac{\partial \ell}{\partial \omega^{(3)}_{kl}}$ depend on the input $\boldsymbol{x}$.
As a result, $\Omega_{F_1}$, $\Omega_{F_2}$, and $\Omega_{F_Y}$ depend on the input $\boldsymbol{x}$, because $\Omega$ is determined by $\frac{\partial \ell}{\partial \omega(t)}$.

In summary, the back-propagation training establishes the relation between the sample space of each layer and the input/labels in two aspects.
First, $\Omega_{F_1}$, $\Omega_{F_2}$, and $\Omega_{F_Y}$ depend on $y$, i.e., $\Omega_{F_Y}$ is a function $y$ and $\Omega_{F_i}$ is a function of $\Omega_{F_{i+1}}$.
Second, $\Omega_{F_1}$, $\Omega_{F_2}$, and $\Omega_{F_Y}$ depend on $\boldsymbol{x}$.

Finally, we visualize the probabilistic explanation for the MLP in Figure \ref{Img_mlp_prob_space}.
The blue arrows indicate that the three layers $\boldsymbol{f}_1$, $\boldsymbol{f}_2$, and $\boldsymbol{f}_Y$ form three conditional distributions, i.e., $P(F_1|\boldsymbol{X})$, $P(F_2|F_1)$, and $P(F_Y|F_2)$, respectively. 
The green arrows indicate that the back-propagation optimize the three sample space $\Omega_1$, $\Omega_2$, and $\Omega_Y$ by updating the weights of each layer in the backward direction.
In addition, the black arrows indicate the mutual effect between the sample space and the corresponding Gibbs distributions. 

\begin{figure}
\centering
\includegraphics[scale=0.5]{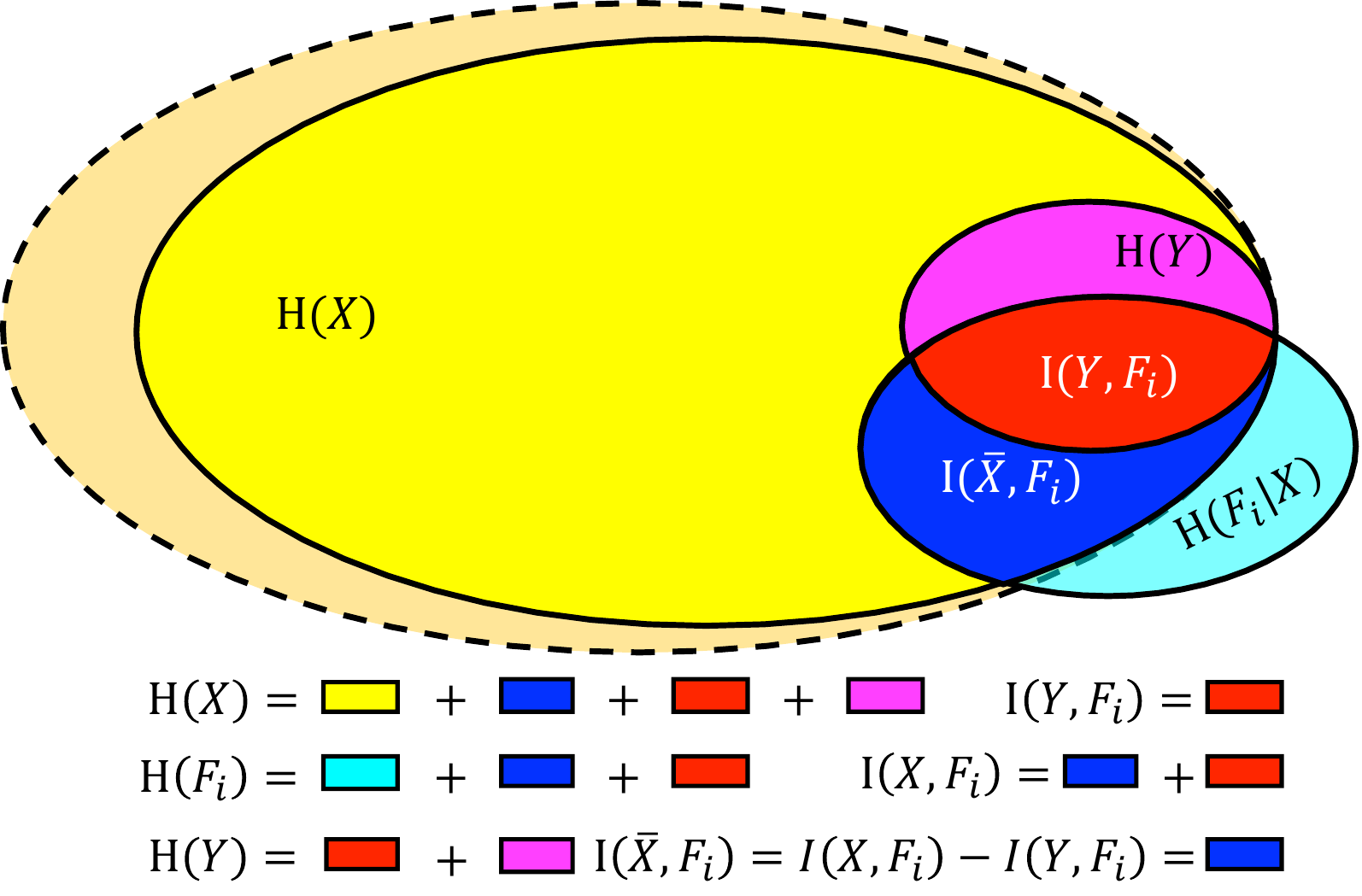}
\caption{\small{
The Venn diagram shows the relationship between the information of $\boldsymbol{X}$, $Y$, and $F_i$.
Since $P(\boldsymbol{X},Y)$ is unknown, the information of $P(\boldsymbol{X},Y)$ is denoted by the largest oval with dashed boundary.
To facilitate subsequent discussions, we still use $H(\boldsymbol{X})$ to denote the information of the \textit{i.i.d.}\ samples $\{\boldsymbol{x}^j\}_{j=1}^J$, because the information of the samples converges to $H(\boldsymbol{X})$ as long as the number of samples is large enough.
}}  
\label{Img_mlp_information_venn}
\end{figure}

\section{Novel information theoretic explanations}
\label{info_exp}

Based on the probabilistic representations for MLPs, we propose five information theoretic explanations for MLPs. 
First of all, the entropy of a fully connected layer is finite. 
Second, we specify the information theoretic relationship between $\boldsymbol{X}$, $Y$ and $F_i$.
Third, IB cannot correctly explain MLPs because MLPs not satisfy the probabilistic premise for IB.
Forth, we specify the information flow of $\boldsymbol{X}$ and ${Y}$ in MLPs.
Fifth, we propose a novel information theoretic explanation for the generalization of MLPs.

\subsection{The entropy of a layer is finite}

A controversy about information theoretic explanations for MLPs is that the random variable ${F_i}: \Omega_{F_i} \rightarrow E_{F_i}$ for a fully connected layer $\boldsymbol{f}_i$ is continuous or discrete \citep{goldfeld2019estimating}. 
All the previous works assume activations as $E_i$, thus ${F_i}$ is continuous and $H({F_i}|\boldsymbol{X}) = -\infty$ under the assumption that MLPs are deterministic models, which contradicts simulation results, i.e., $H({F_i}|\boldsymbol{X}) < \infty$. 

The definition of $(\Omega_{F_i}, \mathcal{T}, P_{F_i})$ resolves the controversy.
Since $\Omega_{F_i}$ is discrete, ${F_i}$ is discrete, thereby $H({F_i}|\boldsymbol{X}) < \infty$.
In particular, the Gibbs measure $P_{F_i}$ regards activations of $\boldsymbol{f}_i$ as the negative energy, i.e., activations are the intermediate variables of $P_{F_i}$, rather than $E_{F_i}$.

\subsection{The relationship between $\boldsymbol{X}$, ${Y}$ and ${F_i}$}
\label{xyfi}
Since we suppose $(\boldsymbol{x}^j, y^j) \in \boldsymbol{\mathcal{D}}$ being \textit{i.i.d.}, we can derive $H(Y) = I(\boldsymbol{X},Y)$ (the proof is presented in Appendix \ref{YbelongstoX}), which indicates that all the information of $Y$ stems from $\boldsymbol{X}$, i.e., the information of $Y$ is a subset of that of $\boldsymbol{X}$ in the Venn diagram of $\boldsymbol{X}$, ${Y}$ and ${F_i}$ (Figure \ref{Img_mlp_information_venn}).

Since the weights of $\boldsymbol{f}_i$ are randomly initialized, $F_i$ does not contain any information of $\boldsymbol{X}$ and $Y$ before training. 
In addition, we cannot guarantee that all the information of $F_i$ is learned from $\boldsymbol{\mathcal{D}}$ after training, thus $H(F_i|\boldsymbol{X}) \neq 0$.

Section \ref{prob_training} demonstrates that $\Omega_{F_i}$ depends on both the input $\boldsymbol{x}$ and the label $y$, thus $I(F_i, \boldsymbol{X}) \neq 0$ and $I(F_i, Y) \neq 0$.
Since all the information of $Y$ stems from $\boldsymbol{X}$, $I(F_i, Y)$ is a subset of $I(F_i, \boldsymbol{X})$, which is shown in Figure (\ref{Img_mlp_information_venn}).

\subsection{The limitation of IB}
\label{limitation_IB}

IB assumes that ${F}$ does not contain any information about ${Y}$ except the information given by $\boldsymbol{X}$, i.e., $P({F}|\boldsymbol{X}, {Y}) = P({F}| \boldsymbol{X})$.
Supposing MLPs satisfy the probabilistic premise, \citet{DNN-information} propose the Markov chain (Equation \ref{markov_chain_dnn}) and two DPIs (Equation \ref{dpi_dnn}) for the MLP. 

However, we demonstrate ${F_i}: \Omega_{F_i} \rightarrow E_{F_i}$ depending on both $\boldsymbol{X}$ and ${Y}$,  because Section \ref{prob_training} show $\Omega_{F_i}$ depending on both $\boldsymbol{x}$ and ${y}$.
As a result, MLPs not satisfy the probabilistic premise for IB if taking into account the back-propagation. 

Notably, the information that $Y$ transfers to $F_i$ during training will retain in $\Omega_{F_i}$ after training, because ${\Omega_{F_i}}$ is fixed after training.
In other words, MLPs still cannot satisfy the probabilistic premise for IB even after training.
Therefore, the information flow of $\boldsymbol{X}$ and $Y$ in MLPs cannot satisfy the DPIs (Equation \ref{dpi_dnn}) derived from IB after training.

\begin{figure}
\centering
\includegraphics[scale=0.65]{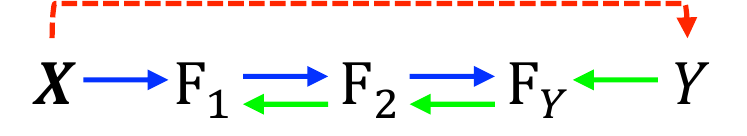}
\caption{\small{
The information flow of $\boldsymbol{X}$ and $Y$ in the MLP. 
}}
\label{Img_mlp_information_flow1}
\end{figure}

\subsection{The information flow of ${Y}$ and $\boldsymbol{X}$}
\label{info_flowyx}

Section \ref{prob_training} shows that $\Omega_{F_Y}$ is a function $y$, $\Omega_{F_2}$ is a function of $\Omega_{F_Y}$, and $\Omega_{F_1}$ is a function of $\Omega_{F_2}$ in the MLP.
Based on the definition of $F_{F_i}: \Omega_{F_i} \rightarrow E_{F_i}$,
we can derive that $F_Y$ is a function $Y$, $F_{2}$ is a function of $F_Y$, and $F_1$ is a function of $F_2$, which indicates the Markov chain $Y \leftrightarrow F_Y \leftrightarrow F_2 \leftrightarrow F_1$. 
As a result, the information flow of $Y$ in the MLP can be expressed as 
\begin{equation}
\label{dpi_y}
I({Y}, {F_1}) \leq I({Y}, {F_2}) \leq I({Y}, {F_Y}).
\end{equation}

Since $\boldsymbol{X}$ is the input of the MLP, the information of $\boldsymbol{X}$ seems to flow in the forward direction in the MLP (the blue arrows in Figure \ref{Img_mlp_information_flow1}).
However, all the information of $Y$ stems from $\boldsymbol{X}$ (the red dashed arrow in Figure \ref{Img_mlp_information_flow1}) and flows in the backward direction (the green arrows in Figure \ref{Img_mlp_information_flow1}) imply the information of $\boldsymbol{X}$ flows in both the forward and the backward directions, i.e., it cannot satisfy any DPI in the MLP.

\subsection{A novel information theoretic explanation for the generalization of MLPs}
\label{generalization_it}

In terms of deep learning, generalization indicates the ability of neural networks adapting to new data, which does not belong to the training dataset $\mathcal{D}$ but is drawn from the same distribution $P(\boldsymbol{X},Y)$.
Based on the above information theoretic explanations, we propose an information theoretic explanation for the generalization of MLPs.
Specifically, the performance of the MLP on the training dataset $\mathcal{D}$ can be measured by $I(Y, F_Y)$, and the generalization performance of the MLP can be measured by $I(\bar{X}, F_1)$.

Based on $H(Y) = I(\boldsymbol{X}, Y)$, we can derive 
\begin{equation}
\label{xyxc}
I(\boldsymbol{X}, F_i) = I(Y, F_i) + I(\bar{X}, F_i), 
\end{equation}
where $\bar{X} = Y^c \cap \boldsymbol{X}$ is the relative complement $Y$ in $\boldsymbol{X}$.
The information theoretic relationship is shown in Figure \ref{Img_mlp_information_venn}.

The performance the MLP on the training dataset $\mathcal{D}$ can be measured by $I(Y, F_Y)$.
Since we prove the MLP as the Gibbs distribution $P(F_Y|\boldsymbol{X})$ for learning $P(Y|\boldsymbol{X})$ in Section \ref{probabilistic_mlp} and \ref{prob_training}, we have $P_{F_Y|\boldsymbol{X}}(y^j|\boldsymbol{x}^j) = P_{Y|\boldsymbol{X}}(y^j|\boldsymbol{x}^j) = 1$\footnote{$P_{F_Y|\boldsymbol{X}}(y^j|\boldsymbol{x}^j) = P(F_Y=y^j|\boldsymbol{X}=\boldsymbol{x}^j)$ for simplicity.} if the cross entropy loss decreases to zero.
As a result, we have $H(F_Y|\boldsymbol{X}) = 0$, thereby
\begin{equation}
\label{mi_fYX}
I(F_Y, X) = H(F_Y) - H(F_Y|\boldsymbol{X})  = H(F_Y).
\end{equation}

To derive $H(F_Y)$, we can reformulate $P(F_Y)$ as
\begin{equation}
P(F_Y) = \sum_{\boldsymbol{x} \in \mathcal{X}}P(F_Y|\boldsymbol{X}=\boldsymbol{x})P(\boldsymbol{X}=\boldsymbol{x}).
\end{equation}
Though $P(\boldsymbol{X}=\boldsymbol{x})$ is intractable due to $P(\boldsymbol{X})$ is unknown, we can simplify $P(F_Y)$ as follows when $\mathcal{D}$ includes large enough \textit{i.i.d.}\ samples, i.e., $\{\boldsymbol{x}^j\}_{j=1}^J \approx \mathcal{X}$.
\begin{equation}
P(F_Y = y^j) = \frac{1}{J}\sum_{\boldsymbol{x}^j \in \mathcal{D}}P(F_Y = y^j|\boldsymbol{X}=\boldsymbol{x}^j).
\end{equation}
where $P(\boldsymbol{X}=\boldsymbol{x}^j) = 1/J$ because $\boldsymbol{x}^j$ are \textit{i.i.d.}\ samples.
If $P_{F_Y|\boldsymbol{X}}(y^j|\boldsymbol{x}^j) = P_{Y|\boldsymbol{X}}(y^j|\boldsymbol{x}^j) = 1$, we derive $P(F_Y) = P(Y)$ and $H(F_Y) = H(Y)$.
Overall, if the cross entropy loss decreases to zero, $I(\boldsymbol{X}, F_Y) = H(Y)$, i.e., $F_Y$ contains all the information of $Y$, otherwise $I(\boldsymbol{X}, F_Y) < H({Y})$.

The generalization performance of the MLP can be measured by $I(\bar{X}, F_1)$.
First, $H(Y) = I(\boldsymbol{X}, Y)$ implies that the generalization of the MLP is entirely determined by how much information of $\boldsymbol{X}$ the MLP has.
Second, when the cross entropy loss decreases to zero, $I(Y, \text{MLP})$ achieves the maximum $H({Y})$ in $\boldsymbol{f}_Y$, thus $I(\boldsymbol{X}, \text{MLP})$ only depends on $I(\bar{X}, \text{MLP})$ based on Equation \ref{xyxc}.
In other words, if $I(\bar{X}, \text{MLP})$ is large, then $I(\boldsymbol{X}, \text{MLP})$ is large and the MLP has good generalization.

The information flow of $\bar{X}$ in the MLP satisfies a DPI. 
Specifically, $\bar{X}$ does not contain information of $Y$, thus it cannot flow in the backward direction, i.e., it can only flow in the forward direction (the blue arrows in Figure \ref{Img_mlp_information_flow1}).
We introduce a Markov chain $\bar{X} \leftrightarrow F_1 \leftrightarrow F_2 \leftrightarrow F_Y$ and the corresponding DPI is
\begin{equation}
\label{dpi_xbar}
I(\bar{X}, F_1) \geq I(\bar{X}, F_2) \geq I(\bar{X}, F_Y),
\end{equation}
thus $I(\bar{X}, \text{MLP})$ can be simplified as $I(\bar{X}, F_1)$.

In summary, the performance the MLP on the training dataset $\mathcal{D}$ can be measured by $I(Y, F_Y)$, and the generalization of the MLP can be measured by $I(\bar{X}, F_1)$.

\section{The mutual information estimation}
\label{mi_calculation}

In this section, we estimate the mutual information $I(F_i, \boldsymbol{X})$ and $I(F_i, Y)$ based on the probability space $(\Omega_{F_i}, \mathcal{T}, P_{F_i})$.

\subsection{The estimation of $I(\boldsymbol{X}, F_i)$}

Based on the definition of mutual information, we have 
\begin{equation}
\label{mi_x}
I(\boldsymbol{X}, {F_i}) = H(F_i) - H(F_i|\boldsymbol{X}).
\end{equation}
Notably, all the previous works estimate $I(\boldsymbol{X}, {F_i})$ as $H(F_i)$, because IB supposes that $F_i$ is entirely determined by $\boldsymbol{X}$, namely $H(F_i|\boldsymbol{X}) = 0$. 
However, we have $H(F_i|\boldsymbol{X}) \neq 0$ in Section \ref{xyfi}.
As a result, we should take into account $H(F_i|\boldsymbol{X})$ for precisely estimating $I(\boldsymbol{X}, {F_i})$.

The key to estimating $I(\boldsymbol{X}, {F_i})$ is specifying $P(F_i|\boldsymbol{X})$ and $P(F_i)$.
Based on the definitions of $(\Omega_{F_i}, \mathcal{T}, P_{F_i})$ and the random variable $F_i : \Omega_{F_i} \rightarrow E_{F_i}$ (Section \ref{prob_space}), we have
\begin{equation}
\label{gibbs_fi}
P_{F_i|\boldsymbol{X}}(n|\boldsymbol{x}^j) = \frac {1}{Z_{F_i}}\text{exp}[\sigma_i(\langle \boldsymbol{\omega}^{(i)}_n, \boldsymbol{f}_{1\rightarrow (i-1)}(\boldsymbol{x}^j) \rangle + b_{in})].
\end{equation}
where $\boldsymbol{f}_i$ has $N$ neurons, i.e., $n \in [1,N]$, and $\boldsymbol{f}_{1\rightarrow (i-1)}(\boldsymbol{x}^j)$ is the input of $\boldsymbol{f}_i$, i.e., the output of the hidden layers from the first one to $(i-1)$th one given $\boldsymbol{x}^j$. 
More specifically, the $P_{F_i|\boldsymbol{X}}(n|\boldsymbol{x}^j)$ corresponding to the three fully connected layers in the MLP can be expressed as
\begin{equation}
\begin{split}
P_{F_1|\boldsymbol{X}}(n|\boldsymbol{x}^j) &= \frac {1}{Z_{F_1}}\text{exp}[\sigma_1(\langle \boldsymbol{\omega}^{(1)}_n, \boldsymbol{x}^j \rangle + b_{1n})]\\
P_{F_2|\boldsymbol{X}}(k|\boldsymbol{x}^j) &= \frac {1}{Z_{F_2}}\text{exp}[\sigma_2(\langle \boldsymbol{\omega}^{(2)}_k, \boldsymbol{f}_1(\boldsymbol{x}^j) \rangle + b_{2k})]\\
P_{F_Y|\boldsymbol{X}}(l|\boldsymbol{x}^j) &= \frac {1}{Z_{F_Y}}\text{exp}[\langle \boldsymbol{\omega}^{(3)}_l, \boldsymbol{f}_2(\boldsymbol{f}_1(\boldsymbol{x}^j)) \rangle + b_{yk}]\\
\end{split}
\end{equation}
  
In addition, we derive the marginal distribution $P_{F_i}(n)$ from the joint distribution $P(F_i, \boldsymbol{X})$ as
\begin{equation}
\begin{split}
P(F_i = n) &= \sum_{\boldsymbol{x} \in \mathcal{X}} P(F_Y = n, \boldsymbol{X} = \boldsymbol{x})\\
&= \sum_{\boldsymbol{x} \in \mathcal{X}}P(\boldsymbol{X}=\boldsymbol{x})P(F_Y = n|\boldsymbol{X}=\boldsymbol{x}).
\end{split}
\end{equation}
Since $P(\boldsymbol{X})$ is unknown and $\mathcal{D} \rightarrow \mathcal{X}$ as long as $\mathcal{D}$ includes large enough \textit{i.i.d.}\ samples, we relax $P(F_i = n)$ as
\begin{equation}
\label{average_gibbs_fi}
\begin{split}
P(F_i=n) & = \sum_{\boldsymbol{x}^j \in \mathcal{D}}P(\boldsymbol{X} =\boldsymbol{x}^j)P({F_i} = n|\boldsymbol{X} =\boldsymbol{x}^j)\\
& = \frac{1}{J}\sum_{\boldsymbol{x}^j \in \mathcal{D}}P({F_i} = n|\boldsymbol{X} =\boldsymbol{x}^j).
\end{split}
\end{equation}

We can observe that $P_{F_i}(n)$ measures the average probability of $\boldsymbol{\omega}_n^{(i)}$ occurring in the entire dataset $\{\boldsymbol{x}^j\}_{j=1}^J$, and $P_{F_i|\boldsymbol{X}}(n|\boldsymbol{x}^j) $ measures the probability of $\boldsymbol{\omega}_n^{(i)}$ occurring in the single data $\boldsymbol{x}^j$.
Based on the equivalence between the Kullback-Leibler (KL) divergence and mutual information, i.e., $I(\boldsymbol{X}, {F_i}) = E_{\boldsymbol{X}}\text{KL}[P(F_i|\boldsymbol{X})||P(F_i)]$ \citep{it_book}, we can conclude that $I(\boldsymbol{X}, {F_i})$ would be small if the probability of $\{\boldsymbol{\omega}_n^{(i)}\}_{n=1}^N$ occurring in each single data is close to that of $\{\boldsymbol{\omega}_n^{(i)}\}_{n=1}^N$ occurring the entire dataset, otherwise $I(\boldsymbol{X}, {F_i})$ would be large.

\subsection{The estimation of $I(\boldsymbol{F_i}, \boldsymbol{Y})$}

Based on the definition of mutual information, we have 
\begin{equation}
\label{mi_y}
I(Y, {F_i}) = H(F_i) - H(F_i|Y).
\end{equation}
We use the same method to estimate $P(F_i)$. 
However, since $P_{F_i|Y}(n|l)$ is intractable, we alternatively extend it as
\begin{equation}
\begin{split}
P_{F_i|Y}(n|l) = \sum_{\boldsymbol{x}^j \in \boldsymbol{\mathcal{D}}}P_{F_i|\boldsymbol{X}}(n|\boldsymbol{x}^j)P_{\boldsymbol{X}|Y}(\boldsymbol{x}^j|l).
\end{split}
\end{equation}

Since $\{\boldsymbol{x}^j\}_{j=1}^J$ is supposed to be \textit{i.i.d.}, $P_{\boldsymbol{X}|Y}(\boldsymbol{x}^j|l) = \frac{1}{N(l)}$ if $y^j = l$, otherwise $P_{\boldsymbol{X}|Y}(\boldsymbol{x}^j|l) = 0$, where $N(l)$ denotes the number of samples with the label $l$.
As a result, we have
\begin{equation}
\label{pdf_f_given_y}
P_{F_i|Y}(n|l) = \frac{1}{N(l)}\sum_{\boldsymbol{x}^j \in \mathcal{D}, y^j = l}P_{F_i|\boldsymbol{X}}(n|\boldsymbol{x}^j),
\end{equation}
which measures the probability of $\boldsymbol{\omega^{(i)}_n}$ occurring in the entire dataset with the label $l$.
Finally, we can derive $I(Y, F_i)$ based on Equation \ref{average_gibbs_fi} and \ref{pdf_f_given_y}.

Similarly, based on $I(Y, {F_i}) = E_{Y}\text{KL}[P(F_i|Y)||P(F_i)]$, we can conclude that $I(Y, {F_i})$ would be small if the probability of $\{\boldsymbol{\omega}_n^{(i)}\}_{n=1}^N$ occurring in the dataset with each label is close to that of $\{\boldsymbol{\omega}_n^{(i)}\}_{n=1}^N$ occurring in the entire dataset, otherwise $I(Y, {F_i})$ would be large.

In summary, we introduces a new method to estimate $I(\boldsymbol{X}, F_i)$ and $I(Y, F_i)$ in this section based on the definitions of $(\Omega_{F_i}, \mathcal{T}, P_{F_i})$ and the random variable $F_i : \Omega_{F_i} \rightarrow E_{F_i}$.

\begin{figure*}
\centering
\includegraphics[scale=0.58]{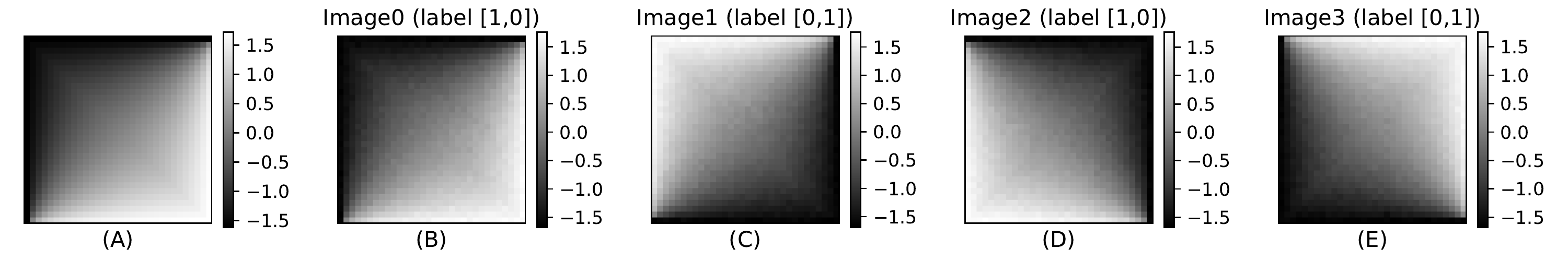}
\caption{\small{
(A) shows the deterministic image $\hat{\boldsymbol{x}}$. 
All the synthetic images $\boldsymbol{x}$ are generated by rotating $\hat{\boldsymbol{x}}$ and adding the Gaussian noise $\mathcal{N}(\mu_{\boldsymbol{x}}, 0.1)$, i.e., $\boldsymbol{x} = r(\hat{\boldsymbol{x}}) + \mathcal{N}(\mu_{\boldsymbol{x}}, 0.1)$, where $r(\cdot)$ defines totally four different ways to rotate $\hat{\boldsymbol{x}}$ and $\mu_{\boldsymbol{x}}$ is the expectation of ${\boldsymbol{x}}$.
Specifically, Image0 is the synthetic image generated by adding $\mathcal{N}(\mu_{\boldsymbol{x}}, 0.1)$ without rotation,
Image1 is the synthetic image generated by rotating $\hat{\boldsymbol{x}}$ along the secondary diagonal direction and adding $\mathcal{N}(\mu_{\boldsymbol{x}}, 0.1)$,
Image2 is the synthetic image generated by rotating $\hat{\boldsymbol{x}}$ along the vertical direction and adding $\mathcal{N}(\mu_{\boldsymbol{x}}, 0.1)$,
and Image3 is the synthetic image generated by rotating $\hat{\boldsymbol{x}}$ along the horizontal direction and adding $\mathcal{N}(\mu_{\boldsymbol{x}}, 0.1)$.
The four images are categorized into two different classes: Image0 and Image2 with label $[1,0]$, and Image1 and Image3 with label $[0, 1]$.
}}
\label{fig_iid_images1}
\end{figure*}

\section{Experiments}
\label{experiments}

In this section, we present two set of experiments based on the synthetic dataset and benchmark datasets to demonstrate the probabilistic representation and the information theoretic explanations for MLPs in Section \ref{prob_exp}, \ref{info_exp}, and \ref{mi_calculation}. All the simulation codes are available online\footnote{\url{https://github.com/EthanLan/DNN_Information_theory}}.

\subsection{Setup}

We generate a synthetic dataset consisting of 256 $32 \times 32$ grayscale images based on the deterministic image $\hat{\boldsymbol{x}}$ shown in Figure \ref{fig_iid_images1}(A). 
A synthetic image $\boldsymbol{x}$ is generated by rotating $\hat{\boldsymbol{x}}$ and adding the Gaussian noise $\mathcal{N}(\mu_{\boldsymbol{x}}, \sigma^2 = 0.1)$,
\begin{equation}
\boldsymbol{x} = r(\hat{\boldsymbol{x}}) + \mathcal{N}(\mu_{\boldsymbol{x}}, 0.1)
\end{equation}
where $r(\cdot)$ totally defines four different ways to rotate $\hat{\boldsymbol{x}}$ shown in Figure \ref{fig_iid_images1}(B)-(E), and $\mu_{\boldsymbol{x}}$ denotes the expectation of ${\boldsymbol{x}}$.
The reason for adding Gaussian noise is to avoid MLPs directly memorize the deterministic image.

The synthetic dataset evenly consists of 64 images with the four different rotation ways shown in Figure \ref{fig_iid_images1}(B)-(E).
Compared to benchmark datasets with complicated features, the synthetic dataset only has four simple features, namely the four different rotation ways.
As a result, we can clearly demonstrate the proposed probabilistic explanations for MLPs by visualizing the weights of MLPs.

Compared to benchmark datasets with unknown entropy, the entropy of the synthetic dataset is known.
If we do not take into account the additive Gaussian noise, the entropy of the synthetic dataset would be exactly 2 bits.
Since the noise is Gaussian, the differential entropy of the noise is $\frac{1}{2}log(2\pi e \sigma^2) \approx 0.38$ bits.
Therefore,  the total entropy of the synthetic dataset is 2.38 bits, because the additive noise is independent on the rotation ways.
Since the labels $[1, 0]$ and $[0, 1]$ evenly divide the synthetic dataset into two classes, the entropy of the labels is 1 bit.
As a result, the synthetic dataset enables us to precisely examine the existing and the proposed information theoretic explanations for MLPs.

\subsection{The simulations on the synthetic dataset}

This section demonstrates five aspects: (i) the probability space $(\Omega_{F_i}, \mathcal{T}, P_{F_i})$ for a fully connected layer $\boldsymbol{f}_i$; (ii) the effect of an activation function $\sigma_i(\cdot)$ on $P_{F_i}$; (iii) the effect of $\sigma_i(\cdot)$ on $H(F_i)$, $I(\boldsymbol{X},F_i)$, and $I(Y, F_i)$; (iv) the information flow in the MLP, i.e., the variation of $I(\boldsymbol{X},F_i)$, $I(Y, F_i)$, and $I(\bar{X},F_i)$ over different layers, and (v) the comparison of the proposed mutual information estimator and two existing non-parametric estimators on the synthetic dataset.

\pagebreak
To classify the synthetic dataset, we specify the MLP as follows: (i) since a single image is $32 \times 32$, the input layer $\boldsymbol{x}$ has $M = 1024$ nodes, (ii) two hidden layers $\boldsymbol{f_1}$ and $\boldsymbol{f_2}$ have $N = 8$ and $K = 6$ neurons, respectively, and (iii) the output layer $\boldsymbol{f_Y}$ has $L = 2$ nodes corresponding to the labels of the dataset. In addition, all the activation functions are chosen as ReLU $\sigma(x) = max(0, x)$ unless otherwise specified.

\subsubsection{The probability space for a layer}

\begin{figure*}
\centering
\includegraphics[scale=0.62]{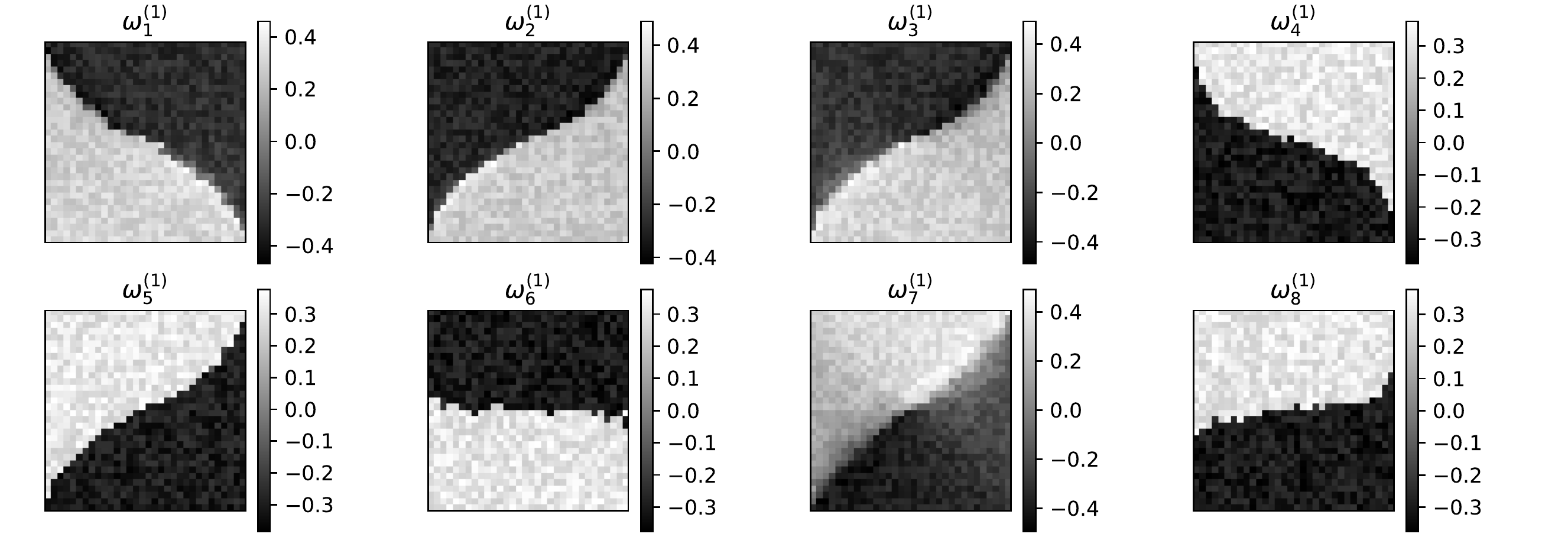}
\caption{\small{
The eight possible outcomes $\{\boldsymbol{\omega}^{(1)}_{n}\}_{n=1}^{8}$ represented by the learned weights of the eight neurons, where $\boldsymbol{\omega}^{(1)}_n = \{\omega^{(1)}_{mn}\}_{m=1}^{1024}$.
All the weights $\{\omega^{(1)}_{mn}\}_{m=1}^{1024}$ are reshaped into $32 \times 32$ dimension for visualizing the spatial structure.
}}
\label{Img_mlp_gibbs_sim1_features}
\end{figure*}

\begin{table*}
\caption{The Gibbs measure of the first hidden layer $\boldsymbol{f_1}$ given the synthetic images in Figure \ref{fig_iid_images1}}
\label{fc_gibbs}
\vskip 0.1in
\begin{center}
\begin{small}
\begin{threeparttable}
\begin{tabular}{ccccccccc}
\toprule
 & $\boldsymbol{\omega^{(1)}_{1}}$ & $\boldsymbol{\omega^{(1)}_{2}}$ & $\boldsymbol{\omega^{(1)}_{3}}$ & $\boldsymbol{\omega^{(1)}_{4}}$ & $\boldsymbol{\omega^{(1)}_{5}}$ & $\boldsymbol{\omega^{(1)}_{6}}$ & $\boldsymbol{\omega^{(1)}_{7}}$ & $\boldsymbol{\omega^{(1)}_{8}}$ \\
\midrule
$g_{1n}(\boldsymbol{x})$& {45.3} & \textbf{215.7} & 206.2 & -62.7 & -222.9 & {137.1} & -202.5 & -171.6\\
$f_{1n}(\boldsymbol{x})$& {45.3} & \textbf{215.7} & 206.2 & 0.0 & 0.0 & {137.1} & 0.0 & 0.0\\
 $\text{exp}[f_{1n}(\boldsymbol{x})]$ & {4.71e+19} & \textbf{4.75e+93} & 3.56e+89 & 1.0 & {1.0} & {3.48e+59} & 1.0 & 1.0   \\
 $P_{F_1|\boldsymbol{X}}(\boldsymbol{\omega^{(1)}_{n}}|\text{Image0})$ & 0.0 & \textbf{1.0} & 0.0 & 0.0 & 0.0 & 0.0 & 0.0 & 0.0   \\
\midrule
$g_{1n}(\boldsymbol{x})$& {-53.5} & -217.7 & -208.4 & 69.0 & \textbf{224.8} & {-134.6} & 204.1 & 171.3\\
$f_{1n}(\boldsymbol{x})$& {0.0} & {0.0} & 0.0 & 69.0 & \textbf{224.8} & {0.0} & 204.1 & 171.3\\
$\text{exp}[f_{1n}(\boldsymbol{x})]$ & 1.0 & 1.0 & 1.0 & {9.25e+29} & \textbf{4.25e+97} & 1.0 & 4.36e+88 & {2.48e+74}   \\
$P_{F_1|\boldsymbol{X}}(\boldsymbol{\omega^{(1)}_{n}}|\text{Image1})$ & 0.0 & {0.0} & {0.0} & 0.0 & \textbf{1.0} & 0.0 & 0.0 & 0.0   \\
\midrule
$g_{1n}(\boldsymbol{x})$& \textbf{219.4} & {54.9} & 78.9 & -211.3 & -37.4 & {153.6} & -106.6 & -116.4\\
$f_{1n}(\boldsymbol{x})$& \textbf{219.4} & {54.9} & 78.9 & 0.0 & 0.0 & {153.6} & 0.0 & 0.0\\
$\text{exp}[f_{1n}(\boldsymbol{x})]$ & \textbf{1.92e+95} & {6.96e+23} & {1.84e+34} & 1.0 & 1.0 & 5.10e+66 & 1.0 & 1.0   \\
$P_{F_1|\boldsymbol{X}}(\boldsymbol{\omega^{(1)}_{n}}|\text{Image2})$ & \textbf{1.0} & 0.0 & 0.0 & 0.0 & 0.0 & 0.0 & 0.0 & {0.0}   \\
 \midrule
$g_{1n}(\boldsymbol{x})$& {-219.0} & -55.9 & -81.6 & \textbf{208.0} & {41.3} & -159.6 & 111.8 & 122.1\\
$f_{1n}(\boldsymbol{x})$& {0.0} & {0.0} & 0.0 & \textbf{208.0} & 41.3 & {0.0} & 111.8 & 122.1\\
$\text{exp}[f_{1n}(\boldsymbol{x})]$ & {1.0} & 1.0 & 1.0 & \textbf{2.15e+90} & 8.63e+17 & 1.0 & {3.58e+48} & 1.06e+53   \\
$P_{F_1|\boldsymbol{X}}(\boldsymbol{\omega^{(1)}_{n}}|\text{Image3})$ & {0.0} & 0.0 & 0.0 & \textbf{1.0} & 0.0 & 0.0 & 0.0 & 0.0   \\
\bottomrule
\end{tabular}
\begin{tablenotes}
            \item $g_{1n}(\boldsymbol{x})$ and $f_{1n}(\boldsymbol{x})$ are the linear output and the activation, respectively, where $g_{1n}(\boldsymbol{x}) = \langle \boldsymbol{\omega}^{(1)}_{n}, \boldsymbol{x} \rangle + b_{1n}$ and $f_{1n} = \sigma_1[g_{1n}(\boldsymbol{x})]$.
\end{tablenotes}
\end{threeparttable}
\end{small}
\end{center}
\vskip -0.1in
\end{table*}

To demonstrate the proposed probability space for all the fully connected layers in the $\text{MLP} = \{\boldsymbol{x; f_1; f_2; f_Y}\}$, we only need to demonstrate $(\Omega_{F_1}, \mathcal{T}, P_{F_1})$ for $\boldsymbol{f_1}$, because we derive $(\Omega_{F_i}, \mathcal{T}, P_{F_i})$ for each layer in the backward direction based on the mathematical induction in Section \ref{prob_space}.
\pagebreak

First, we demonstrate the sample space $\Omega_{F_1} = \{\boldsymbol{\omega}^{(1)}_{n}\}_{n=1}^N$ for $\boldsymbol{f_1}$.
We train the MLP on the synthetic dataset until the training accuracy is $100\%$ and visualize the learned weights of the eight neurons, i.e., $\boldsymbol{\omega}^{(1)}_{n} = \{\omega^{(1)}_{mn}\}_{m=1}^{1024}, n \in [1,8],$ in Figure \ref{Img_mlp_gibbs_sim1_features}, from which we observe that (i) $\boldsymbol{\omega}^{(1)}_{n}$ can be regarded as a possible outcome (i.e., the feature of $\boldsymbol{x}$), e.g., $\boldsymbol{\omega}^{(1)}_2$ has low magnitude at top-left positions and high magnitude at bottom-right positions, which describes the spatial feature of $\boldsymbol{x}= \text{Image0}$ in Figure \ref{fig_iid_images1}; 
and (ii) the weights of different neurons formulate different features. 
Though the weights of some neurons, e.g., $\boldsymbol{\omega}^{(1)}_2$ and $\boldsymbol{\omega}^{(1)}_3$, are similar, they still can be viewed as different features, because their weights with the same index are different, i.e., $\forall n \neq n', \omega^{(1)}_{mn} \neq \omega^{(1)}_{mn'}$.
\pagebreak

Second, we demonstrate the Gibbs measure $P_{F_1}$ for $\boldsymbol{f}_1$.
Based on Equation \ref{Gibbs_f1},
we derive ${\textstyle P_{F_1|\boldsymbol{X}}(\boldsymbol{\omega}^{(1)}_{n}|\boldsymbol{x})}$ given the four images in Figure \ref{fig_iid_images1}(B)-(E). 
Table \ref{fc_gibbs} shows that $P_{F_1}$ correctly measures the probability of ${\textstyle \{\boldsymbol{\omega}^{(1)}_{n}\}_{n=1}^8}$.
For instance, $\boldsymbol{\omega}^{(1)}_{2}$ correctly describes the feature of Image0, thus it has the largest linear output $g_{12}(\boldsymbol{x}) = 215.7$ and activation $f_{12}(\boldsymbol{x}) = 215.7$, thereby ${\textstyle P_{F_1|\boldsymbol{X}}(\boldsymbol{\omega}^{(1)}_{2}|\text{Image0}) = 1}$.
As a comparison, $\boldsymbol{\omega}^{(1)}_{5}$ incorrectly describes the feature of Image0, thus it has the lowest linear output $g_{15}(\boldsymbol{x}) = -222.9$ and activation $f_{12}(\boldsymbol{x}) = 0.0$, so $P_{F_1|\boldsymbol{X}}(\boldsymbol{\omega}^{(1)}_{5}|\text{Image0}) = 0$.

\subsubsection{The effect of activation functions on the Gibbs probability measure $P_{F}$}
\label{act_function_pf}


\begin{table*}
\caption{The Gibbs probability $P_{F_1|\boldsymbol{X}}(\boldsymbol{\omega^{(1)}_{n}}|\text{Image0})$ with four different activation functions and the corresponding conditional entropy $H(F_1|\boldsymbol{X}=\text{Image0})$}
\label{activation_gibbs}
\vskip 0.1in
\begin{center}
\begin{small}
\begin{threeparttable}
\begin{tabular}{cccccccccc}
\toprule
& $\boldsymbol{\omega^{(1)}_{1}}$ & $\boldsymbol{\omega^{(1)}_{2}}$ & $\boldsymbol{\omega^{(1)}_{3}}$ & $\boldsymbol{\omega^{(1)}_{4}}$ & $\boldsymbol{\omega^{(1)}_{5}}$ & $\boldsymbol{\omega^{(1)}_{6}}$ & $\boldsymbol{\omega^{(1)}_{7}}$ & $\boldsymbol{\omega^{(1)}_{8}}$ & $H(F_1|\boldsymbol{X})$ \\
\midrule
$g_{1n}(\boldsymbol{x})$& {45.3} & \textbf{215.7} & 206.2 & -62.7 & -222.9 & {137.1} & -202.5 & -171.6\\
\midrule
$f_{1n}^{\text{Linear}}(\boldsymbol{x})$& {45.3} & \textbf{215.7} & 206.2 & -62.7 & -222.9 & {137.1} & -202.5 & -171.6\\
$\text{exp}[f_{1n}^{\text{Linear}}(\boldsymbol{x})]$ & {4.71e+19} & \textbf{4.75e+93} & 3.56e+89 & 5.88e-28 & {1.56e-97} & {3.48e+59} & 1.13e-88 & 2.98e-75   \\
$P(F_1|\boldsymbol{X})$ & 0.0 & \textbf{1.0} & 0.0 & 0.0 & 0.0 & {0.0} & 0.0 & 0.0   & 0.00\\
\midrule
$f_{1n}^{\text{ReLU}}(\boldsymbol{x})$& {45.3} & \textbf{215.7} & 206.2 & 0.0 & 0.0 & {137.1} & 0.0 & 0.0\\
$\text{exp}[f_{1n}^{\text{ReLU}}(\boldsymbol{x})]$ & {4.71e+19} & \textbf{4.75e+93} & 3.56e+89 & 1.0 & {1.0} & {3.48e+59} & 1.0 & 1.0   \\
$P(F_1|\boldsymbol{X})$ & 0.0 & \textbf{1.0} & 0.0 & 0.0 & 0.0 & 0.0 & 0.0 & 0.0  & 0.00\\
 \midrule
$f_{1n}^{\text{Tanh}}(\boldsymbol{x})$ & \textbf{1.0} & \textbf{1.0} & \textbf{1.0} & -1.0 & -1.0 & \textbf{1.0} & -1.0 & -1.0\\
$\text{exp}[f_{1n}^{\text{Tanh}}(\boldsymbol{x})]$ & \textbf{2.71} & \textbf{2.71} & \textbf{2.71} & 0.36 & 0.36 & \textbf{2.71} & 0.36 & 0.36   \\
$P(F_1|\boldsymbol{X})$ & \textbf{0.22} & \textbf{0.22} & \textbf{0.22} & 0.03 & {0.03} & \textbf{0.22} & 0.03 & 0.03 & 2.53   \\
\midrule
$f_{1n}^{\text{Sigmoid}}(\boldsymbol{x})$ & \textbf{1.0} & \textbf{1.0} & \textbf{1.0} & 0.0 & 0.0 & \textbf{1.0} & 0.0 & 0.0\\
$\text{exp}[f_{1n}^{\text{Sigmoid}}(\boldsymbol{x})]$ & \textbf{2.71} & \textbf{2.71} & \textbf{2.71} & 1.0 & {1.0} & \textbf{2.71} & 1.0 & 1.0   \\
$P(F_1|\boldsymbol{X})$ & \textbf{0.18} & \textbf{0.18} & \textbf{0.18} & {0.07} & {0.07} & \textbf{0.18} & {0.07} & {0.07} & 2.84   \\
\bottomrule
\end{tabular}
\begin{tablenotes}
            \item $f_{1n}^{\text{Linear}}(\boldsymbol{x})$ denotes the activation without activation function. $f_{1n}^{\text{ReLU}}(\boldsymbol{x})$, $f_{1n}^{\text{Tanh}}(\boldsymbol{x})$, and $f_{1n}^{\text{Sigmoid}}(\boldsymbol{x})$ denote the activations with different activation functions given the same linear output $g_{1n}(\boldsymbol{x})$. $H(F_1|\boldsymbol{X})$ denotes $H(F_1|\boldsymbol{X}=\text{Image0})$ for simplicity.
\end{tablenotes}
\end{threeparttable}
\end{small}
\end{center}
\vskip -0.1in
\end{table*}

To demonstrate the effect of activation functions on the Gibbs measure, we examine $P_{\boldsymbol{F}_1|\boldsymbol{X=x}}(\boldsymbol{\omega}^{(1)}_{n}|\text{Image0})$ in four different cases: (i)  the linear activation function $\sigma_1(x) = x$, (ii) ReLU $\sigma_1(x) = max(0, x)$, (iii) the hyperbolic tangent function (abbr. Tanh) $\sigma_1(x) = (\text{e}^{x}-\text{e}^{-x})/(\text{e}^{x}+\text{e}^{-x})$, and (iv) the sigmoid function $\sigma_1(x) = 1/(1+ \text{e}^{-x})$ in Table \ref{activation_gibbs}.

ReLU guarantees an accurate Gibbs measure because it only sets negative linear outputs as zeros.
For instance,  $\boldsymbol{\omega}^{(1)}_{5}$ is an irrelevant feature of Image0, Table \ref{activation_gibbs} shows ${\textstyle f_{15}^{\text{Linear}}(\boldsymbol{x})=}$ $-222.9$ and ${\textstyle \text{exp}[f_{15}^{\text{Linear}}(\boldsymbol{x})] = 1.56e-97}$. 
As a comparison, if we use ReLU, ${\textstyle f_{15}^{\text{ReLU}}(\boldsymbol{x}) = 0.0}$ and ${\textstyle \text{exp}[f_{15}^{\text{ReLU}}(\boldsymbol{x})] = 1.0}$.
The difference between ${\textstyle \text{exp}[f_{15}^{\text{ReLU}}(\boldsymbol{x})]}$ and ${\textstyle \text{exp}[f_{15}^{\text{Linear}}(\boldsymbol{x})]}$ are small, thus ${\textstyle P^{\text{ReLU}}_{F_1|\boldsymbol{X}}(\boldsymbol{\omega}^{(1)}_{5}|\text{Image0})}$ also should be close to ${\textstyle P^{\text{Linear}}_{F_1|\boldsymbol{X}}(\boldsymbol{\omega}^{(1)}_{5}|\text{Image0})}$, 
which is validated by the experiment, namely ${\textstyle P^{\text{ReLU}}_{F_1|\boldsymbol{X}}(\boldsymbol{\omega}^{(1)}_{5}|\text{Image0}) = P^{\text{Linear}}_{F_1|\boldsymbol{X}}(\boldsymbol{\omega}^{(1)}_{5}|\text{Image0}) = 0.0}$.
We observe similar results on other neurons in Table \ref{activation_gibbs}.

\pagebreak
Tanh cannot guarantee an accurate Gibbs measure as it decreases the difference between the activation of relevant features and that of irrelevant features. 
For instance, $\boldsymbol{\omega}^{(1)}_{2}$ is a relevant feature of Image0 with ${\textstyle f_{12}^{\text{Linear}}(\boldsymbol{x}) = 215.7}$, and $\boldsymbol{\omega}^{(1)}_{1}$ is an irrelevant feature of Image0 with ${\textstyle f_{11}^{\text{Linear}}(\boldsymbol{x}) = 45.3}$, thus ${\textstyle |f_{12}^{\text{Linear}}(\boldsymbol{x}) - f_{11}^{\text{Linear}}(\boldsymbol{x})| = 170.4}$.
As a comparison, if use Tanh, $|f_{12}^{\text{Tanh}}(\boldsymbol{x}) - f_{11}^{\text{Tanh}}(\boldsymbol{x})| = 0.0$, thus $P^{\text{Tanh}}_{F_1|\boldsymbol{X}}(\boldsymbol{\omega}^{(1)}_{1}|\text{Image0}) = P^{\text{Tanh}}_{F_1|\boldsymbol{X}}(\boldsymbol{\omega}^{(1)}_{2}|\text{Image0}) = 0.22$.
In other words, we cannot distinguish $\boldsymbol{\omega}^{(1)}_{2}$ and $\boldsymbol{\omega}^{(1)}_{1}$ based on Tanh.

Sigmoid cannot guarantee an accurate Gibbs measure due to the same reason.
In particular, since Sigmoid confines activations to the smaller range $[0,1]$, it further decreases the difference between the activation of relevant features and that of irrelevant features.
For instance, if we use Tanh, ${\textstyle |f_{12}^{\text{Tanh}}(\boldsymbol{x}) - f_{15}^{\text{Tanh}}(\boldsymbol{x})| = 2.0}$.
As a comparison, if we use Sigmoid, ${\textstyle |f_{12}^{\text{Sigmoid}}(\boldsymbol{x}) - f_{15}^{\text{Sigmoid}}(\boldsymbol{x})| = 1.0}$.
Consequently, ${\scriptstyle |P^{\text{Sigmoid}}_{F_1|\boldsymbol{X}}(\boldsymbol{\omega}^{(1)}_{2}|\text{Image0}) - P^{\text{Sigmoid}}_{F_1|\boldsymbol{X}}(\boldsymbol{\omega}^{(1)}_{5}|\text{Image0})| = 0.11}$ becomes smaller than ${\scriptstyle |P^{\text{Tanh}}_{F_1|\boldsymbol{X}}(\boldsymbol{\omega}^{(1)}_{2}|\text{Image0}) - P^{\text{Tanh}}_{F_1|\boldsymbol{X}}(\boldsymbol{\omega}^{(1)}_{5}|\text{Image0})| = 0.19}$, i.e., it becomes more difficult to distinguish $\boldsymbol{\omega}^{(1)}_{2}$ and $\boldsymbol{\omega}^{(1)}_{5}$ based on Sigmoid.

The experiment provides a probabilistic explanation for the limitation of saturating (i.e., bounded) activation functions (e.g., Tanh and Sigmoid) for training neural networks
\citep{glorot2010understanding}.
Since saturating activation functions confine activations in a very small range and decrease the difference between the activation of relevant features and that of irrelevant features, they make difficult to distinguish relevant features and irrelevant ones.
As a result, neural networks with saturating activation functions require more computation cost, e.g., more training time or more hidden layers, to achieve the same training result.

In summary, activation functions have a great effect on the Gibbs measure of a fully connected layer.
Consequently, a fully connected layer with different activation functions should have different entropy and mutual information, which is discussed in the next section.

\begin{table}
\caption{The distribution $P(F_1)$ based on different activation functions and their respective $H(F_1)$, $I(\boldsymbol{X},F_1)$, and $I(Y,F_1)$}
\label{f1_gibbs}
\vskip 0.1in
\begin{center}
\begin{small}
\begin{threeparttable}
\begin{tabular}{ccccccccc}
\toprule
& Linear & ReLU & Tanh & Sigmoid \\
\midrule
$P(F_1=\boldsymbol{\omega^{(1)}_{1}})$ & \textbf{0.25} & \textbf{0.25} & 0.12   & 0.12 \\
$P(F_1=\boldsymbol{\omega^{(1)}_{2}})$ & \textbf{0.25} & \textbf{0.25}  & 0.12  & 0.13   \\
$P(F_1=\boldsymbol{\omega^{(1)}_{3}})$ & 0.00 & 0.00 & 0.13 & 0.12\\
$P(F_1=\boldsymbol{\omega^{(1)}_{4}})$ & \textbf{0.25} & \textbf{0.25} & 0.13  & 0.13\\
$P(F_1=\boldsymbol{\omega^{(1)}_{5}})$ & \textbf{0.25} & \textbf{0.25} & 0.12 & 0.13\\
$P(F_1=\boldsymbol{\omega^{(1)}_{6}})$ & 0.00 & 0.00 & 0.12 & 0.13\\
$P(F_1=\boldsymbol{\omega^{(1)}_{7}})$ & 0.00 & 0.00 & 0.13 & 0.12\\
$P(F_1=\boldsymbol{\omega^{(1)}_{8}})$ & 0.00 & 0.00 & 0.13 & 0.12\\
\midrule
$H(F_1)$ & {2.00} & {2.00} & 3.00 & 3.00\\
$I(\boldsymbol{X},F_1)$ & {2.00} & {2.00} & 0.47 & 0.16\\
$I(Y,F_1)$ & 1.00 & {1.00} & 0.35 & 0.16\\
\bottomrule
\end{tabular}
\end{threeparttable}
\end{small}
\end{center}
\vskip -0.1in
\end{table}

\subsubsection{The effect of activation functions on $H(F_1)$, $I(\boldsymbol{X}, F_1)$, and $I(Y, F_1)$}
\label{act_function_mi}


Since Gaussian noise is not helpful for classifying the synthetic dataset, the upper bound of $I(\boldsymbol{X}, F_1) = 2$.
Since the label evenly divides the entire dataset into two groups, the upper bound of $I(Y, F_1) = H(Y) = 1$.
In addition, since each synthetic image only has one feature, $H({F_1}|\boldsymbol{X} = \boldsymbol{x})$ should be close to zero if $\boldsymbol{f}_1$ models the input precisely.

Table \ref{activation_gibbs} summarizes $H({F_1}|\boldsymbol{X} = \text{Image0})$ given different activation functions:
$H({F_1}|\boldsymbol{X} = \text{Image0}) = 0$ given ReLU, and $H({F_1}|\boldsymbol{X} = \text{Image0})  >  2$ given Tanh and Sigmoid.
We can conclude that $\boldsymbol{f}_1$ with Tanh or Sigmoid does not contain much information of Image0.

Table \ref{f1_gibbs} summarizes $I(\boldsymbol{X},F_1)$ given different activation functions based on Equation \ref{mi_x}. 
$I(\boldsymbol{X}, F_1) = H(\boldsymbol{X}) = 2.0$ given ReLU indicates that $\boldsymbol{f}_1$ with ReLU contains all the information of the entire dataset.
In contrast, $I(\boldsymbol{X}, F_1) = 0.47$ given Tanh indicates that $\boldsymbol{f}_1$ with Tanh does not contain much information of the entire dataset.

\begin{figure*}
\begin{minipage}[b]{0.99\linewidth}
\centering
\centerline{\includegraphics[scale=0.5]{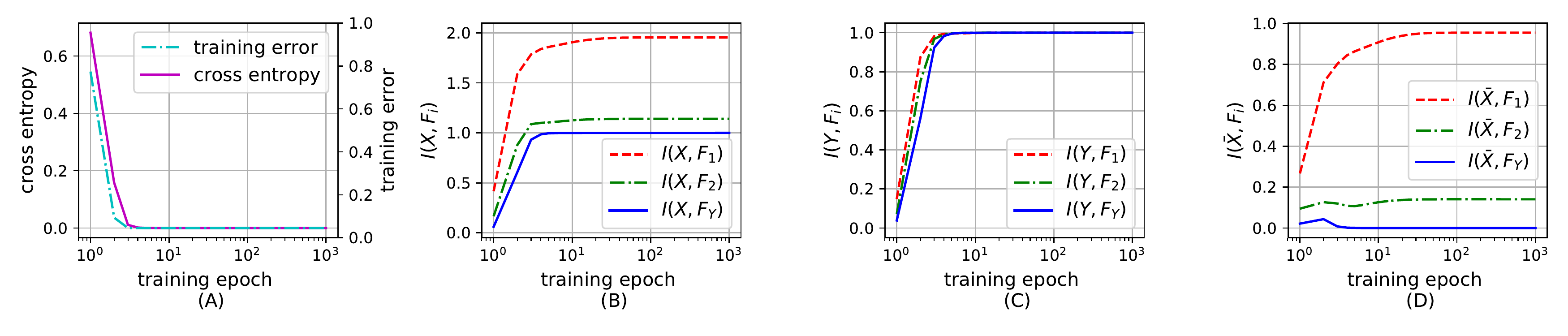}}
\end{minipage}
\vfill 
\begin{minipage}[b]{0.99\linewidth}
\centering
\centerline{\includegraphics[scale=0.5]{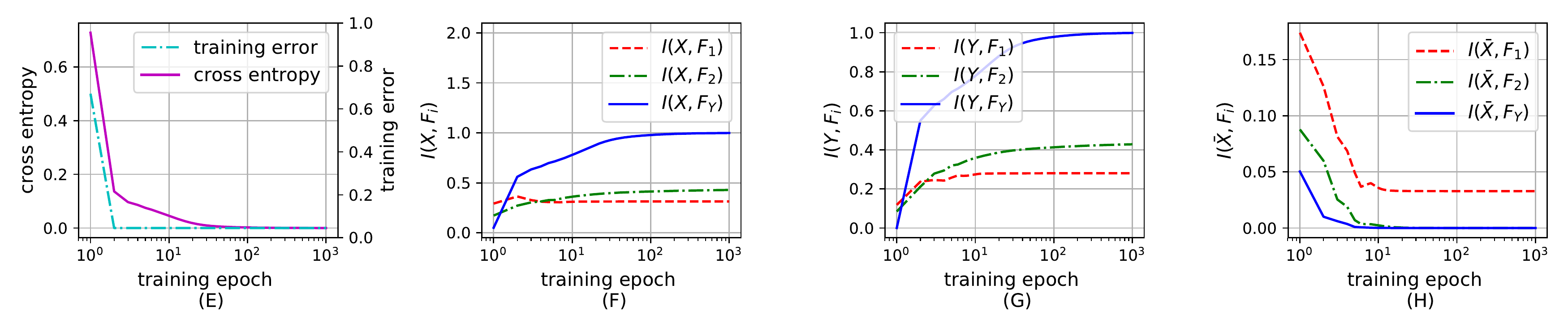}}
\end{minipage}
\caption{(A) and (E) visualize the variation of the training error and the cross entropy loss of MLP1 and MLP2 during training.
(B)-(D) visualize the variations of $I(\boldsymbol{X}, F_i)$,  $I(Y, F_i)$, and $I(\bar{X}, F_i)$ over all the layers in MLP1 during training, respectively.
Similarly, (F)-(H) visualize the variations of $I(\boldsymbol{X}, F_i)$,  $I(Y, F_i)$, and $I(\bar{X}, F_i)$ over all the layers in MLP2 during training, respectively.}
\label{Img_info_flow_(8_6_2)}
\end{figure*}

In addition, we derive $I(Y,F_1)$ based on Equation \ref{mi_y}.
$I(Y, F_1) = H(Y) = 1.0$ given ReLU indicates that $\boldsymbol{f}_1$ with ReLU contains all the information of labels.
As a comparison, $I(Y, F_1) = 0.35$ given Tanh indicates that $\boldsymbol{f}_1$ with Tanh only contains partial information of labels. 

\subsubsection{The information flow in the MLPs}
\label{if_mlp_synthetic}

\begin{table}
\caption{The number neurons(nodes) of the layer and the activation function of all the layers in the three MLPs}
\label{MLP3}
\vskip 0.1in
\begin{center}
\begin{small}
\begin{threeparttable}
\begin{tabular}{cccccc}
\toprule
 & $\boldsymbol{x}$ & $\boldsymbol{f}_1$ & $\boldsymbol{f}_2$ & $\boldsymbol{f}_Y$ & $\sigma(\cdot)$ \\
\midrule
MLP1 & 1024 & 8 & 6 & 2  & ReLU\\
MLP2 & 1024 & 8 & 6 & 2 & Tanh \\
MLP3 & 1024 & 1 & 6 & 2 & ReLU\\
\bottomrule
\end{tabular}
\end{threeparttable}
\end{small}
\end{center}
\vskip -0.1in
\end{table}

In this section, we demonstrate the proposed information theoretic explanations for the MLP in Section \ref{info_exp}.
We design three MLPs, namely MLP1, MLP2, and MLP3.
The difference between MLP1 and MLP2 is the activation functions, and the difference between MLP1 and MLP3 is the number of neurons in $\boldsymbol{f}_1$, which are summarized in Table \ref{MLP3}.

All the weights of the three MLPs are randomly initialized by a uniform distribution unless otherwise specified.
We choose the Adam algorithm  \citep{kingma2014adam}, a variant of the Stochastic Gradient Descent (SGD), to learn the weights of the three MLPs on the entire synthetic dataset over 1000 epochs with the learning rate 0.01.

Based on the synthetic dataset and the learned weights at each epoch, we can derive $I(\boldsymbol{X}, F_i)$,  $I(Y, F_i)$, and $I(\bar{X}, F_i)$ based on Equation (\ref{mi_x}), (\ref{mi_y}), (\ref{xyxc}), respectively.
To keep consistent with previous works \citep{chelombiev2018adaptive}, we train the MLPs with 50 random initializations and use the averaged mutual information to indicate the information flow in the MLPs.
Figure \ref{Img_info_flow_(8_6_2)}(A) and \ref{Img_info_flow_(8_6_2)}(E) show the variation of the cross entropy loss and the training error of MLP1 and MLP2, respectively. 
Figure \ref{Img_info_flow_(8_6_2)}(B)-\ref{Img_info_flow_(8_6_2)}(D) and \ref{Img_info_flow_(8_6_2)}(F)-\ref{Img_info_flow_(8_6_2)}(H) show the information flow in MLP1 and MLP2, respectively.

Since all the weights of MLPs are randomly initialized, $F_i$ initially is independent on $\boldsymbol{X}$ and $Y$.
As a result, $I(\boldsymbol{X}, F_i)$ and $I(Y, F_i)$ initially should be close to zero in MLP1 and MLP2, which is validated in Figure \ref{Img_info_flow_(8_6_2)}(B)-\ref{Img_info_flow_(8_6_2)}(C) and \ref{Img_info_flow_(8_6_2)}(F)-\ref{Img_info_flow_(8_6_2)}(G).
As training continues, Figure \ref{Img_info_flow_(8_6_2)}(B)-\ref{Img_info_flow_(8_6_2)}(C) and \ref{Img_info_flow_(8_6_2)}(F)-\ref{Img_info_flow_(8_6_2)}(G) show that $I(\boldsymbol{X}, F_i)$ and $I(Y, F_i)$ quickly converge to fixed values.
Specifically, Figure \ref{Img_info_flow_(8_6_2)}(B) and \ref{Img_info_flow_(8_6_2)}(F) show $I(\boldsymbol{X}, F_1)$ in MLP1 converging to 2 bits and $I(\boldsymbol{X}, F_1)$ in MLP2 converging to 0.47 bits, which are consistent with the results in Table \ref{f1_gibbs}.
Figure \ref{Img_info_flow_(8_6_2)}(C) and \ref{Img_info_flow_(8_6_2)}(G) show that MLP1 and MLP2 spend about 20 and 200 epochs on making $I(Y, F_Y)$ to converge to $H(Y) = 1$ bit, respectively.
That further confirms that saturating activation functions like Tanh cannot guarantee a precise Gibbs measure and require more training time to achieve the same training result.

In terms of the information flow of $\boldsymbol{X}$ in MLP1 and MLP2, Figure \ref{Img_info_flow_(8_6_2)}(B) shows ${\textstyle I(\boldsymbol{X}, F_1) \geq I(\boldsymbol{X}, F_2) \geq I(\boldsymbol{X}, F_Y)}$ in MLP1 after the cross entropy loss decreases to zero.
In contrast, Figure \ref{Img_info_flow_(8_6_2)}(F) shows ${\textstyle I(\boldsymbol{X}, F_Y) \geq I(\boldsymbol{X}, F_1) \geq I(\boldsymbol{X}, F_2)}$ in MLP2 after the cross entropy loss decreases to zero.
The results demonstrate that the information flow of $\boldsymbol{X}$ in MLPs cannot satisfy any DPI (Section \ref{info_flowyx}).

In terms of the information flow of $Y$ in MLP1 and MLP2,
Figure \ref{Img_info_flow_(8_6_2)}(C) shows $I(Y, F_1) = I(Y, F_2) = I(Y, F_Y)$ in MLP1 after the cross entropy loss decreases to zero.
In addition, Figure \ref{Img_info_flow_(8_6_2)}(G) shows $I(Y, F_Y) \geq I(Y, F_2) \geq I(Y, F_1)$ in MLP2 after the cross entropy loss decreases to zero.
The results demonstrate that the information flow of $Y$ in MLPs satisfies $I(Y, F_Y) \geq I(Y, F_2) \geq I(Y, F_1)$ (Section \ref{info_flowyx}).
\pagebreak

\begin{figure*}
\centering
\centerline{\includegraphics[scale=0.5]{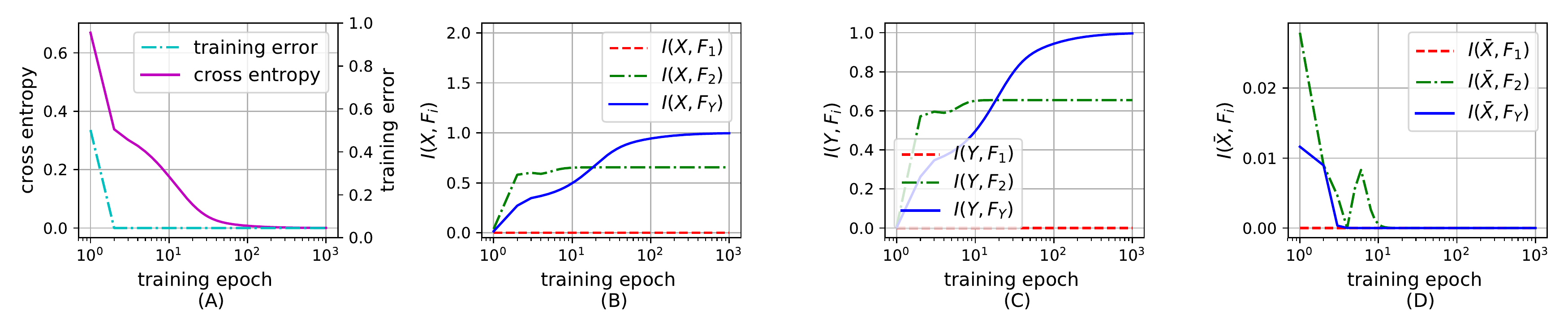}}
\caption{(A) visualizes the variation of the training error and the cross entropy loss of MLP3 during training.
(B)-(D) visualize the variations of $I(\boldsymbol{X}, F_i)$,  $I(Y, F_i)$, and $I(\bar{X}, F_i)$ over all the layers of MLP3 during training, respectively.}
\label{Img_info_flow_(1_4_2)}
\end{figure*}


Compared to MLP1, MLP3 only has one neuron in $\boldsymbol{f}_1$, which makes MLP3 to spend more than 100 epochs in minimizing the cross entropy loss to zero in Figure \ref{Img_info_flow_(1_4_2)}(A).
More importantly, it significantly changes the information flow in MLP3.
The probability space $(\Omega_{F_1}, \mathcal{T}, P_{F_1})$ indicates that the single neuron only defines one possible outcome with $100\%$ occurring probability, thus $\boldsymbol{f}_1$ becomes a deterministic function and cannot transfer information to $\boldsymbol{f}_2$ and $\boldsymbol{f}_Y$ in the forward direction, i.e., the second and third blue arrows are blocked in Figure \ref{Img_mlp_information_flow1}.
As a result, the information of $\boldsymbol{X}$ and $Y$ can flow into MLP3 in the backward direction.

Figure \ref{Img_info_flow_(1_4_2)}(B)-\ref{Img_info_flow_(1_4_2)}(D) visualize the information flow in MLP3 and demonstrate the above theoretical discussion.
First, we observe $I(\boldsymbol{X}, F_1) = I(Y, F_1) = 0$, which validates $\boldsymbol{f}_1$ being a deterministic function.
Second, the information flow of $\boldsymbol{X}$ and $Y$ in MLP3 satisfy ${\textstyle I(\boldsymbol{X}, F_Y) \geq I(\boldsymbol{X}, F_2) \geq I(\boldsymbol{X}, F_1)}$ and ${\textstyle I(Y, F_Y) \geq I(Y, F_2) \geq I(Y, F_1)}$ in most training epochs, respectively, which validates the information of $\boldsymbol{X}$ and $Y$ can flow into MLP3 in the backward direction.
Third, $I(\bar{X}, F_i)$ being very close to zero validates that all the information of each layer stems from $Y$ based on Equation \ref{xyxc}.




In summary, this section demonstrates the proposed information theoretic explanations in Section \ref{limitation_IB} and \ref{info_flowyx}.
First, we observe three different information flows of $\boldsymbol{X}$ in the three MLPs, i.e., the information flow of $\boldsymbol{X}$ cannot satisfy any DPI in MLPs.
Second, the information flow of $Y$ in the tree MLPs has backward direction, i.e., it satisfy the DPI ${\textstyle I(Y, F_Y) \geq I(Y, F_2) \geq I(Y, F_1)}$, especially after the cross entropy loss decreases to zero. 
Third, we demonstrate that MLPs cannot satisfy the DPI derived from IB (Equation \ref{dpi_dnn}), especially the information of $\boldsymbol{X}$ can only flow into MLP3 in the backward direction.

To further demonstrate the proposed information theoretic explanations, the next section compares the proposed mutual information estimators (Section \ref{mi_calculation}) to commonly used mutual information estimators based on two non-parametric inference methods, i.e., empirical distributions \citep{DNN-information} and Gaussian KDE \citep{IP-argue}. 
We use the same experimental methods and synthetic dataset as before to show the information flow of $\boldsymbol{X}$ and $Y$ in MLP1 and MLP2 based on the three mutual information estimators.

\begin{figure*}
\centering
\centerline{\includegraphics[scale=0.5]{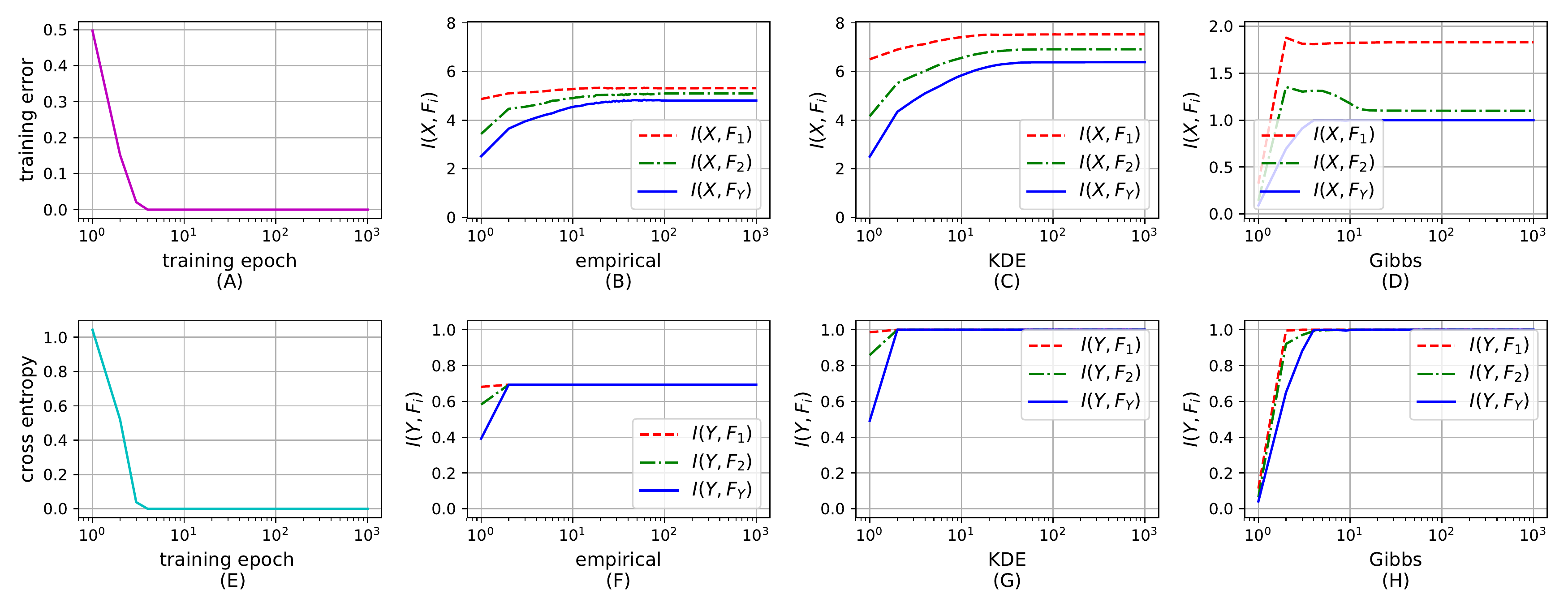}}
\caption{(A) and (E) visualize the variation of the training error and the cross entropy loss of MLP1 during training, respectively.
(B), (C), and (D) visualize the variation of $I(\boldsymbol{X}, F_i)$ over all the layers in MLP1 based on empirical distributions, Gaussian KDE, and Gibbs distribution, respectively.
(F), (G), and (H) visualize the variation of $I(Y, F_i)$ over all the layers in MLP1 based on empirical distributions, Gaussian KDE, and Gibbs distribution, respectively.}
\label{Img_info_flow_(8_6_2)_relu}
\end{figure*}

\begin{figure*}
\centering
\centerline{\includegraphics[scale=0.5]{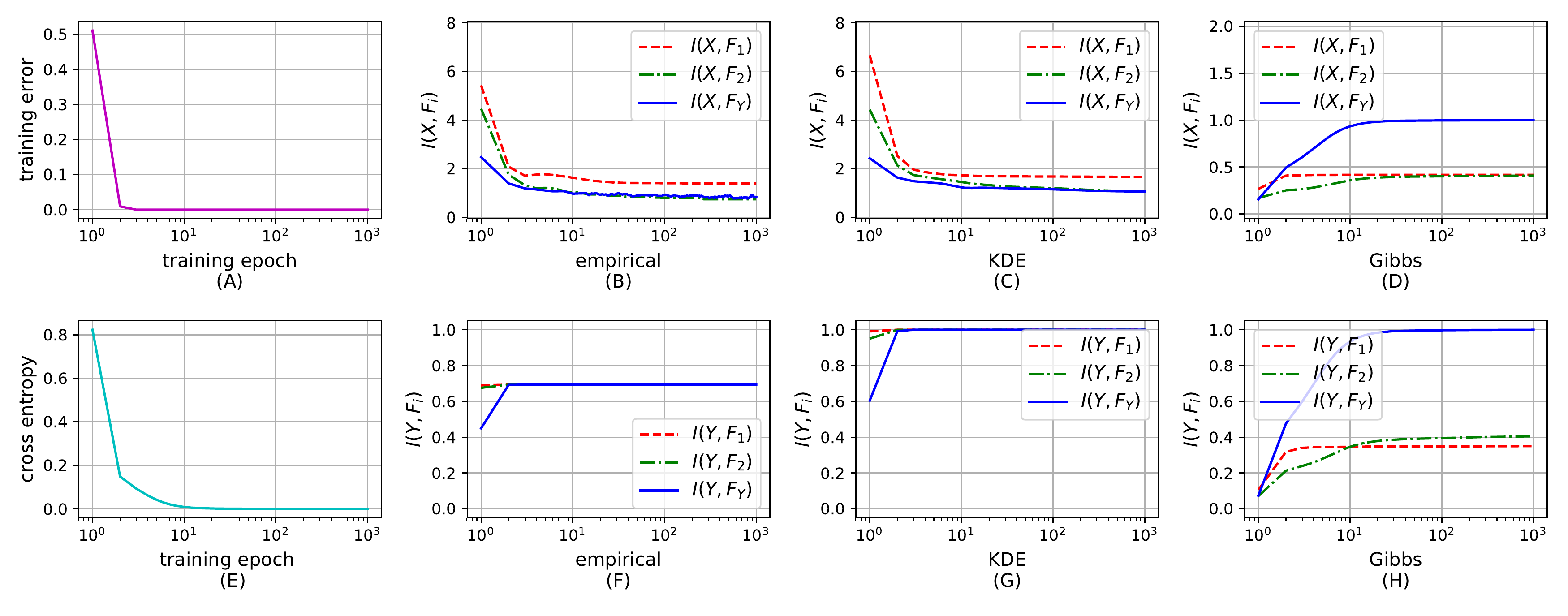}}
\caption{(A) and (E) visualize the variation of the training error and the cross entropy loss of MLP2 during training, respectively.
(B), (C), and (D) visualize the variation of $I(\boldsymbol{X}, F_i)$ over all the layers in MLP2 based on empirical distributions, Gaussian KDE, and Gibbs distribution, respectively.
(F), (G), and (H) visualize the variation of $I(Y, F_i)$ over all the layers in MLP2 based on empirical distributions, Gaussian KDE, and Gibbs distribution, respectively.
}
\label{Img_info_flow_(8_6_2)_tanh}
\end{figure*}

\subsubsection{The comparison with existing methods}

Figure \ref{Img_info_flow_(8_6_2)_relu}(B)-(C) show $I(\boldsymbol{X}, F_1) > 2$ in MLP1 based on empirical distributions and KDE, which contradicts the fact that the synthetic dataset only has 2 bits information. 
As a result, the two estimators cannot correctly estimate $I(\boldsymbol{X}, F_1)$.
Figure \ref{Img_info_flow_(8_6_2)_relu}(F)-(G) show $I(Y, F_Y) = 0.8$ and $I(Y, F_Y) = 1$ based on empirical distributions and KDE, respectively.
It contradicts that $I(Y, F_Y) = H(Y)$ if the cross entropy loss becomes zero (Section \ref{generalization_it}).
Specifically, Figure \ref{Img_info_flow_(8_6_2)_relu}(F) show $I(Y, F_Y) < H(Y)$ after the cross entropy loss decreases to zero, and Figure \ref{Img_info_flow_(8_6_2)_relu}(G) show $I(Y, F_Y) = H(Y)$ before the cross entropy loss decreases to zero.

Section \ref{act_function_pf} shows that Tanh cannot guarantee a precise Gibbs measure and Section \ref{act_function_mi} derives that $\boldsymbol{f}_1$ with Tanh only contains 0.47 bit information of the synthetic dataset. However, Figure \ref{Img_info_flow_(8_6_2)_tanh}(B)-(C) show $I(\boldsymbol{X}, F_1) > 1$ based on empirical distributions and KDE.
As a result, the two mutual information estimators cannot correctly estimate $I(\boldsymbol{X}, F_1)$ in MLP2.
In addition, Figure \ref{Img_info_flow_(8_6_2)_tanh}(F)-(G) demonstrate the same limitation of the two estimators for estimating $I(Y, F_Y)$ in MLP2 as in MLP1.

In summary, the mutual information estimators based on empirical distributions and KDE cannot correctly estimate the information flow of $\boldsymbol{X}$ and $Y$ in MLP1 and MLP2.

\begin{figure*}
	\begin{minipage}[b]{0.99\linewidth}
  		\centering
  		\centerline{\includegraphics[scale=0.5]{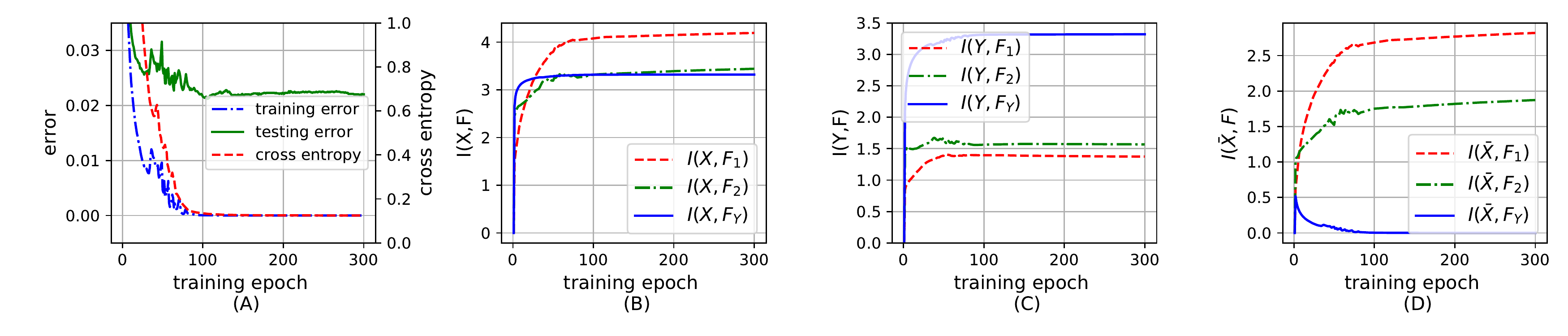}}
	\end{minipage}
	\vfill	
	\begin{minipage}[b]{0.99\linewidth}
		\centering
		\centerline{\includegraphics[scale=0.5]{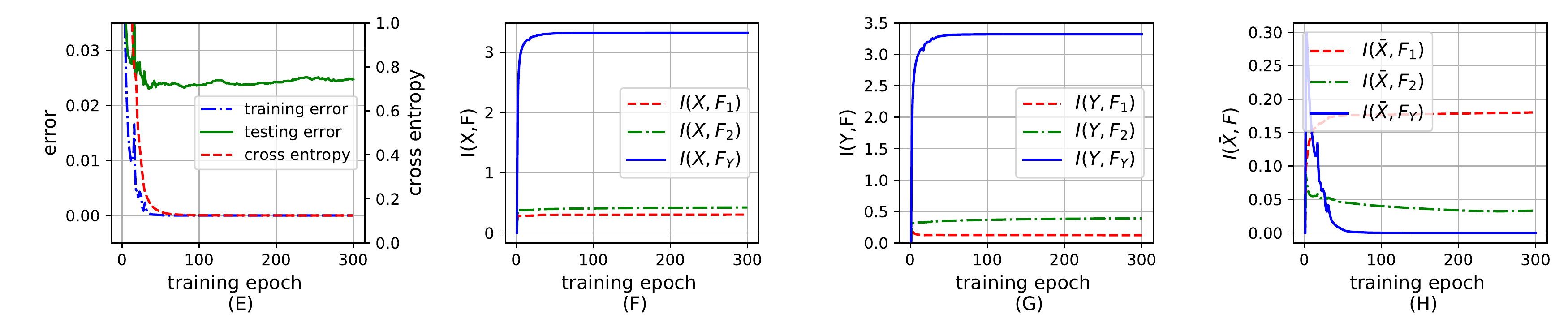}}		
	\end{minipage}
	\vfill	
	\begin{minipage}[b]{0.99\linewidth}
		\centering
		\centerline{\includegraphics[scale=0.5]{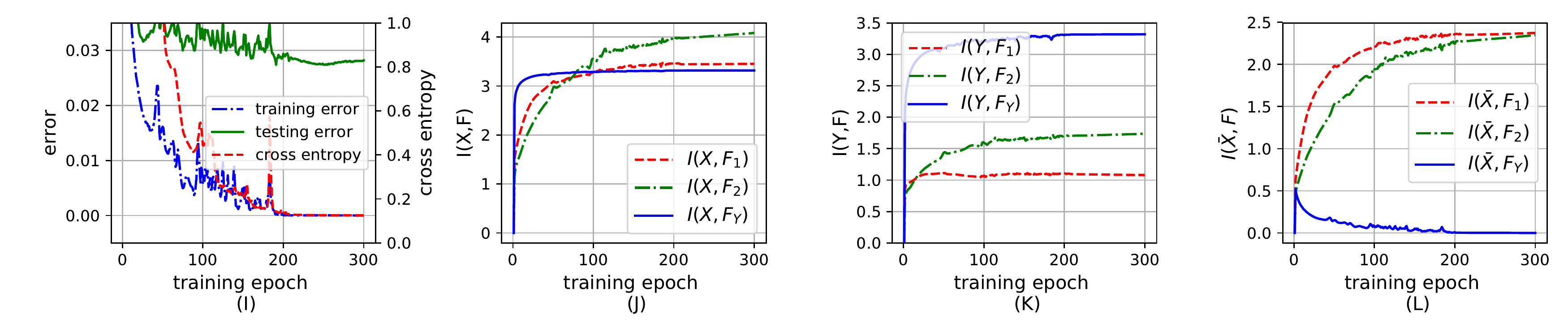}}		
	\end{minipage}
\caption{(A), (E), and (I) visualize the variation of the training/testing error and the cross entropy loss of MLP4, MLP5 and MLP6 during training, respectively.
(B), (F), and (J) visualize the variation of $I(\boldsymbol{X}, F_i)$ over all the layers in MLP4, MLP5, and MLP6, respectively.
(C), (G), and (K) visualize the variation of $I(Y, F_i)$ over all the layers in MLP4, MLP5, and MLP6, respectively.
(D), (H), and (L) visualize the variation of $I(\bar{X}, F_i)$ over all the layers in MLP4, MLP5, and MLP6, respectively.}
\label{if_mlp_mnist}
\end{figure*} 

\subsection{The simulations on the benchmark datasets}

In this section, we use MNIST dataset to demonstrate the proposed explanations for MLPs: (i) the information flow of $\boldsymbol{X}$ and ${Y}$ in MLPs (Section \ref{info_flowyx} and \ref{limitation_IB}) and (ii) the information theoretic explanations for generalization (Section \ref{generalization_it}).
In addition, Appendix \ref{it_mlps_fmnist} presents extra experiments based on more complex MLPs on Fashion-MNIST dataset to further validate the proposed explanations for MLPs.

\subsubsection{The information flow in the MLPs}

We design three MLPs, i.e., MLP4, MLP5, and MLP6, and their differences are summarized in Table \ref{MLP3_MNIST}.
All the weights of the MLPs are randomly initialized by truncated normal distributions.
We still choose the Adam method to learn the weights of the MLPs on MNIST dataset over 300 epochs with the learning rate 0.001, and use the same method as Section \ref{if_mlp_synthetic} to derive $I(\boldsymbol{X}, F_i)$,  $I(Y, F_i)$, and $I(\bar{X}, F_i)$ based on Equation (\ref{mi_x}), (\ref{mi_y}), (\ref{xyxc}), respectively.

\begin{table}
\caption{The number neurons(nodes) of each layer and the activation functions in each MLP.}
\label{MLP3_MNIST}
\vskip 0.1in
\begin{center}
\begin{small}
\begin{threeparttable}
\begin{tabular}{cccccc}
\toprule
 & $\boldsymbol{x}$ & $\boldsymbol{f}_1$ & $\boldsymbol{f}_2$ & $\boldsymbol{f}_Y$ & $\sigma(\cdot)$ \\
\midrule
MLP4 & 784 & 96 & 32 & 10  & ReLU\\
MLP5 & 784 & 96 & 32 & 10 & Tanh \\
MLP6 & 784 & 32 & 96 & 10 & ReLU \\
\bottomrule
\end{tabular}
\end{threeparttable}
\end{small}
\end{center}
\vskip -0.1in
\end{table}

The information flow in the MLPs on MNIST dataset is consistent with the results on the synthetic dataset.
More specifically, Figure \ref{if_mlp_mnist}(B), \ref{if_mlp_mnist}(F) and \ref{if_mlp_mnist}(J) visualize three different information flows of $\boldsymbol{X}$ in MLP4, MLP5, and MLP6, respectively.
It confirms that the information flow of $\boldsymbol{X}$ in MLPs does not satisfy any DPI.
In addition, Figure \ref{if_mlp_mnist}(C), \ref{if_mlp_mnist}(G) and \ref{if_mlp_mnist}(K) demonstrate that the information flow $Y$ satisfies $I(Y, F_Y) \geq I(Y, F_2) \geq I(Y, F_1)$ in all the tree MLPs.
The experiment also demonstrates that IB cannot correctly explain the information flow of $\boldsymbol{X}$ and ${Y}$ in MLPs, because they cannot satisfy the DPIs (Equation \ref{dpi_dnn}) derived from IB in Figure \ref{if_mlp_mnist}(B, C), \ref{if_mlp_mnist}(F, G) and \ref{if_mlp_mnist}(J, K).


Figure \ref{if_mlp_mnist}(D), \ref{if_mlp_mnist}(H) and \ref{if_mlp_mnist}(J) demonstrate that the information flow $\bar{X}$ in all the tree MLPs satisfies $I(\bar{X}, F_1) \geq I(\bar{X}, F_2) \geq I(\bar{X}, F_Y)$.
The next section will demonstrate that the $I(\bar{X}, F_1)$ can measure the generalization of MLPs along with two variables: (i) the number of neurons, (ii) and the number of training samples.

\subsubsection{The information theoretic explanation for the generalization performance of MLPs}

\begin{figure*}
\centering
\includegraphics[scale=0.6]{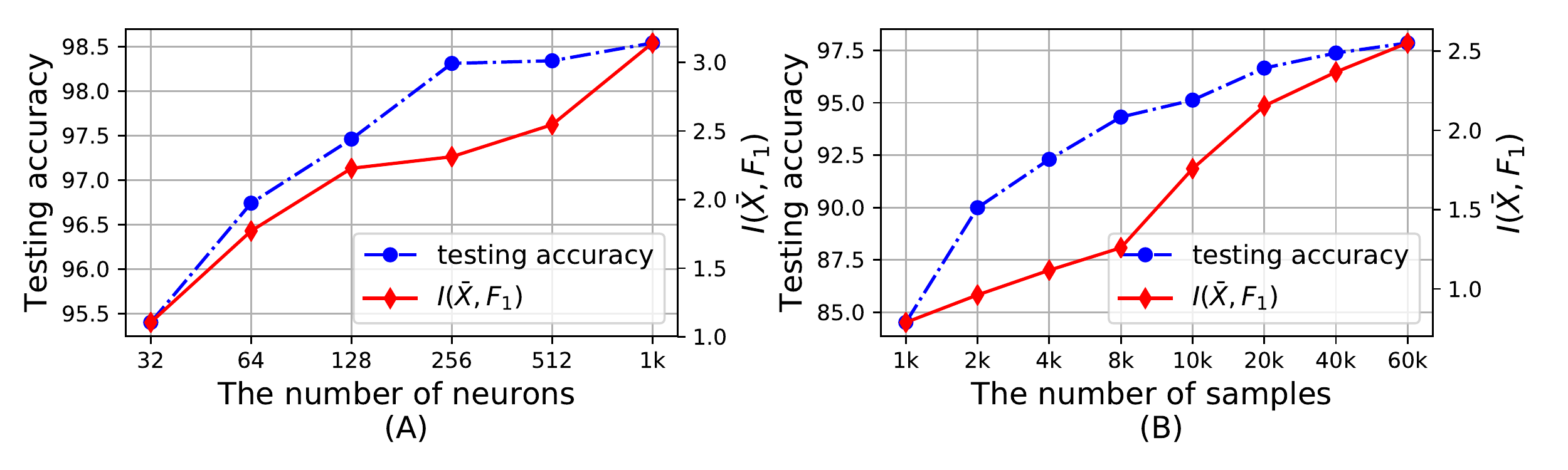}
\caption{\small{
(A) shows the variation of the testing accuracy and $I(\bar{X}, F_1)$ given different MLPs with different number of neurons.
(B) shows the variation of the testing accuracy and $I(\bar{X}, F_1)$ given different number of training samples.
}}
\label{Img_mixbar_neurons_samples}
\end{figure*}


First, $I(\bar{X}, F_1)$ can measure the generalization of MLPs with different numbers of neurons.
In general, a MLP with more neurons would have better generalization performance, thus $I(\bar{X},F_1)$ of the MLP should be larger.
We design six different $\text{MLPs} = \{\boldsymbol{x}, \boldsymbol{f}_1, \boldsymbol{f}_2, \boldsymbol{f}_Y\}$, of which the two hidden layers have the same number of neurons with the same ReLU activation function, but different MLPs have different numbers of neurons, i.e., 32, 64,128, 256, 512, 1024.
After the MLPs achieve 100\% training accuracy on MNIST dataset, we observe a positive correlation with the testing accuracy and $I(\bar{X},F_1)$ in Figure \ref{Img_mixbar_neurons_samples}(A).

Second, $I(\bar{X}, F_1)$ can measure the generalization of MLPs with different numbers of training samples.
In general, a MLP with larger number of training samples would have better generalization, thus $I(\bar{X},F_1)$ of the MLP should be larger.
We generate 8 different training sets with different number of MNIST training samples and train MLP4 on the 8 training sets.
After MLP4 achieves 100\% training accuracy on the 8 training sets, we also observe a positive correlation with the testing accuracy and $I(\bar{X},F_1)$ in Figure \ref{Img_mixbar_neurons_samples}(B).

In summary, $I(\bar{X}, F_1)$ demonstrates positive correlation with the testing error of MLPs, thus we conclude that $I(\bar{X}, F_1)$ can be regarded as a criterion for the generalization of MLPs along with two variables: (i) the number of neurons, (ii) and the number of training samples.The experiment shows potential for explaining the generalization of general DNNs from perspective of information theory, we leave a rigorous study of this as future work.

\section{Conclusions}
\label{conclusions}

In this paper, we introduce a probabilistic representation for improving the information theoretic interpretability.
The probabilistic representation for MLPs includes three parts. 

First, we demonstrate that the activations being \textit{i.i.d.} is not valid for all the hidden layers of MLPs.
As a result, the mutual information estimators based on non-parametric inference methods, e.g., empirical distributions and Kernel Density Estimate (KDE), are invalid for measuring the mutual information in MLPs because the prerequisite of these non-parametric inference methods is samples being \textit{i.i.d.}.

Second, we define the probability space $(\Omega_F, \mathcal{T}, P_F)$ for a fully connected layer $\boldsymbol{f}$ with $N$ neurons given the input $\boldsymbol{x}$.
Let the experiment be $\boldsymbol{f}$ extracting a single feature of $\boldsymbol{x}$, the sample space $\Omega_F$ consists of $N$ possible outcomes (i.e., features), and each outcome is defined by the weights of each neuron; 
the event space $\mathcal{T}$ is the $\sigma$-algebra; 
and the probability measure $P_F$ is a Gibbs measure for quantifying the probability of each outcome occurring the experiment. 

Third, we propose probabilistic explanations for MLPs and the back-propagation training: (i) the entire architecture of MLPs formulates a Gibbs distribution based on the Gibbs distribution $P_F$ for each layer;
and (ii) the back-propagation training aims to optimize the sample space of all the layers of MLPs for modeling the statistical connection between the input $\boldsymbol{x}$ and the label $\boldsymbol{y}$, because the weights of each layer define the sample space $\Omega_F$.

To the best of our knowledge, most existing information theoretic explanations for MLPs lack a solid probabilistic foundation.
It not only weakens the validity of the information theoretic explanations but also could derive incorrect explanations for MLPs. 
To resolve the fundamental issue, we first introduce the probabilistic representation for MLPs, and then improve the information theoretic interpretability of MLPs in three aspects.

Above all, we explicitly define the random variable of $\boldsymbol{f}$ as ${F}: \Omega_F \rightarrow E$ based on $(\Omega_F, \mathcal{T}, P_F)$. 
Since $\Omega_F$ is discrete, $E$ denotes discrete measurable space. Hence, $F$ is a discrete random variable and $H(F) < \infty$.
In other words, we resolve the controversy regarding $F$ being discrete or continuous.

Furthermore, the probabilistic explanation for the back-propagation training indicates that $\Omega_F$ depends on both $\boldsymbol{x}$ and $y$, thereby $F$ depending on both $\boldsymbol{X}$ and $Y$.
It contradicts the probabilistic assumption of IB, i.e., $F$ is independent on $Y$ given $\boldsymbol{X}$.
As a result, the information flow of $\boldsymbol{X}$ and $Y$ in MLPs does not satisfy IB if we take into account the back-propagation training.

In addition, we demonstrate that the performance of a MLP depends on the mutual information between the MLP and the input $\boldsymbol{X}$, i.e., $I(\boldsymbol{X}, \text{MLP})$.
Specifically, we prove all the information of $Y$ steming from $\boldsymbol{X}$, i.e., $H(Y) = I(\boldsymbol{X}, Y)$ (the relation is visualized by the Venn diagram in Figure \ref{Img_mlp_information_venn}), thus $I(\boldsymbol{X}, \text{MLP})$ can be divided into two parts $I(Y, \text{MLP})$ and $I(\bar{X}, \text{MLP})$, where $\bar{X} = Y^c \cap \boldsymbol{X}$ denotes the relative complement $Y$ in $\boldsymbol{X}$.
We show that the training accuracy of the MLP depends on $I(Y, \text{MLP})$, and the generalization of the MLP depends on $I(\bar{X}, \text{MLP})$.

It is noteworthy that we design a synthetic dataset to fully demonstrate the proposed probabilistic representation and information theoretic explanations for MLPs.
Compared to all the existing information theoretical explanations merely using benchmark datasets for validation, the synthetic dataset enables us to demonstrate the proposed information theoretic explanations clearly and comprehensively, because all the features of the synthetic dataset are known and much simpler than benchmark dataset.

The proposed information theoretic explanations for MLPs provides a novel viewpoint to understand the generalization of MLPs.
It deserves more efforts as future research.
First, since the cross entropy loss only guarantees the performance of MLPs on the training dataset, incorporating $I(\bar{X},F_1)$ into the cross entropy loss could be a promising approach to improve the generalization performance of MLPs. 
Second, we are planning to extend the information theoretic explanations for generalization to general DNNs, which could shed light on understanding the generalization of DNNs. 


\pagebreak

\appendix

\section{The necessary conditions for activations being \textit{i.i.d.}}
\label{appendix:necess_iid}

\subsection{The necessary conditions for activations being independent}

The necessary condition for two random variables $A$ and $B$ being independent is that they are uncorrelated, namely the covariance $Cov(A, B) = 0$.
Therefore, the necessary condition for $\{G_{2k} \}_{k=1}^K$ being independent is that $\forall (k, k') \in S_1 = \{(k, k') \in \mathbb{Z}^2 |k \neq k', 1 \leq k \leq K, 1 \leq k' \leq K \}$, $Cov(G_{2k}, G_{2k'}) = 0$, which can be formulated as 
\begin{equation} 
{
Cov(\sum_{n=1}^N\omega^{(2)}_{nk}F_{1n} + b_{2k}, \sum_{n'=1}^N\omega^{(2)}_{n'k'}F_{1n'} + b_{2k'}) = 0.
}
\end{equation}

Since the covariance between a random variable and a constant are zero, namely $Cov(X, c) = 0$, we derive
\begin{equation} 
\begin{split}
Cov(G_{2k}, G_{2k'}) = Cov(\sum_{n=1}^N\omega^{(2)}_{nk}F_{1n}, \sum_{n'=1}^N\omega^{(2)}_{n'k'}F_{1n'}).
\end{split}
\end{equation}

Based on $Cov(X, Y+Z) = Cov(X, Y) + Cov(X, Z)$, $Cov(G_{2k}, G_{2k'})$ can be extended as
\begin{equation} 
\begin{split}
Cov(G_{2k}, G_{2k'}) = &\sum_{n=1}^N \omega^{(2)}_{nk}\omega^{(2)}_{nk'} Var(F_{1n})\\
&+ \sum_{n \neq n'}\omega^{(2)}_{nk}\omega^{(2)}_{n'k'}Cov(F_{1n}, F_{1n'}).
\end{split}
\end{equation}

Assuming $\{F_{1n} \}_{n=1}^N$ are \textit{i.i.d.} and $\{f_{1n}\}_{n=1}^N \sim P(F_1)$, we have $Cov(F_{1n}, F_{1n'}) = 0$, thus we have
\begin{equation} 
\begin{split}
Cov(G_{2k}, G_{2k'}) = Var(F_{1})\sum_{n=1}^N \omega^{(2)}_{nk}\omega^{(2)}_{nk'}.
\end{split}
\end{equation}
Since $Var(F_{1}) > 0$, the necessary condition for $\{G_{2k} \}_{k=1}^K$ being independent can be formulated as $\forall (k, k') \in S_1$,
\begin{equation}
\label{betabeta}
\begin{split}
\sum_{n=1}^N \omega^{(2)}_{nk}\omega^{(2)}_{nk'} = 0.
\end{split}
\end{equation}

Based on the theorem in Appendix \ref{function_independent}, we can derive that $\{G_{2k} \}_{k=1}^K$ being independent is equivalent to $\{F_{2k} \}_{k=1}^K$ being independent as long as the activation function $\sigma_2(\cdot)$ is invertible.
In other words, if $\sigma_2(\cdot)$ is invertible, the necessary condition for $\{F_{2k} \}_{k=1}^K$ being independent is the same as the necessary condition for $\{G_{2k} \}_{k=1}^K$ being \textit{i.i.d.}.

In summary, if the activations of a hidden layer are independent in the context of frequentist probability, the weights of the hidden layer must satisfy Equation \ref{betabeta} given the assumption that the inputs of the hidden layer are \textit{i.i.d.}\ and the activation function is invertible.

\pagebreak
\subsection{The necessary conditions for activations being identically distributed}

The necessary condition for $\{G_{2k} \}_{k=1}^K$ being identically distributed is that $\forall (k, k') \in S_1$, we have
\begin{equation} 
\begin{split}
E(G_{2k}) = E(G_{2k'}),
\end{split}
\end{equation}
where $E(\cdot)$ denotes the expectation.

Since $G_{2k} = \sum_{n=1}^N\omega^{(2)}_{nk}F_{1n} + b_{2k}$, we can derive
\begin{equation} 
\begin{split}
E(G_{2k}) & = \sum_{n=1}^N \omega^{(2)}_{nk}E(F_{1n}) + b_{2k}.
\end{split}
\end{equation}

Assuming $\{F_{1n} \}_{n=1}^N$ are \textit{i.i.d.}\ and $\{f_{1n}\}_{n=1}^N \sim P(F_1)$, we have $E(F_{1n}) = E(F_{1})$.
Hence, we can further derive
\begin{equation}
\begin{split}
E(G_{2k}) & = E(F_{1})\sum_{n=1}^N \omega^{(2)}_{nk} + b_{2k}.
\end{split}
\end{equation}

Based on $E(G_{2k}) = E(G_{2k'})$, we can derive
\begin{equation} 
\label{identical}
\begin{split}
-E(F_{1})\sum_{n=1}^N (\omega^{(2)}_{nk} -  \omega^{(2)}_{nk'}) = b_{2k} - b_{2k'}.
\end{split}
\end{equation}

We assume that $\sigma_2(\cdot)$ is strictly increasing and differentiable, thus $\sigma_2(\cdot)$ is invertible and its inverse $\sigma_2^{-1}(\cdot)$ is also strictly increasing.
As a result, the cumulative distribution function of $F_{2k}$ can be expressed as 
\begin{equation} 
\label{neuron_cumulative} 
\begin{split}
\Phi_{F_{2k}}(f) & = \phi(F_{2k} \leq f) \\
 & = \phi(\sigma_1(G_{2k}) \leq f) \\
 & = \phi(G_{2k} \leq \sigma_1^{-1}(f)) \\
  & = \Phi_{G_{2k}}(\sigma_1^{-1}(f)). \\
\end{split}
\end{equation}
where $\phi(F_{2k} \leq f)$ is the probability of $F_{2k}$ takes on a value less than or equal to $f$. 
Subsequently, we can obtain
\begin{equation} 
\label{neuron_density} 
\begin{split}
P_{F_{2k}}(f) & = \frac{\partial \Phi_{F_{2k}}(f)}{\partial f} =  \frac{\partial \Phi_{G_{2k}}(\sigma_2^{-1}(f))}{\partial f} \\
& = P_{G_{2k}}(\sigma_2^{-1}(f))\frac{\partial \sigma_2^{-1}(f)}{\partial f}. \\
\end{split}
\end{equation}
Equation \ref{neuron_density} indicates that if $\sigma_2(\cdot)$ is strictly increasing and differentiable and $\{G_{2k} \}_{k=1}^K$ are identically distributed, then $\{F_{2k} \}_{k=1}^K$ are identically distributed as well.

In summary, if the activations of a hidden layer are identically distributed in the context of frequentist probability, the weights and the biases of the hidden layer must satisfy Equation \ref{identical} under the assumption that inputs of the layer are \textit{i.i.d.}\ and the activation functions are strictly increasing and differentiable.

\pagebreak
\subsection{Conclusion}

In summary, assuming $\{F_{1n} \}_{n=1}^N$ are \textit{i.i.d.}\ and $\{f_{1n}\}_{n=1}^N \sim P(F_1)$, the necessary conditions for the activations $\{F_{2k} \}_{k=1}^K$, being \textit{i.i.d.} can be summarized as
\begin{equation} 
\label{condition_indep_a}
\begin{split}
\forall (k, k') \in S_1 = \{(k, k') \in \mathbb{Z}^2 |&{\scriptstyle k \neq k', 1 \leq k \leq K, 1 \leq k' \leq K} \} \\
\sum_{n=1}^N \omega^{(2)}_{nk}&\omega^{(2)}_{nk'} = 0, \\
-E(F_{1})\sum_{n=1}^N (\omega^{(2)}_{nk} &-  \omega^{(2)}_{nk'}) = b_{2k} - b_{2k'},
\end{split}
\end{equation}
\begin{equation}
\label{condition_act}
\sigma_2(\cdot) \text{ is strictly increasing and differentiable.}
\end{equation}

Equation \ref{condition_indep_a} shows the necessary conditions for $\{G_{2k}\}_{k=1}^K$ being independent and identically distributed.
Equation \ref{condition_act} specifies the condition of the activation function $\sigma_2(\cdot)$ such that if $\{G_{2k}\}_{k=1}^K$ are \textit{i.i.d.}, then $\{F_{2k}\}_{k=1}^K$ are also \textit{i.i.d.}.
The invertible condition is not required here because strictly increasing and differentiable imply it.
It is important to note that the necessary conditions hold for arbitrary fully connected layers as long as we properly change the superscript of $\omega^{(2)}_{nk}$ and the subscript of $b_{2k}$.

\subsection{functions of independent random variables are independent}
\label{function_independent}

\textbf{Theorem:} Assuming ${X}$ and ${Y}$ are independent random variables on a probability space $(\Omega, \mathcal{T}, P)$.
Let $g$ and $h$ be real-valued functions defined on the codomains of $\boldsymbol{X}$ and $\boldsymbol{Y}$, respectively.
Then $g({X})$ and $h({Y})$ are independent random variables.

\textbf{Proof:}
Let $\boldsymbol{A} \subseteq \mathbb{R}$ and $\boldsymbol{B} \subseteq \mathbb{R}$ be the range of $g$ and $h$, the joint distribution between $g({X})$ and $h({Y})$ can be formulate as ${P}(g(\boldsymbol{X}) \in \boldsymbol{A}, h(\boldsymbol{Y}) \in \boldsymbol{B})$.
Let $g^{-1}(\boldsymbol{A})$ and $h^{-1}(\boldsymbol{B})$ denote the preimages of $\boldsymbol{A}$ and $\boldsymbol{B}$, respectively, we have
\begin{equation} 
\begin{split}
{P}(g({X}) \in \boldsymbol{A}, h({Y}) \in \boldsymbol{B}) &= {P}({X} \in g^{-1}(\boldsymbol{A}), {Y} \in h^{-1}(\boldsymbol{B}))
\end{split}
\end{equation}
Based on the definition of independence, we can derive that 
\begin{equation} 
\begin{split}
{P}(g({X}) \in \boldsymbol{A}, h({Y}) \in \boldsymbol{B}) & = {P}({X} \in g^{-1}(\boldsymbol{A})){P}({Y} \in h^{-1}(\boldsymbol{B})) \\
& = {P}(g({X}) \in \boldsymbol{A}){P}(h({Y}) \in \boldsymbol{B}) \\
\end{split}
\end{equation}
Based on the definition of preimage, we can derive that
\begin{equation} 
\begin{split}
{P}(g({X}) \in \boldsymbol{A}, h({Y}) \in \boldsymbol{B}) & = {P}(g({X}) \in \boldsymbol{A}){P}(h({Y}) \in \boldsymbol{B}) \\
\end{split}
\end{equation}
Therefore, $g({X})$ and $h({Y})$ are independent random variables.

\pagebreak

\section{Activations are not \textit{i.i.d.} in more complex MLPs on the Fashion-MNIST dataset}
\label{MLP_IID}

\begin{figure*}
\centering
\includegraphics[scale=0.55]{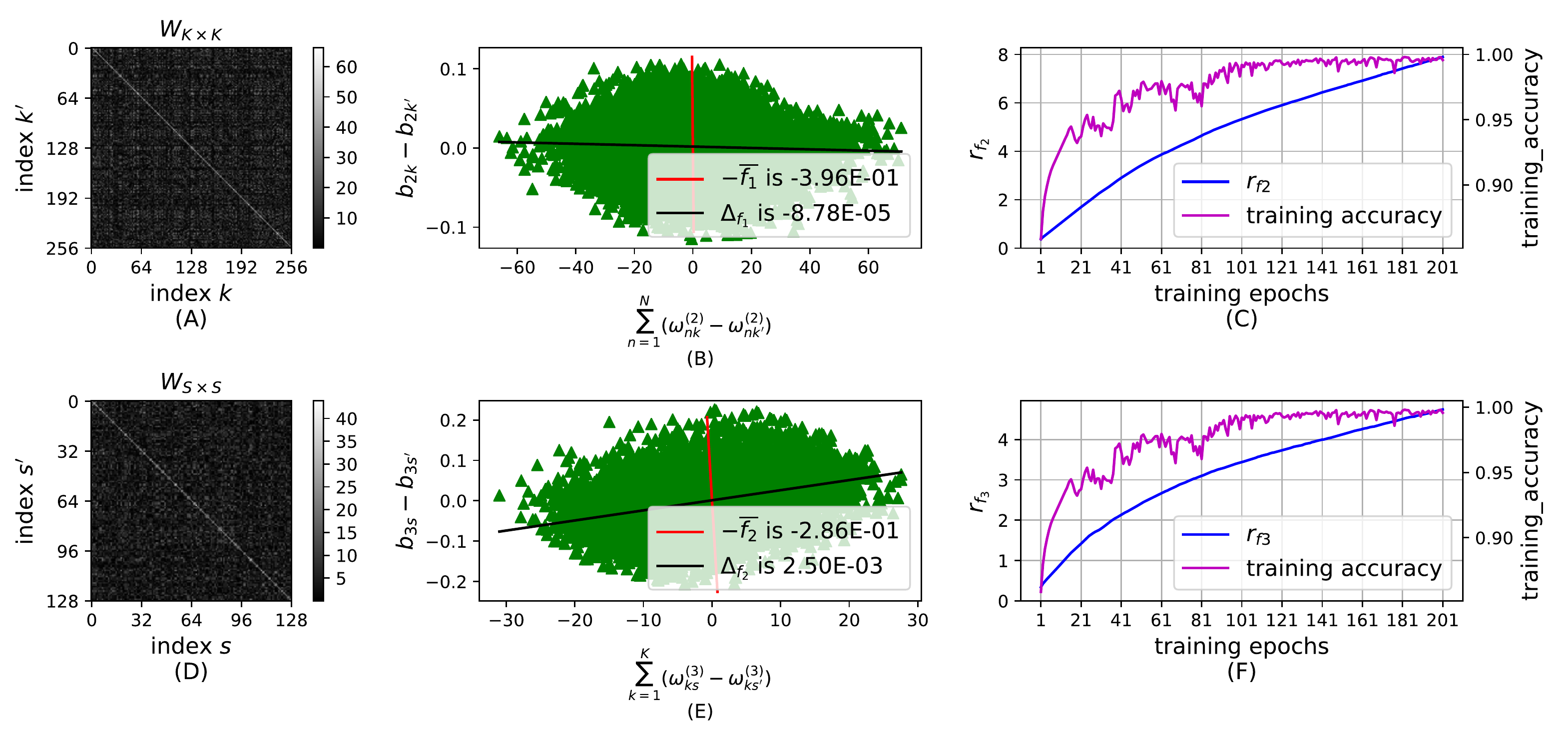}
\caption{\small{
(A) $W_{K\times K}$ contains the value of $|\sum_{n=1}^N \omega^{(2)}_{nk}\omega^{(2)}_{nk'}|$ for all the neurons in $\boldsymbol{f_2}$.
(B) The green triangles show all the samples $[\sum_{n=1}^N (\omega^{(2)}_{nk} - \omega^{(2)}_{nk'}), b_{2k}  - b_{2k'}]$, the black line shows the linear regression result based on these samples, and the red line shows the linear relation indicated by the slope $-\overline{\boldsymbol{f_1}} \approx -E(F_{1})$.
(C) The blue curve and the magenta curve show the variation of $r_{\boldsymbol{f_2}} = \frac{1}{\sum_{k=1}^K\sum_{k'=1}^k 1}|\sum_{n=1}^N \omega^{(2)}_{nk}\omega^{(2)}_{nk'}|$ and the training accuracy in 201 training epochs, respectively.
(D) $W_{S\times S}$ contains the value of $|\sum_{k=1}^K \omega^{(3)}_{ks}\omega^{(3)}_{ks'}|$ for all the neurons in $\boldsymbol{f_3}$.
(E) The green triangles show all the samples $[\sum_{k=1}^K (\omega^{(3)}_{ks} - \omega^{(3)}_{ks'}), b_{3s}  - b_{3s'}]$, the black line shows the linear regression result based on these samples, and the red line shows the linear relation indicated by the slope $-\overline{\boldsymbol{f_2}} \approx -E(F_{2})$.
(F) The blue curve and the magenta curve show the variation of $r_{\boldsymbol{f_3}} = \frac{1}{\sum_{s=1}^S\sum_{s'=1}^s 1}|\sum_{s=1}^S \omega^{(3)}_{ks}\omega^{(3)}_{ks'}|$ and the training accuracy in 201 training epochs, respectively.
}}
\label{fig_mlp_iid3}
\end{figure*}

In this section, we demonstrate that activations cannot satisfy the necessary conditions in a more complex MLP on the Fashion-MNIST dataset, thus activation being \textit{i.i.d.}\ is not valid for all the fully connected layers of the MLP.

To check if activations satisfy the necessary conditions, we specify a $\text{MLP} = \{\boldsymbol{x}, \boldsymbol{f}_1, \boldsymbol{f}_2, \boldsymbol{f}_3, \boldsymbol{f}_Y\}$ for classifying the Fashion-MNIST dataset \citep{xiao2017fashion}.
The dimension of each Fashion-MNIST image is $28 \times 28$, thus the number of the input nodes is $M = 784$.
In addition, $\boldsymbol{f_1}$, $\boldsymbol{f_2}$, and $\boldsymbol{f_3}$ have $N = 512$, $K= 256$, and $S = 128$ neurons, respectively, and $\boldsymbol{f_Y}$ has $L = 10$ nodes.
All hidden layers choose the sigmoid function, which satisfies the third necessary condition (Equation \ref{condition_indep}), thus we only need to examine the first two necessary conditions.

After the training accuracy is very close to $100\%$, we obtain $\omega^{(2)}_{nk}$ and $\omega^{(3)}_{ks}$, and construct two matrixes $W_{K \times K}$ and $W_{S \times S}$ to contain $|\sum_{n=1}^N \omega^{(2)}_{nk}\omega^{(2)}_{nk'}|$ and $|\sum_{k=1}^K \omega^{(3)}_{ks}\omega^{(3)}_{ks'}|$ for each activation in $\boldsymbol{f_2}$ and $\boldsymbol{f_3}$, respectively.
Figure \ref{fig_mlp_iid2}(A) and \ref{fig_mlp_iid2}(D) show that $\sum_{n=1}^N \omega^{(2)}_{nk}\omega^{(2)}_{nk'}$  and $\sum_{k=1}^K \omega^{(3)}_{ks}\omega^{(3)}_{ks'}$ are far from zero for many different activations after training. 
As a result, activations cannot be independent after training even if we consider the estimation error.

We obtain $C_2^{256} = 65280$ samples of $\sum_{n=1}^N (\omega^{(2)}_{nk} -  \omega^{(2)}_{nk'})$ and $b_{2k} - b_{2k'}$, and $C_2^{128} = 16256$ samples of $\sum_{k=1}^K (\omega^{(3)}_{ks} -  \omega^{(3)}_{ks'})$ and $b_{3s} - b_{3s'}$, which are shown by green triangles in Figure \ref{fig_mlp_iid2}(B) and \ref{fig_mlp_iid2}(E), respectively.
Based on the linear regression, we learn the linearities with the slope $\Delta_{f_1} = -8.75E-05$ and $\Delta_{f_2} = -2.50E-03$ from the samples.
In addition, we derive the sample mean $\overline{\boldsymbol{f_1}} = 3.96E-01$ and $\overline{\boldsymbol{f_2}} = 2.96E-01$.
We observe that $\Delta_{f_1}$ and $\Delta_{f_2}$ have huge difference as $-\overline{\boldsymbol{f_1}}$ and $-\overline{\boldsymbol{f_2}}$, respectively.
As a result, activations cannot be identically distributed after training even if we consider the estimation error.

Moreover, we demonstrate that activations being \textit{i.i.d.} is also not valid during the training procedure.
We use $r_{\boldsymbol{f_2}} = \frac{1}{\sum_{k=1}^K\sum_{k'=1}^k 1}|\sum_{n=1}^N \omega^{(2)}_{nk}\omega^{(2)}_{nk'}|$ (i.e., the mean of $|\sum_{n=1}^N \omega^{(2)}_{nk}\omega^{(2)}_{nk'}|$ over all the activations) and $r_{\boldsymbol{f_3}} = \frac{1}{\sum_{s=1}^S\sum_{s'=1}^s 1}|\sum_{s=1}^S \omega^{(3)}_{ks}\omega^{(3)}_{ks'}|$ to indicate if all the activations of $\boldsymbol{f}_2$ and $\boldsymbol{f}_3$ are independent during training.
Figure \ref{fig_mlp_iid2}(C) and \ref{fig_mlp_iid2}(F) show the variation of $r_{\boldsymbol{f_2}}$ and $r_{\boldsymbol{f_3}}$ during 301 training epochs, respectively.
At the beginning, $r_{\boldsymbol{f_2}}$ and $r_{\boldsymbol{f_3}}$ are close to zero because all the weights $\beta_{nk}$ and $\beta_{ks}$  are randomly initialized. 
However, as the training procedure goes on, $r_{\boldsymbol{f_2}}$  and $r_{\boldsymbol{f_3}}$ show an increasing trend.
Therefore, all the activations cannot keep being independent, thereby not being \textit{i.i.d.}\ during training.

Overall, since $\{{F_{2k}}\}_{k=1}^K$ and $\{{F_{3s}}\}_{s=1}^S$ cannot satisfy the necessary conditions, they cannot be \textit{i.i.d.}\ under the assumption that $\{{F_{1n}}\}_{n=1}^N$ and $\{{F_{2k}}\}_{k=1}^K$ are \textit{i.i.d.}\ during and after training.
In other words, $\{{F_{1n}}\}_{n=1}^N$, $\{{F_{2k}}\}_{k=1}^K$ and $\{{F_{3s}}\}_{s=1}^S$ cannot be simultaneously \textit{i.i.d.}\, during and after training the MLP.
Therefore, activations being \textit{i.i.d.}\ is not valid for all the hidden layers of the MLP.

\section{The equivalence between the Stochastic Gradient Descent (SGD) algorithm and the first order approximation}
\label{GD_Approx}

If an arbitrary function ${f}$ is differentiable at point ${p}^* \in {\mathbb{R}}^{N}$ and its differential is represented by the Jacobian matrix $\nabla_{p}{f}$, the first order approximation of ${f}$ near the point ${p}$ can be formulated as 
\begin{equation} 
{f}({p}) - {f}({p}^*) = (\nabla_{{p}^*}{f}) \cdot ({p} - {p}^*) + o(||{p} - {p}^*||),
\end{equation}
where $o(||{p} - {p}^*||)$ is a quantity that approaches zero much faster than $||{p} - {p}^*||$ approaches zero.

Based on the first order approximation \cite{GD_Taylor}, the activations in $\boldsymbol{f_2}$ and $\boldsymbol{f_1}$ in the $\text{MLP} = \{\boldsymbol{x}, \boldsymbol{f_1}, \boldsymbol{f_2}, \boldsymbol{f_Y}\}$ can be expressed as follows:
\begin{equation} 
\label{neuron_taylor0}
\begin{split}
\boldsymbol{f_2}[\boldsymbol{f_1}, {\theta}_{j+1}(2)] &\approx \boldsymbol{f_2}[\boldsymbol{f_1}, {\theta}_{j}(2)] + (\nabla_{{\theta}_{j}(2)} \boldsymbol{f_2}) \cdot [{\theta}_{j+1}(2) - {\theta}_{j}(2)], \\
\boldsymbol{f_1}[\boldsymbol{x}, {\theta}_{j+1}(1)] &\approx \boldsymbol{f_1}[\boldsymbol{x}, {\theta}_{j}(1)] + (\nabla_{{\theta}_{j}(1)} \boldsymbol{f_1}) \cdot [{\theta}_{j+1}(1) -{\theta}_{j}(1)], \\
\end{split}
\end{equation}
where $\boldsymbol{f_2}[\boldsymbol{f_1}, {\theta}_{j+1}(2)]$ are the activations of $\boldsymbol{f_2}$ based on the parameters of $\boldsymbol{f_2}$ learned in the $j+1$th iteration, i.e., ${\theta}_{j+1}(2)$, given the activations of $\boldsymbol{f_1}$.
The definitions of $\boldsymbol{f_2}[\boldsymbol{f_1}, {\theta}_{j}(2)]$, $\boldsymbol{f_1}(\boldsymbol{x}, {\theta}_{j+1}(1))$, and $\boldsymbol{f_1}(\boldsymbol{x}, {\theta}_{j}(1))$ are the same as $\boldsymbol{f_2}[\boldsymbol{f_1}, {\theta}_{j+1}(2)]$.

Since $\boldsymbol{f_2} = \{f_{2k} = \sigma_2(\sum_{t=1}^T\omega^{(2)}_{tk} \cdot f_{1t} + b_{2k})\}_{k=1}^K$ has $K$ neurons and each neuron has $T+1$ parameters, namely ${\theta}(2) = \{\omega^{(2)}_{1k}; \cdots; \omega^{(2)}_{Tk}; b_{2k}\}_{k=1}^K$, the dimension of $\nabla_{{\theta}_j(2)} \boldsymbol{f_2}$ is equal to $K \times (T+1)$ and $\nabla_{{\theta}_j(2)} \boldsymbol{f_2}$ can be expressed as
\begin{equation} 
\begin{split}
\nabla_{{\theta}_j(2)} \boldsymbol{f_2} =  (\nabla_{\sigma_2} \boldsymbol{f_2}) \cdot [\boldsymbol{f_1}; 1]^T
\end{split}
\end{equation}
where $\nabla_{\sigma_2} \boldsymbol{f_2} = \frac{\partial \boldsymbol{f_2}[\boldsymbol{f_1}, {\theta}_t(2)]}{\partial \sigma_2}$. 
Substituting $(\nabla_{\sigma_2} \boldsymbol{f_2}) \cdot [\boldsymbol{f_1}; 1]^T$ for $\nabla_{{\theta}_j(2)} \boldsymbol{f_2}$ in Equation \ref{neuron_taylor0}, we derive 
\begin{equation}
\label{neuron_taylor}
\begin{split}
&\boldsymbol{f_2}[\boldsymbol{f_1}, {\theta}_{j+1}(2)] \approx \boldsymbol{f_2}[\boldsymbol{f_1}, {\theta}_j(2)]\\ &+ (\nabla_{\sigma_2} \boldsymbol{f_2}) \cdot [\boldsymbol{f_1}; 1]^T \cdot {\theta}_{j+1}(2) - (\nabla_{\sigma_2} \boldsymbol{f_2}) \cdot [\boldsymbol{f_1}; 1]^T \cdot {\theta}_j(2)
\end{split}
\end{equation}

If we only consider a single neuron, e.g., $f_{2k}$, we define ${\theta}_{j+1}(2k) = [\omega^{(1)}_{1k}; \cdots; \omega^{(1)}_{Tk}; b_{2k}]$ and ${\theta}_j(2k) = [\omega'^{(1)}_{1k}; \cdots; \omega'^{(1)}_{Tk}; b'_{2k}]$, thus $[{f_1}; 1]^T \cdot {\theta}_{j+1}(2k) = \sum_{t=1}^T\omega^{(2)}_{tk} \cdot f_{1t} + b_{2k}$.
As a result, for a single neuron, Equation \ref{neuron_taylor} can be expressed as 
\begin{equation} 
\label{neuron_taylor2}
\begin{split}
&{f_{2k}}[{f_1}, {\theta}_{j+1}(2k)] \approx \underbrace{(\nabla_{\sigma_2} f_{2k}) \cdot [\sum_{t=1}^T\omega^{(2)}_{tk} \cdot f_{1t} + b_{2k}]}_{\text{Approximation}}\\ &+ \underbrace{f_{2k}[{f_1}, {\theta}_j(2k)] - (\nabla_{\sigma_2} f_{2k}) \cdot [\sum_{t=1}^T\omega'^{(2)}_{tk} \cdot f_{1t} + b'_{2k}]}_{\text{Bias}}
\end{split}
\end{equation}

Equation \ref{neuron_taylor2} indicates that ${f_{2k}}[{f_1}, {\theta}_{j+1}(2k)]$ can be reformulated as two components:  the approximation and the bias.
Since $\nabla_{\sigma_2} f_{2k} = \frac{\partial f_{2k}[{f_1}, {\theta}_j(2)]}{\partial \sigma_2}$ is only related to ${f_1}$ and ${\theta}_j(2)$, it can be regarded as a constant with respect to ${\theta}_{j+1}(2)$.
The bias component also does not contain any parameters in the $(j+1)$th training iteration. 

In summary, ${f_{2k}}({f_1}, {\theta}_{j+1}(2k))$ can be reformulated as
\begin{equation} 
\label{neuron_taylor3}
\begin{split}
{f_{2k}}({f_1}, {\theta}_{j+1}(2k)) & \approx C_1 \cdot [\sum_{t=1}^T\omega^{(2)}_{tk} \cdot f_{1t} + b_{2k}]  + C_2 \\
\end{split}
\end{equation}
where $C_1 = \nabla_{\sigma_2} f_{2k}$ and $C2 = f_{2k}({f_1}, {\theta}_j(2k)) - (\nabla_{\sigma_2} f_{2k}) \cdot [\sum_{t=1}^T\omega^{(2)}_{tk} \cdot f_{1t} + b^*_{2k}]$.
Similarly, the activations in $\boldsymbol{f_1}$ also can be formulated as the approximation.

\begin{table*}
\caption{One iteration of SGD training procedure for the MLP}
\label{backpropagation_dnn}
\vskip 0.15in
\begin{center}
\begin{small}
\begin{threeparttable}
\begin{tabular}{ccccccc}
\toprule
Layer & Gradients $\nabla_{\theta(i)} \hat{\mathcal{L}}^{\ell}_{\mathcal{S}}(h)$ & & Parameters & &Activations & \\
\midrule
${f_Y}$		&  ${\scriptstyle \nabla_{\boldsymbol{\theta}(y)} \hat{\mathcal{L}}^{\ell}_{\mathcal{S}}(h)}$ & $\downarrow$& ${\scriptstyle {\theta}_{t+1}(Y) = {\theta}_{t+1}(Y) - \alpha[\nabla_{{\theta}(y)} \hat{\mathcal{L}}^{\ell}_{\mathcal{S}}(h)]}$ & $\uparrow$ & ${\scriptstyle {f_Y}({f_2}, {\theta}_{t+1}(Y))}$ & $\uparrow$ \\
${f_2}$		& ${\scriptstyle \nabla_{\boldsymbol{\theta}(2)} \hat{\mathcal{L}}^{\ell}_{\mathcal{S}}(h)} $ & $\downarrow$& ${\scriptstyle {\theta}_{t+1}(2) = {\theta}_{t+1}(2) - \alpha[\nabla_{{\theta}(2)} \hat{\mathcal{L}}^{\ell}_{\mathcal{S}}(h)]}$ & $\uparrow$ & ${\scriptstyle {f_2}({f_1}, {\theta}_{t+1}(2))}$ & $\uparrow$ \\
${f_1}$    	& ${\scriptstyle \nabla_{\boldsymbol{\theta}(1)} \hat{\mathcal{L}}^{\ell}_{\mathcal{S}}(h)} $ & $\downarrow$& ${\scriptstyle {\theta}_{t+1}(1) = {\theta}_{t+1}(1) - \alpha[\nabla_{{\theta}(1)} \hat{\mathcal{L}}^{\ell}_{\mathcal{S}}(h)]}$ & $\uparrow$ & ${\scriptstyle {f_1}({x}, {\theta}_{t+1}(1))}$ & $\uparrow$\\
${x}$    	        & ---	& & --- & & ---\\

\bottomrule
\end{tabular}
\begin{tablenotes}
            \item The up-arrow and down-arrow indicate the order of gradients and parameters(activations) update, respectively. 
\end{tablenotes}
\end{threeparttable}
\end{small}
\end{center}
\vskip -0.1in
\end{table*}

To demonstrate the first order approximation for activations of the MLP (Equation \ref{neuron_taylor0}), we only need to prove ${\theta}_{j+1}(2) - {\theta}_{j}(2)$ approaching zero, which can be guaranteed by SGD.
Given the $\text{MLP} = \{\boldsymbol{x}, \boldsymbol{f_1}, \boldsymbol{f_2}, \boldsymbol{f_Y}\}$ and the empirical risk $\hat{\mathcal{L}}^{\ell}_{\mathcal{S}}(h)$, 
SGD aims to optimize the parameters of the MLP through minimizing $\hat{\mathcal{L}}^{\ell}_{\mathcal{S}}(h)$ \citep{backpropagation}.
\begin{equation} 
\label{gd_theta}
{\theta}_{t+1} = {\theta}_t - \alpha \nabla_{\theta_t} \hat{\mathcal{L}}^{\ell}_{\mathcal{S}}(h),
\end{equation}
where $\nabla_{\theta_t} \hat{\mathcal{L}}^{\ell}_{\mathcal{S}}(h)$ denotes the Jacobian matrix of $\hat{\mathcal{L}}^{\ell}_{\mathcal{S}}(h)$ with respect to ${\theta}_t$ at the $t$th iteration, and $\alpha > 0$ denotes the learning rate.
Since the functions of all the layers are differentiable, the Jacobian matrix of $\hat{\mathcal{L}}^{\ell}_{\mathcal{S}}(h)$ with respect to the parameters of the $i$th hidden layer, i.e., $\nabla_{\theta(i)} \hat{\mathcal{L}}^{\ell}_{\mathcal{S}}(h)$, can be expressed as

\begin{equation} 
\label{jacobian_theta}
\begin{split}
\nabla_{{\theta}(y)} \hat{\mathcal{L}}^{\ell}_{\mathcal{S}}(h) &= \nabla_{{f_Y}} \hat{\mathcal{L}}^{\ell}_{\mathcal{S}}(h)\nabla_{{\theta_Y}}f_Y\\
\nabla_{{\theta}(2)} \hat{\mathcal{L}}^{\ell}_{\mathcal{S}}(h) &= \nabla_{{f_Y}} \hat{\mathcal{L}}^{\ell}_{\mathcal{S}}(h)\nabla_{f_2}f_Y \nabla_{\theta_2} f_2\\
\nabla_{{\theta}(1)} \hat{\mathcal{L}}^{\ell}_{\mathcal{S}}(h) &= \nabla_{{f_Y}} \hat{\mathcal{L}}^{\ell}_{\mathcal{S}}(h) \nabla_{{f_2}} f_Y \nabla_{{f_1}} f_2 \nabla_{\theta_1} f_1,
\end{split}
\end{equation}
where ${\theta}(i)$ denote the parameters of the $i$th layer.
Equation \ref{gd_theta} and \ref{jacobian_theta} indicate that ${\theta}(i)$ can be learned as
\begin{equation} 
{\theta}_{t+1}(i) = {\theta}_t(i) - \alpha [\nabla_{\theta_t(i)} \hat{\mathcal{L}}^{\ell}_{\mathcal{S}}(h)].
\end{equation}
Table \ref{backpropagation_dnn} summarizes SGD training procedure for the MLP shown in Figure \ref{fig_mlp_bhm}.
SGD minimizing $\hat{\mathcal{L}}^{\ell}_{\mathcal{S}}(h)$ makes $\nabla_{\theta_t(i)} \hat{\mathcal{L}}^{\ell}_{\mathcal{S}}(h)$ to converging zero, thereby ${\theta}_{t+1}(i) - {\theta}_t(i)$ converging to zero.

\pagebreak
 
\section{The Gibbs explanation for the entire architecture of the MLP}
\label{posterior_MLP}

Since the entire architecture of the $\text{MLP} = \{\boldsymbol{x}, \boldsymbol{f_1}, \boldsymbol{f_2}, \boldsymbol{f_Y}\}$ in Figure \ref{fig_mlp_bhm} corresponds to a joint distribution 
\begin{equation} 
P(F_Y; F_2; F_1|{X}) = P(F_Y|F_2)P(F_2|F_1)P(F_1|{X}),
\end{equation}
the marginal distribution $P(F_Y|{X})$ can be formulated as 
\begin{equation} 
\begin{split}
P_{F_Y|X}(l|{x}) &= \sum_{k=1}^K \sum_{t=1}^T P(F_Y = l, F_2 = k, F_1=t|X={x})\\ &= \sum_{k=1}^K P_{F_Y|F_2}(l|k) \sum_{t=1}^T P_{F_2|F_1}(k|t)P_{F_1|X}(t|{x}).
\end{split}
\end{equation}

Based on the definition of the Gibbs probability measure (Equation \ref{Gibbs_f}), we have
\begin{equation} 
P_{F_1|X}(t|{x}) = \frac{1}{Z_{{F_1}}}\text{exp}(f_{1t}) = \frac{1}{Z_{{F_1}}}\text{exp}[\sigma_1(\langle {\boldsymbol{\omega}'}^{(1)}_{t}, {x}' \rangle)],
\end{equation}
where ${\boldsymbol{\omega}'}^{(1)}_{t} = [{\boldsymbol{\omega}}^{(1)}_{t}, b_{1n}]$ and ${x}' = [{x}, 1]$, i.e., $\langle {\boldsymbol{\omega}'}^{(1)}_{t}, {x}' \rangle = \langle {\boldsymbol{\omega}}^{(1)}_{t}, {x} \rangle + b_{1n}$.
Similarly, we have
\begin{equation} 
P_{F_2|F_1}(k|t) = \frac{1}{Z_{{F_2}}}\text{exp}(f_{2k}) = \frac{1}{Z_{{F_2}}}\text{exp}[\sigma_2(\langle {\boldsymbol{\omega}'}^{(2)}_{k}, {f}'_1 \rangle)],
\end{equation}
where ${f}_1 = \{f_{1t}\}_{t=1}^T = \{\sigma_1(\langle {\boldsymbol{\omega}'}^{(1)}_{t}, {x}' \rangle)\}_{t=1}^T$, ${\boldsymbol{\omega}'}^{(2)}_{k} = [{\boldsymbol{\omega}}^{(2)}_{k}, b_{2k}]$ and ${f}'_1 = [{f}_1, 1]$, i.e., $\langle {\boldsymbol{\omega}'}^{(2)}_{k}, {f}'_1 \rangle = \langle {\boldsymbol{\omega}}^{(2)}_{k}, {f}_1 \rangle + b_{2k}$, thus we have 
\begin{equation} 
\begin{split}
&\sum_{t=1}^T P_{F_2|F_1}(k|t)P_{F_1|X}(t|{x})\\
& = \frac{1}{Z_{{F_2}}}\frac{1}{Z_{{F_1}}}\sum_{t=1}^T\text{exp}[\sigma_2(\langle {\boldsymbol{\omega}'}^{(2)}_{k}, {f}'_1 \rangle)]\text{exp}[\sigma_1(\langle {\boldsymbol{\omega}'}^{(1)}_{t}, {x}' \rangle)]. \\
\end{split}
\end{equation}

Since $\langle {\boldsymbol{\omega}'}^{(2)}_{k}, {f}'_1 \rangle = \langle {\boldsymbol{\omega}}^{(2)}_{k}, {f}_1 \rangle + b_{2k} = \sum_{t=1}^T\omega^{(2)}_{kt}f_{1t} + b_{2k}$ is a constant with respect to $t$, we have
\begin{equation} 
\begin{split}
&\sum_{t=1}^T P_{F_2|F_1}(k|t)P_{F_1|X}(t|{x})\\ &= \frac{1}{Z_{{F_2}}}\frac{1}{Z_{{F_1}}}\text{exp}[\sigma_2(\langle {\boldsymbol{\omega}'}^{(2)}_{k}, {f}'_1 \rangle)]\sum_{t=1}^T\text{exp}[\sigma_1(\langle {\boldsymbol{\omega}'}^{(1)}_{t}, {x}' \rangle)]. 
\end{split}
\end{equation}
In addition, $\sum_{t=1}^T\text{exp}[\sigma_1(\langle {\boldsymbol{\omega}'}^{(1)}_{t}, {x}' \rangle)] = Z_{F_1}$, thus we have 
\begin{equation}
\sum_{t=1}^T P_{F_2|F_1}(k|t)P_{F_1|X}(t|{x}) = \frac{1}{Z_{{F_2}}}\text{exp}[\sigma_2(\langle {\boldsymbol{\omega}'}^{(2)}_{k}, {f}'_1 \rangle)].
\end{equation}
Therefore, we can simplify $P_{F_Y|X}(l|{x})$ as
\begin{equation}
\begin{split}
P_{F_Y|X}(l|{x}) &= \sum_{k=1}^K P_{F_Y|F_2}(l|k) \sum_{t=1}^T P_{F_2|F_1}(k|t)P_{F_1|X}(t|{x})\\ &= \sum_{k=1}^K P_{F_Y|F_2}(l|k) \frac{1}{Z_{{F_2}}}\text{exp}[\sigma_2(\langle {\boldsymbol{\omega}'}^{(2)}_{k}, {f}'_1 \rangle)].
\end{split}
\end{equation}
Similarly, since $P_{F_Y|F_2}(l|k) = \frac{1}{Z_{{F_Y}}}\text{exp}[\sigma_3(\langle {\boldsymbol{\omega}}^{(3)}_{l}, {f}_2 \rangle + b_{yl})]$ and $\langle {\boldsymbol{\omega}}^{(3)}_{l}, {f}_2 \rangle = \sum_{k=1}^K\omega^{(3)}_{lk}f_{2k}$ is also a constant with respect to $k$, we can derive 
\begin{equation}
\begin{split}
P_{F_Y|X}(l|{x}) = P_{F_Y|F_2}(l|k) = \frac{1}{Z_{F_Y}}\text{exp}[\langle {\boldsymbol{\omega}}^{(3)}_{l}, {f}_2 \rangle + b_{yl}].
\end{split}
\end{equation}

In addition, since ${f}_2 = \{f_{2k}\}_{k=1}^K = \{\sigma_2(\langle {\boldsymbol{\omega}}^{(2)}_{k}, {f}_1 \rangle + b_{2k}) \}_{k=1}^K$, we can extend $P_{F_Y|X}(l|{x}_i)$ as 
\begin{equation}
\begin{split}
&P_{F_Y|X}(l|{x}) = P_{F_Y|F_2}(l|k) = \frac{1}{Z_{F_Y}}\text{exp}[\langle {\boldsymbol{\omega}}^{(3)}_{l}, {f}_2 \rangle + b_{yl}]\\
&= \frac{1}{Z_{F_Y}}\text{exp}[\langle {\boldsymbol{\omega}}^{(3)}_{l}, \left( \begin{array}{c} \sigma_2(\langle {\boldsymbol{\omega}}^{(2)}_{1}, {f}_1 \rangle + b_{21}) \\ \vdots \\ \sigma_2(\langle {\boldsymbol{\omega}}^{(2)}_{K}, {f}_1 \rangle + b_{2K}) \end{array} \right) \rangle + b_{yl}].
\end{split}
\end{equation}

Since ${f}_1 = \{f_{1t}\}_{t=1}^T = \{\sigma_1(\langle {\boldsymbol{\omega}}^{(1)}_{t}, {x} \rangle + b_{1n})\}_{t=1}^T$, we can further extend $P_{F_Y|X}(l|{x})$ as
\begin{table*}
\centering
\begin{center}
\begin{tabular}{p{16cm}}
\begin{equation}
\begin{split}
P_{F_Y|X}(l|{x})
= \frac{1}{Z_{F_Y}}\text{exp}[\langle {\boldsymbol{\omega}}^{(3)}_{l}, \left( \begin{array}{c} \sigma_2(\langle {\boldsymbol{\omega}}^{(2)}_{1}, \left( \begin{array}{c} \sigma_1(\langle {\boldsymbol{\omega}}^{(1)}_{1}, {x} \rangle + b_{11}) \\ \vdots \\ \sigma_1(\langle {\boldsymbol{\omega}}^{(1)}_{1}, {x} \rangle + b_{11}) \end{array} \right) \rangle + b_{21}) \\ \vdots \\ \sigma_2(\langle {\boldsymbol{\omega}}^{(2)}_{K}, \left( \begin{array}{c} \sigma_1(\langle {\boldsymbol{\omega}}^{(1)}_{T}, {x} \rangle + b_{1T}) \\ \vdots \\ \sigma_1(\langle {\boldsymbol{\omega}}^{(1)}_{T}, {x} \rangle + b_{1T}) \end{array} \right) \rangle + b_{2K}) \end{array} \right) \rangle + b_{yl}]
= \frac{1}{Z_{\text{MLP}}(x_i)}\text{exp}[{f_{yl}(f_2(f_1({x})))}].
\end{split}\end{equation}\\
\end{tabular}
\end{center}
\end{table*}

Overall, we prove $P_{F_Y|X}(l|{x})$ as a Gibbs distribution and it can be expressed as 
\begin{equation}
\label{posterior_dnn} 
P_{F_Y|X}(l|{x}) = \frac{1}{Z_{\text{MLP}}({x}_i)}\text{exp}[f_{yl}(f_2(f_1({x})))].
\end{equation}
where $E_{yl}(x) = - f_{yl}(f_2(f_1({x})))$ is the energy function of $l \in \mathcal{Y}$ given $x$ and the partition function 
\begin{equation}
\begin{split}
Z_{\text{MLP}}({x}) &= \sum_{l=1}^L\sum_{k=1}^K\sum_{t=1}^TQ(F_Y,F_2,F_1|X=x)\\ 
&= \sum_{l=1}^L\text{exp}[f_{yl}(f_2(f_1({x})))].
\end{split}
\end{equation}


\section{The gradient of the cross entropy loss function with respect to the weights }
\label{bp}

If the loss function is the cross entropy, we have
\begin{equation}
\ell = H[P_{Y|\boldsymbol{X}}(l|\boldsymbol{x}), \boldsymbol{f_{y}}(\boldsymbol{x})],
\end{equation}
where $\boldsymbol{f_{y}(x)} = \{f_{yl}\}_{l=1}^L$ is the output of the MLP given $\boldsymbol{x}$, and $P_{Y|\boldsymbol{X}}(l|\boldsymbol{x})$ is the one-hot probability of $\boldsymbol{x}$ given the label ${y}$, i.e., if $l = y$,$P_{Y|\boldsymbol{X}}(l|\boldsymbol{x}) = 1$, otherwise $P_{Y|\boldsymbol{X}}(l|\boldsymbol{x}) = 0$.

Based on the definition of the cross entropy, $\ell$ can be expressed as
\begin{equation}
\ell = -\sum_{l=1}^LP_{Y|\boldsymbol{X}}(l|\boldsymbol{x})\text{log}f_{yl}.
\end{equation}
Therefore, the derivative of $\ell$ with respect to $f_{yl}$ is 
\begin{equation}
\frac{\partial \ell}{\partial f_{yl}} = -\frac{P_{Y|\boldsymbol{X}}(l|\boldsymbol{x})}{f_{yl}}.
\end{equation}
In addition, we have 
\begin{equation}
\frac{\partial f_{yt}}{\partial g_{yl}} = \frac{\frac{1}{Z_Y}\text{exp}(g_{yt})}{\partial g_{yl}} = \left\{ \begin{array}{rcl}
         f_{yl}(1-f_{yl}) & \mbox{for}& t=l \\ 
         -f_{yl}f_{yt} & \mbox{for} & t \neq l
                \end{array}\right..
\end{equation}
As a result, the derivative of $\ell$ with respect to $g_{yl}$ can be expressed as
\begin{equation}
\begin{split}
\frac{\partial \ell}{\partial g_{yl}} &= \sum_{t=1}^L\frac{\partial \ell}{\partial f_{yt}} \frac{\partial f_{yt}}{\partial g_{yl}}\\
&= -P_Y(l)(1-f_{yl}) + \sum_{t\neq l}{P_{Y|\boldsymbol{X}}(t|\boldsymbol{x})}f_{yl}\\
&= f_{yl} - P_{Y|\boldsymbol{X}}(l|\boldsymbol{x})
\end{split}
\end{equation}

Therefore, the derivative of $\ell$ with respect to $\omega^{(3)}_{kl}$ can be expressed as 
\begin{equation}
\frac{\partial \ell}{\partial \omega^{(3)}_{kl}} = \sum_{l=1}^L\frac{\partial \ell}{\partial g_{yl}} \frac{\partial g_{yl}}{\partial \omega^{(3)}_{kl}} = [f_{yl} - P_{Y|\boldsymbol{X}}(l|\boldsymbol{x})]f_{2k}.
\end{equation}

Similarly, the derivative of $\ell$ with respect to $g_{2k}$ can be expressed as 
\begin{equation}
\begin{split}
\frac{\partial \ell}{\partial g_{2k}} &= \sum_{l=1}^L\frac{\partial \ell}{\partial g_{yl}} \frac{\partial g_{yl}}{\partial f_{2k}} \frac{\partial f_{2k}}{\partial g_{2k}}\\
&= \sum_{l=1}^L [f_{yl} - P_{Y|\boldsymbol{X}}(l|\boldsymbol{x})]\omega^{(3)}_{kl}\sigma'_2(g_{2k}).
\end{split}
\end{equation}
The derivative of $\ell$ with respect to $\omega^{(2)}_{nk}$ can be expressed as 
\begin{equation}
\begin{split}
\frac{\partial \ell}{\partial \omega^{(2)}_{nk}} &= \frac{\partial \ell}{\partial g_{2k}}\frac{\partial g_{2k}}{\partial \omega^{(2)}_{nk}}\\
&= \sum_{l=1}^L [f_{yl} - P_{Y|\boldsymbol{X}}(l|\boldsymbol{x})]\omega^{(3)}_{kl}\sigma'_2(g_{2k})f_{1n}
\end{split}
\end{equation}

Similarly, the derivative of $\ell$ with respect to $g_{1n}$ can be expressed as 
\begin{equation}
\begin{split}
\frac{\partial \ell}{\partial g_{1n}} &= \sum_{k=1}^K\frac{\partial \ell}{\partial g_{2k}} \frac{\partial g_{2k}}{\partial f_{1n}} \frac{\partial f_{1n}}{\partial g_{1n}}\\
&= \sum_{k=1}^K \sum_{l=1}^L [f_{yl} - P_{Y|\boldsymbol{X}}(l|\boldsymbol{x})]\omega^{(3)}_{kl}\sigma'_2(g_{2k})\omega^{(2)}_{nk}\sigma'_1(g_{1n}).
\end{split}
\end{equation}
The derivative of $\ell$ with respect to $\omega^{(1)}_{mn}$ can be expressed as 
\begin{equation}
\begin{split}
\frac{\partial \ell}{\partial \omega^{(1)}_{mn}} &= \frac{\partial \ell}{\partial g_{1n}}\frac{\partial g_{1n}}{\partial \omega^{(1)}_{mn}}\\
&= \sum_{k=1}^K \sum_{l=1}^L [f_{yl} - P_{Y|\boldsymbol{X}}(l|\boldsymbol{x})]\omega^{(3)}_{kl}\sigma'_2(g_{2k})\omega^{(2)}_{nk}\sigma'_1(g_{1n})x_m.
\end{split}
\end{equation}

Based on the back-propagation algorithm, weights are updated as 
\begin{equation}
\begin{split}
\omega^{(1)}_{mn}(t+1) &= \omega^{(1)}_{mn}(t) - \alpha \frac{\partial \ell}{\partial \omega^{(1)}_{mn}(t)}\\
\omega^{(2)}_{nk}(t+1) &= \omega^{(2)}_{nk}(t) - \alpha \frac{\partial \ell}{\partial \omega^{(2)}_{nk}(t)}\\
\omega^{(3)}_{kl}(t+1) &= \omega^{(3)}_{kl}(t) - \alpha \frac{\partial \ell}{\partial \omega^{(3)}_{kl}(t)}
\end{split}
\end{equation}
where $\alpha$ is the learning rate and $t$ is the $t$th training iteration.

\section{$H(Y) = I(\boldsymbol{X}, Y)$}
\label{YbelongstoX}

In this section, we prove all the information of $Y$ stems from $\boldsymbol{X}$, i.e., $H(Y) = I(\boldsymbol{X}, Y)$.
Based on the definition of mutual information, we have
\begin{equation}
 I(\boldsymbol{X}, Y) = H(Y) - H(Y|\boldsymbol{X}),
\end{equation}
thus $H(Y) = I(\boldsymbol{X}, Y)$ is equivalent to $H(Y|\boldsymbol{X}) = 0$.

Based on the definition of conditional entropy, we have
\begin{equation}
H(Y|\boldsymbol{X}) = \sum_{\boldsymbol{x} \in \mathcal{{X}}}P(\boldsymbol{X}=\boldsymbol{x})H(Y|\boldsymbol{X}=\boldsymbol{x})
\end{equation}
where $H(Y|\boldsymbol{X}=\boldsymbol{x})$ can be formulated as
\begin{equation}
H(Y|\boldsymbol{X}=\boldsymbol{x}) = \sum_{y \in \mathcal{{Y}}}P(Y=y)\text{log}_2P(Y=y|\boldsymbol{X}=\boldsymbol{x}).
\end{equation}

Since $\mathcal{Y} = \{1, \cdots, L\}$ and $(\boldsymbol{x}^j, y^j) \in \mathcal{D}$ are \textit{i.i.d.}, we can simplify $H(Y|\boldsymbol{X}=\boldsymbol{x})$ as
\begin{equation}
H(Y|\boldsymbol{X}=\boldsymbol{x}) = \sum_{l =1}^L\frac{N(l)}{J}\text{log}_2P(Y=l|\boldsymbol{X}=\boldsymbol{x}),
\end{equation}
where $N(l)$ is the number of labels $y^j = l$ and $J$ is the total number of samples.

Since $P(Y|\boldsymbol{X})$ is one-hot format, i.e.,
\begin{equation}
P_{Y|\boldsymbol{X}}(l|\boldsymbol{x}^j) =     \left\{ \begin{array}{rcl}
         1 & \text{if} & l = y^j \\ 
         0 & \text{if}  & l \neq y^j
                \end{array}\right.
\end{equation}
We can derive $H(Y|\boldsymbol{X}=\boldsymbol{x}) = 0$, thereby $H(Y|\boldsymbol{X}) = 0$.
Finally, we have $H(Y) = I(\boldsymbol{X}, Y)$.

\begin{figure*}
	\begin{minipage}[b]{0.99\linewidth}
  		\centering
  		\centerline{\includegraphics[scale=0.5]{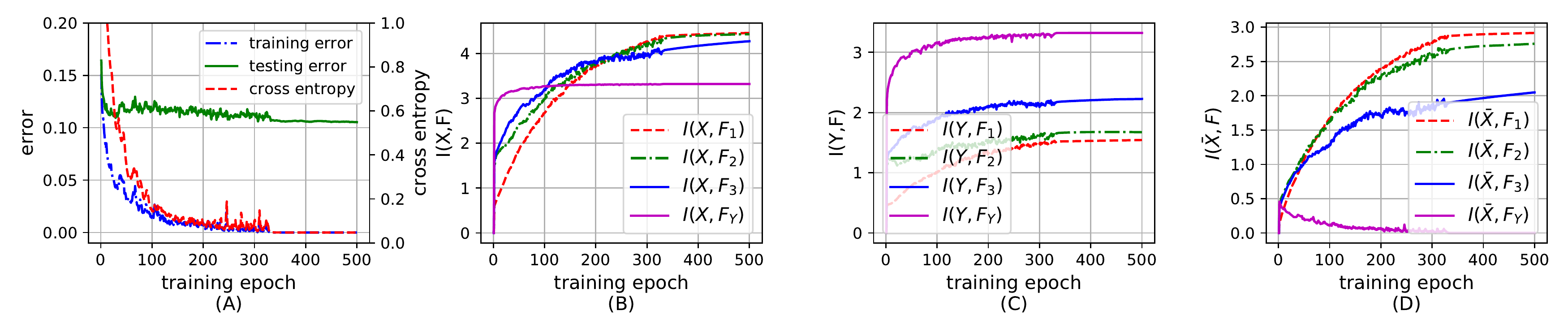}}
	\end{minipage}
	\vfill	
	\begin{minipage}[b]{0.99\linewidth}
		\centering
		\centerline{\includegraphics[scale=0.5]{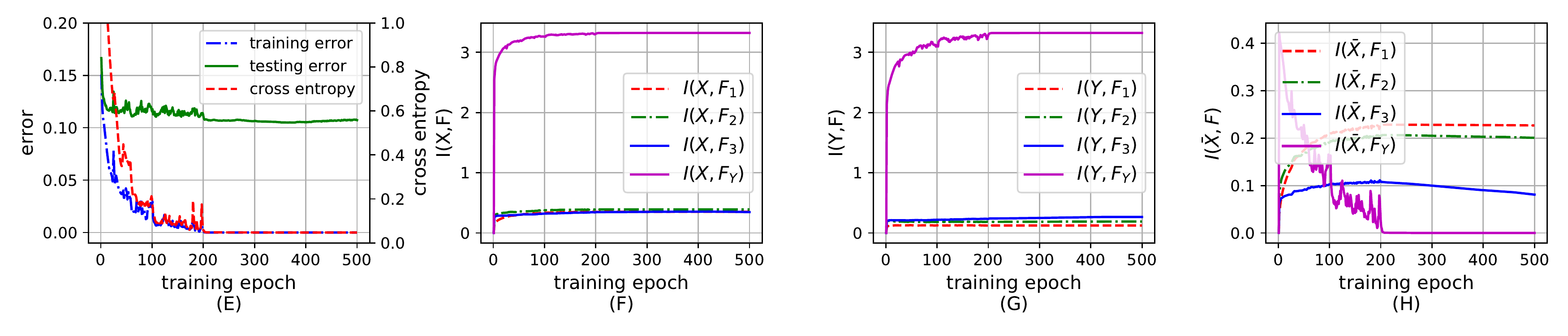}}		
	\end{minipage}
	\vfill	
	\begin{minipage}[b]{0.99\linewidth}
		\centering
		\centerline{\includegraphics[scale=0.5]{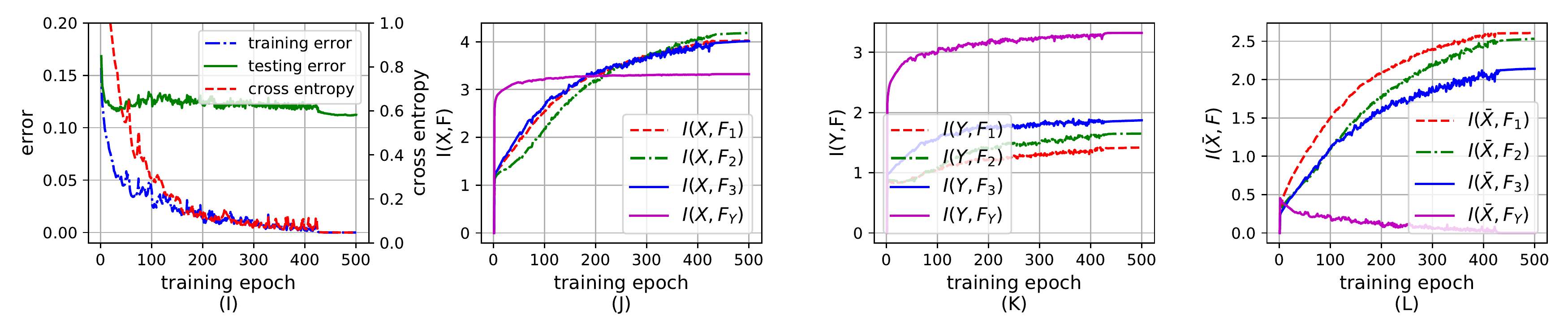}}		
	\end{minipage}
\caption{(A), (E), and (I) visualize the variation of the training/testing error and the cross entropy loss of MLP8, MLP9 and MLP10 during training, respectively.
(B), (F), and (J) visualize the variation of $I(\boldsymbol{X}, F_i)$ over all the layers in MLP8, MLP9, and MLP10, respectively.
(C), (G), and (K) visualize the variation of $I(Y, F_i)$ over all the layers in MLP8, MLP9, and MLP10, respectively.
(D), (H), and (L) visualize the variation of $I(\bar{X}, F_i)$ over all the layers in MLP8, MLP9, and MLP10, respectively.}
\label{if_mlp_fmnist}
\end{figure*} 

\section{Information theoretic explanations for MLPs on Fashion-MNIST dataset}
\label{it_mlps_fmnist}

In this section, we design three MLPs on Fashion-MNIST dataset to demonstrate the proposed explanations for MLPs: (i) the information flow of $\boldsymbol{X}$ and ${Y}$ in MLPs (Section \ref{info_flowyx} and \ref{limitation_IB}) and (ii) the information theoretic explanations for generalization (Section \ref{generalization_it}).

\subsection{The information flow in the MLPs}

To classify the Fashion-MNIST dataset, we design three $\text{MLPs} = \{\boldsymbol{x}, \boldsymbol{f_1}, \boldsymbol{f_2}, \boldsymbol{f_3}, \boldsymbol{f_Y}\}$, i.e., MLP8, MLP9, and MLP10, and their differences are summarized in Table \ref{MLP3_FMNIST}.
All the weights of the MLPs are randomly initialized by truncated normal distributions.
We still choose the Adam method to learn the weights of the MLPs on MNIST dataset over 500 epochs with the learning rate 0.001, and use the same method as Section \ref{if_mlp_synthetic} to derive $I(\boldsymbol{X}, F_i)$,  $I(Y, F_i)$, and $I(\bar{X}, F_i)$ based on Equation (\ref{mi_x}), (\ref{mi_y}), (\ref{xyxc}), respectively.

\begin{table}
\caption{The number neurons(nodes) of each layer and the activation functions in each MLP.}
\label{MLP3_FMNIST}
\vskip 0.1in
\begin{center}
\begin{small}
\begin{threeparttable}
\begin{tabular}{ccccccc}
\toprule
 & $\boldsymbol{x}$ & $\boldsymbol{f}_1$ & $\boldsymbol{f}_2$ & $\boldsymbol{f}_3$ & $\boldsymbol{f}_Y$ & $\sigma(\cdot)$ \\
\midrule
MLP8 & 784 & 256 & 128 & 96 & 10  & ReLU\\
MLP9 & 784 & 256 & 128 & 96 & 10  & Tanh \\
MLP10 & 784 & 96 & 128 & 256 & 10  & ReLU \\
\bottomrule
\end{tabular}
\end{threeparttable}
\end{small}
\end{center}
\vskip -0.1in
\end{table}

The information flow in the MLPs on Fashion-MNIST dataset is consistent with the results on the synthetic dataset.
More specifically, Figure \ref{if_mlp_fmnist}(B), \ref{if_mlp_fmnist}(F) and \ref{if_mlp_fmnist}(J) visualize three different information flows of $\boldsymbol{X}$ in MLP8, MLP9, and MLP10, respectively, which confirms that the information flow of $\boldsymbol{X}$ in MLPs does not satisfy any DPI.
Figure \ref{if_mlp_fmnist}(C), \ref{if_mlp_fmnist}(G) and \ref{if_mlp_fmnist}(K) demonstrate that the information flow $Y$ satisfies $I(Y, F_Y) \geq I(Y, F_3) \geq I(Y, F_2) \geq I(Y, F_1)$ in all the tree MLPs.
The experiment further demonstrates that IB cannot correctly explain the information flow of $\boldsymbol{X}$ and ${Y}$ in MLPs, because they cannot satisfy the DPIs (Equation \ref{dpi_dnn}) derived from IB in Figure \ref{if_mlp_fmnist}(B, C), \ref{if_mlp_fmnist}(F, G) and \ref{if_mlp_fmnist}(J, K).
In addition, Figure \ref{if_mlp_fmnist}(D), \ref{if_mlp_fmnist}(H) and \ref{if_mlp_fmnist}(J) demonstrate that the information flow $\bar{X}$ in all the tree MLPs satisfies $I(\bar{X}, F_1) \geq I(\bar{X}, F_2) \geq I(\bar{X}, F_3) \geq I(\bar{X}, F_Y)$.

\subsection{The information theoretic explanation for the generalization performance of MLPs}

\begin{figure*}
\centering
\includegraphics[scale=0.65]{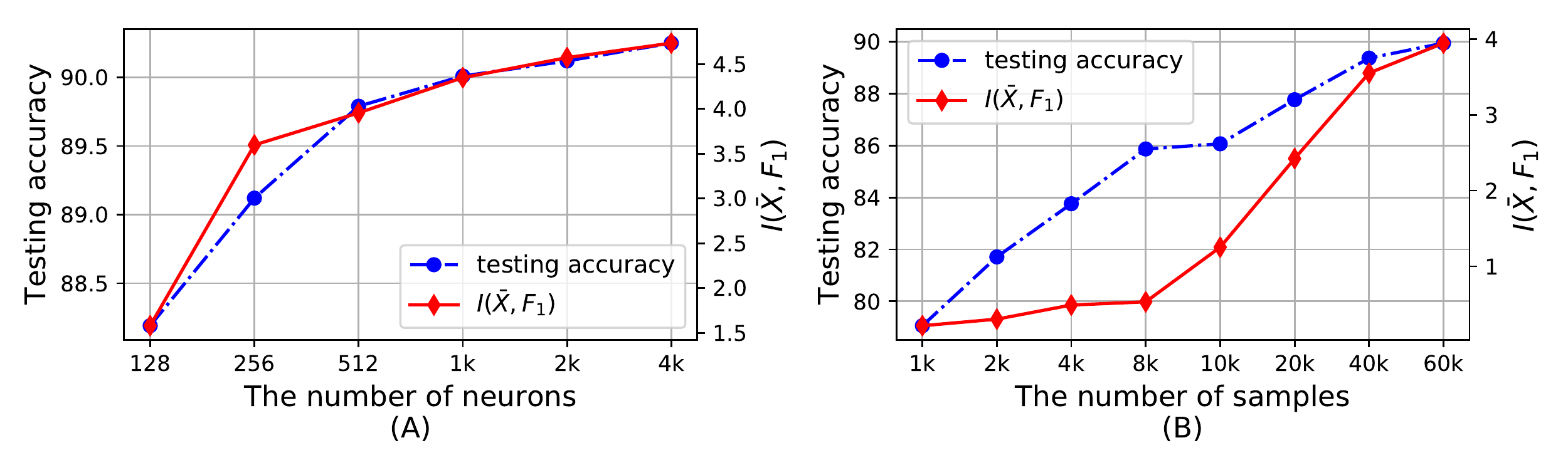}
\caption{\small{
(A) shows the variation of the testing accuracy and $I(\bar{X}, F_1)$ given different MLPs with different number of neurons.
(B) shows the variation of the testing accuracy and $I(\bar{X}, F_1)$ given different number of training samples.
}}
\label{Img_mixbar_neurons_samples_fmnist}
\end{figure*}

First, $I(\bar{X}, F_1)$ can measure the generalization of MLPs with different number of neurons in MLPs.
In general, a MLP with more neurons would have better generalization, thus $I(\bar{X},F_1)$ of the MLP should be larger.
We design six different $\text{MLPs} = \{\boldsymbol{x}, \boldsymbol{f}_1, \boldsymbol{f}_2, \boldsymbol{f}_3, \boldsymbol{f}_Y\}$. 
The number of neurons in the three hidden layers of the six MLPs has the same ratio, i.e., $\#(\boldsymbol{f}_1): \#(\boldsymbol{f}_2): \#(\boldsymbol{f}_3) = 4:3:1$.
However, different MLPs have different number of neurons, especially $\#(\boldsymbol{f}_1) = \{64,128,256,512, 1024, 2048\}$.
After all the six MLPs achieve 100\% training accuracy on MNIST dataset, we observe a positive correlation with the testing accuracy and $I(\bar{X},F_1)$ in Figure \ref{Img_mixbar_neurons_samples_fmnist}(A).

Second, $I(\bar{X}, F_1)$ can measure the generalization of MLPs with different number of training samples.
In general, a MLP with larger number of training samples would have better generalization performance, thus $I(\bar{X},F_1)$ of the MLP should be larger.
We generate 8 different training sets with different number of MNIST training samples and train MLP8 on the 8 training sets.
After MLP8 achieves 100\% training accuracy on the 8 training sets, we also observe a positive correlation with the testing accuracy and $I(\bar{X},F_1)$ in Figure \ref{Img_mixbar_neurons_samples_fmnist}(B).

In summary, $I(\bar{X}, F_1)$ demonstrate positive correlation with the testing error of MLPs, which keeps consistent with results based on MLPs on MNIST dataset. 
The experiment further confirms that $I(\bar{X}, F_1)$ can be viewed as a criterion for the generalization of MLPs. 

\pagebreak
\bibliographystyle{cas-model2-names}
\bibliography{it-cnn}




\end{document}